\documentclass[]{interact}

\usepackage{epstopdf}
\usepackage[caption=false]{subfig}

\usepackage[numbers,sort&compress]{natbib}
\bibpunct[, ]{[}{]}{,}{n}{,}{,}
\renewcommand\bibfont{\fontsize{10}{12}\selectfont}

\usepackage[utf8]{inputenc} 
\usepackage[T1]{fontenc}    
\usepackage{hyperref}       
\usepackage{url}            
\usepackage{booktabs}       
\usepackage{amsfonts}       
\usepackage{nicefrac}       
\usepackage{microtype}      
\usepackage{xcolor}         
\usepackage{amsmath}
\usepackage{amssymb}
\usepackage[colorinlistoftodos,bordercolor=orange,backgroundcolor=orange!20,linecolor=orange,textsize=scriptsize]{todonotes}
\usepackage{pifont}
\usepackage{xcolor}
\usepackage{algorithm}
\usepackage{algorithmic}

\newcommand{\minmax}{\mathop{\min\!\max}}
\newcommand{\EndProof}{\begin{flushright}$\square$\end{flushright}}
\allowdisplaybreaks

\newcommand{\E}{{\mathbb E}}

\def\X{\mathcal X}
\def\Y{\mathcal Y}

\def\E{\mathbb E}
\def\EE{\mathbb E}
\def\PP{\mathbb P}

\newcommand{\ans}[1]{{\color{black}#1}}
\newcommand{\add}[1]{{\color{black}#1}}

\theoremstyle{plain}
\newtheorem{theorem}{Theorem}[section]
\newtheorem{lemma}[theorem]{Lemma}
\newtheorem{corollary}[theorem]{Corollary}

\theoremstyle{definition}
\newtheorem{definition}[theorem]{Definition}

\theoremstyle{remark}

\newtheorem{innercustomass}{Assumption}
\newenvironment{customass}[1]
  {\renewcommand\theinnercustomass{#1}\innercustomass}
  {\endinnercustomass}

\begin{document}


\title{Distributed Saddle Point Problems: \\ Lower Bounds, Near-Optimal and Robust Algorithms}

\author{
    \name{
        Aleksandr Beznosikov\textsuperscript{a,b,c,d},
        Valentin Samokhin\textsuperscript{e,f}
        Alexander Gasnikov\textsuperscript{d,g,h}
    }
    \affil{
        \textsuperscript{a} Laboratory of Federated Learning Problems, Ivannikov Institute for System Programming of the RAS, Moscow, Russia;\\
        \textsuperscript{b} BRAIn Lab, Moscow Institute of Physics and Technology, Moscow, Russia;\\
        \textsuperscript{c} Center for Applied Artificial Intelligence, The Russian Presidential Academy of National Economy and Public Administration, Moscow, Russia;\\
        \textsuperscript{d} Center for Artificial Intelligence, Innopolis University, Innopolis, Russia; \\
        \textsuperscript{e} AGI Med Lab, Artificial Intelligence Research Institute, Moscow, Russia;\\
        \textsuperscript{f} Department of Data Analysis in Neuroscience, Institute for Information Transmission Problems of the RAS, Moscow, Russia;\\
        \textsuperscript{g} Artificial Intelligence Center, Ivannikov Institute for System Programming of the RAS, Moscow, Russia;\\
        \textsuperscript{h} Laboratory of Mathematical Methods of Optimization, Moscow Institute of Physics and Technology, Moscow, Russia;
    }
}

\maketitle

\begin{abstract}
 This paper focuses on the distributed optimization of stochastic saddle point problems. The first part of the paper is devoted to lower bounds for the centralized and decentralized distributed methods for smooth (strongly) convex-(strongly) concave saddle point problems, as well as the near-optimal algorithms by which these bounds are achieved. Next, we present a new federated algorithm for centralized distributed saddle-point problems -- Extra Step Local SGD. The theoretical analysis of the new method is carried out for strongly convex-strongly concave and non-convex-non-concave problems. In the experimental part of the paper, we show the effectiveness of our method in practice. In particular, we train GANs in a distributed manner. 
\end{abstract}

\begin{keywords}
distributed optimization; saddle point problems; lower and upper bounds; local methods; convex optimization; stochastic optimization
\end{keywords}

\section{Introduction}

Distributed algorithms have already become an integral part of solving many applied tasks, including machine learning problems \cite{shalev2014understanding, mcdonald2010distributed, mcmahan2017communication}. This paper also deals with distributed methods, we study the saddle point problem (SPP):
\begin{equation}
\label{distr}
    \min_{x \in \X} \max_{y \in \Y} f(x,y) := \frac{1}{M} \sum\limits_{m=1}^M f_m(x,y), 
\end{equation}
where parts of the function $f$ are distributed among $M$ \ans{devices/nodes/workers/machines}, while the function $f_m$ corresponds to the device with number $m$. SPPs, including distributed ones, have many applications. Here we can mention the already classical and long-established applications in equilibrium theory, games and economics \cite{facchinei2007finite}, as well as new and recent trends in image deconvolution \cite{chambolle2011first,esser2010general}, reinforcement and statistical learning \cite{Abadeh,pmlr-v119-jin20f}, adversarial training \cite{Madry2017:adv} and GANs \cite{goodfellow2014generative}. In particular, a number of papers \cite{daskalakis2017training,gidel2018variational,mertikopoulos2018optimistic,chavdarova2019reducing,pmlr-v89-liang19b,peng2020training} showed the connection of the theory for convex SPPs with the training of GANs and provided insights and useful hints for the GANs community. From a machine learning point of view, it can be interesting if $f_m$ is an empirical loss function of the model on the local data of the $m$th device. Therefore, we consider the statement of the problem \eqref{distr} when we have access only to the local stochastic oracle of $f_m(x,y):= \E_{\xi_m \sim \mathcal{D}_m} [f_m(x,y,\xi_m)]$, where the data $\xi_m$ follow unknown distributions $\mathcal{D}_m$.

However, the main problem of distributed learning tasks is not the stochasticity, but the separation of the problem. \ans{The $m$th device only has access to information about the nature of its local distribution $\mathcal{D}_m$. We assume that on the $m$th node it is possible to sample online from $\mathcal{D}_m$ or to interdependently work with already prepared data from this distribution. It is important that all other devices do not know $\mathcal{D}_m$, moreover, transferring local data to other devices may be inefficient or impossible for privacy reasons.} Therefore, to solve \eqref{distr}, it becomes necessary to construct a distributed algorithm that combines local computation on each of the devices and communication between them. Such an algorithm can be organized as follows: all devices communicate only with the main device (server). This approach is called centralized. The main problem is the importance of the server -- it can crash and interrupt the whole process. Therefore, along with the centralized approach, the decentralized \cite{gallager1983distributed} one is also popular. In this case, all devices are connected \add{by} a network, communication occurs along the edges of this network.

Both centralized and decentralized methods are well developed for minimization problems. \ans{However, the direction of distributed algorithms for SPPs has been much less studied.} Our work makes the following contribution to this area.

\subsection{Our contributions} \label{sec:contr}


\hspace{0.45cm}$\bullet$ \textbf{Lower bounds. } 
We present lower bounds for distributed stochastic smooth strongly convex-strongly concave and convex-concave\footnote{\add{For convex-concave problems, we only give intuitions for obtaining lower bounds from corresponding results for strongly convex-strongly concave problems.}} SPPs in both distributed settings: centralized and decentralized. \ans{In particular, for a given budget on the number of communications and on the number of local computations for each node, we provide the lower bound on the resulting accuracy. From such kind of estimates, given the accuracy, we can solve the opposite problem and obtain estimates on the number of communications and local oracle calls.}  

$\bullet$ \textbf{Optimal algorithms. } Next, we get the near-optimal algorithms. They are near-optimal from a theoretical point of view because the upper bounds on \add{their convergence rates} reach lower estimates up to numerical constants and logarithmic factors. For the centralized problem, we construct our method based on the Extra Step method \cite{Korpelevich1976TheEM, juditsky2008solving} (classical and optimal method for non-distributed SPPs) with the correct batch size. In the decentralized case, we also use the Extra Step method as a basis, but communication is done using the accelerated (gossip) consensus procedure \cite{liu2011accelerated}.

\ans{For the summary and comparison of the lower and upper convergence rate bounds, we refer to Table 1.}
\renewcommand{\arraystretch}{1.5}
\renewcommand{\tabcolsep}{10pt} 
\begin{table}[h!]
\begin{center}
\begin{tabular}{ccc}
\hline
\multicolumn{1}{c}{} & \multicolumn{1}{c}{\textbf{lower}} & \multicolumn{1}{c}{\textbf{upper}} \\ \hline
\multicolumn{1}{c}{}   & \multicolumn{2}{c}{{\tt centralized}} \\ \hline
{\tt sc} & {$\Omega\left(R_0^2\exp\left( - \frac{32\mu \min\{K;T\}}{L \Delta} \right) + \frac{\sigma^2}{\mu^2 M T}\right)$} & {$\mathcal{\tilde O}\left(R_0^2 \exp\left( - \frac{\mu \min\{K;T\}}{4L \Delta} \right) + \frac{\sigma^2}{\mu^2 M T}\right)$} \\ \hline
{\tt c} & {$\Omega\left(\frac{L \Omega^2 \Delta}{K} + \frac{\sigma \Omega}{\sqrt{MT}} \right)$} & {$\mathcal{O}\left(\frac{L \Omega_z^2 \Delta}{\min\{K;T\}} + \frac{\sigma \Omega}{\sqrt{MT}} \right)$} \\ \hline
\multicolumn{1}{c}{}   & \multicolumn{2}{c}{{\tt decentralized}} \\ \hline
{\tt sc} & {$\Omega\left(R_0^2 \exp\left( - \frac{128\mu \min\{K;T\}}{L \sqrt{\chi}} \right) + \frac{\sigma^2}{\mu^2 M T}\right)$} & {$\mathcal{\tilde O}\left( R_0^2 \exp\left( - \frac{\mu \min\{K;T\}}{8L \sqrt{\chi}} \right) + \frac{\sigma^2}{\mu^2 M T}\right)$} \\ \hline
{\tt c} & {$\Omega\left(\frac{L \Omega^2 \sqrt{\chi}}{\min\{K;T\}} + \frac{\sigma \Omega}{\sqrt{MT}} \right)$} & {$\mathcal{\tilde O}\left(\frac{L \Omega^2 \sqrt{\chi}}{\min\{K;T\}} + \frac{\sigma \Omega}{\sqrt{MT}} \right)$} \\ \hline
\end{tabular}
\vspace{0.3cm}
\caption{Lower and upper bounds for distributed smooth stochastic strongly convex-strongly concave ({\tt sc}) or convex-concave ({\tt c}) saddle point problems in the centralized and decentralized cases. Notation:
$L$ -- smoothness constant of $f$, $\mu$ -- strong convexity-strong concavity constant, $R_0^2 = \|x^0 - z^* \|_2^2 + \|y^0 - y^* \|_2^2$, $\Omega$ -- diameter of the optimization set, $\Delta$, $\chi$ -- diameter and condition number of the communication graph (condition number of the gossip matrix), $K$ -- number of communication rounds, $T$ -- number of local calls of the gradient oracle on each node. In the convex-concave case, the bounds are in terms of the gap function \add{(see \eqref{gap})}, in the strongly convex-strongly concave case -- in terms of the (squared) distance to the solution.}
\end{center}
\label{table1}
\end{table}

$\bullet$ \textbf{Local method. } We also present an extra-step modification of Local SGD \cite{mcdonald-etal-2010-distributed, stich2018local},
one of the most popular methods in federated learning \cite{konevcny2016federated, kairouz2019advances}. More recently, other versions of the Local SGD methods for SPPs have appeared \cite{deng2021local,hou2021efficient}. All of the methods presented in these papers are based on gradient descent-ascent, but it is known that such methods, even in the non-distributed case, diverge for the most common SPPs \cite{goodfellow2016nips, daskalakis2017training}. Our method is based on the classic method for smooth SPPs -- Extra Step algorithm, which makes it stand out from the competitors.

$\bullet$ \textbf{Non-convex-non-concave analysis. } We analyze our new algorithms: near-optimal and local, not only in the convex-concave case, but even in the non-convex-non-concave case under the Minty assumption \cite{minty62,  diakonikolas2021efficient}. Minty is the weakest additional assumption for a non-convex-non-concave problem found in the literature. \ans{Under this weak assumption only a few results on distributed methods for SPPs are available in the literature \cite{liu2019decentralizedprox, liu2019decentralized}.} In particular, our analysis covers the estimates of the decentralized but deterministic method from \cite{liu2019decentralizedprox}, and also generalizes and overlaps the estimates for the stochastic method for homogeneous data ($f_m = f$) from \cite{liu2019decentralized}.

$\bullet$ \textbf{Experiments. } The first part of our experiments on the classical bilinear problem is devoted to the comparison of the optimal centralized method and the method based on Local SGD, as well as the comparison of our local method with  competitors \cite{deng2021local,hou2021efficient}. 
The second part is devoted to the use of Local SGD  and Local Adam techniques for training GANs in a homogeneous and heterogeneous cases.


\subsection{Related works}

\hspace{0.45cm} $\bullet$ \textbf{SPPs.} First, we highlight two main non-distributed algorithms for SPPs. The first algorithm -- Mirror Descent \cite{nemirovski}, it is customary is usually used in the non-smooth case. For smooth problems, Extra Step/Mirror Prox is applied \cite{Korpelevich1976TheEM, Nemirovski2004, juditsky2008solving}. Also, the following methods \cite{nesterov2007dual, hsieh2019convergence, doi:10.1137/S0363012998338806} can be noted as popular for smooth SPPs.

$\bullet$ \textbf{Lower bounds.} In the non-distributed case, the lower bounds for smooth strongly convex-strongly concave case SPPs are given in \cite{zhang2019lower}, for convex-concave -- in \cite{ouyang2019lower}. In smooth stochastic convex optimization, we highlight works on lower bounds \cite{nemirovskij1983problem, foster2019complexity}. It is also important to note the works devoted to the lower bounds for centralized and decentralized distributed convex optimization \cite{scaman2017optimal,arjevani2015communication}.

$\bullet$ \textbf{Distributed SPPs.} The following works are devoted to decentralized SPPs: in the deterministic case \cite{liu2019decentralizedprox, 9304470, rogozin2021decentralized}, in the stochastic case \cite{liu2019decentralized}. Let us also highlight the local methods for SPPs \cite{deng2021local,hou2021efficient} already noted earlier in Section \ref{sec:contr}.


\section{Settings and assumptions}\label{sec:notation}

We consider the problem \eqref{distr}, where the sets $\mathcal{X} \subseteq \mathbb{R}^{n_x}$ and $\mathcal{Y} \subseteq \mathbb{R}^{n_y}$ are \ans{closed} convex sets. For simplicity, we introduce the set $\mathcal{Z} = \mathcal{X} \times \mathcal{Y}$, $z = (x,y)$ and the operators $F$ and $F_m$:
\begin{equation}
\label{opSP}
\ans{F(z) := F(x,y) := \begin{pmatrix}
\nabla_x f(x,y)\\
-\nabla_y f(x,y)
\end{pmatrix},}
\quad
F_m(z) := F_m(x,y) := \begin{pmatrix}
\nabla_x f_m(x,y)\\
-\nabla_y f_m(x,y)
\end{pmatrix}.
\end{equation}
\ans{As noted above, we consider stochastic formulations of the problem \eqref{distr}, where each $f_m$ is of the form where $f_m(x,y) = \mathbb{E}_{\xi \sim \mathcal{D}_m} [f_m(x,y,\xi)]$ with an unknown distribution $\mathcal{D}_m$. Similar to \eqref{opSP}, one can also introduce $F_m(z, \xi)$. We assume that we do not have access to oracles of $f_m(x,y)$ and $F_m(z)$, but can only call $f_m(x,y,\xi)$ and $F_m(z,\xi)$, where $\xi$ is some sample from the distribution $\mathcal{D}_m$.}

Next, we introduce the following assumptions:

\begin{customass}{1(g)} \label{ass:as1g}
$f(x,y)$ is $L$ - smooth, i.e. for all $z_1, z_2 \in \mathcal{Z}$
    \begin{align}
    \label{as1g}
    \|F(z_1) - F(z_2)\| \leq L\|z_1-z_2\|.
    \end{align}
\end{customass}
\begin{customass}{1(l)} \label{as:as1l}
For all $m$, $f_m(x,y)$ is $L_{\max}$-smooth, i.e. for all $z_1, z_2 \in \mathcal{Z}$
    \begin{align}
    \label{as1l}
    \|F_m(z_1) - F_m(z_2)\| \leq L_{\max}\|z_1-z_2\|.
    \end{align}
\end{customass}
\begin{customass}{2(sc)} \label{ass:as2g}
$f(x,y)$ is strongly convex-strongly concave with constant $\mu$, i.e. for all $z_1, z_2 \in \mathcal{Z}$
    \begin{align} 
    \label{as2g}
    \langle F(z_1) - F(z_2), z_1 - z_2 \rangle \geq \mu\|z_1-z_2\|^2.
    \end{align}
\end{customass}
\begin{customass}{2(c)} \label{ass:as2c}
$f(x,y)$ is convex-concave, i.e. $f(x,y)$ is strongly convex-strongly concave with $\mu=0$.
\end{customass}
\begin{customass}{2(nc)} \label{ass:as2n}
$f$ satisfies the Minty assumption, i.e. there exists $z^* \in \mathcal{Z}$ such that for all $z \in \mathcal{Z}$
\begin{align}
    \label{as2m}
    \langle F(z), z - z^* \rangle\geq 0. 
\end{align}
\end{customass}
\begin{customass}{3} \label{as:as3}
$F_m (z, \xi)$ is unbiased and has bounded variance, i.e. for all $z \in \mathcal{Z}$ it holds that
    \begin{align} 
    \label{as3}
    \E_{\xi \sim \mathcal{D}_m}[F_m(z,\xi)] = F_m(z), \quad \E_{\xi \sim \mathcal{D}_m}[\|F_m(z,\xi) - F_m(z)\|^2] \leq  \sigma^2.
    \end{align}
\end{customass}
\begin{customass}{4} \label{as:as4}
$\mathcal{Z}$ is compact, i.e. for all $z, z'\in \mathcal{Z}$
    \begin{align} \label{as5}
    \| z - z'\| \leq \Omega_z.
    \end{align}
\end{customass}

Hereinafter, we use the standard Euclidean norm $\|\cdot\|$. We also introduce the following notation $\text{proj}_{\mathcal{Z}}(z) = \min_{u \in \mathcal{Z}}\| u - z\|$ -- the Euclidean projection onto $\mathcal{Z}$.


We also assume that all devices are connected to each other in a network, which can be represented as an undirected graph $\mathcal{G}(\mathcal{V}, \mathcal{E})$, where $\mathcal{V}$ is a set of vertices and $\mathcal{E}$ is a set of edges. \ans{We introduce the graph diameter $\Delta$ as the maximum distance between the pair of vertices.}
As mentioned earlier, we are interested in several cases of distributed optimization: centralized, and decentralized. It is important to mention one of the most popular communication procedures in the decentralized setup --
\textit{the gossip protocol} \cite{kempe2003gossip, boyd2006randomized, nedic2009distributed}. This approach uses a particular matrix $W$. \ans{The devices pass their local variables to their neighbors along the edges of the graph, and each node computes a new value of the local variable by averaging its own variable and the information received from its neighbors according to the weights in the matrix $W$.} Therefore, the convergence of decentralized algorithms is determined by the properties of this matrix. Therefore, we introduce its definition:

\begin{definition} \label{def:goss} We call a $M \times M$ matrix $W$ a gossip matrix if it satisfies the following conditions: 1) $W$ is symmetric \ans{and} positive semi-definite, 2) the kernel of $W$ is the set of constant vectors \ans{(vectors with equal components)}: $\text{ker}(W) = \text{span}(\mathbf{1})$, 3) $W$ is defined on the edges of the network: $W_{ij} \neq 0$ \ans{if and} only if $i=j$ or $(i,j) \in \mathcal{E}$.
\end{definition}

Let $\lambda_1(W) \geq \ldots \geq \lambda_M(W) = 0$ \ans{be the spectrum of $W$, and the condition number defined as $\chi = \chi(W) = \frac{\lambda_1(W)}{\lambda_{M-1}(W)}$.} \ans{Note that in many works \cite{boyd2006randomized, scaman2017optimal, liu2011accelerated, ye2020multi}, the authors do not use the matrix $W$, but $\tilde W = I - \frac{W}{\lambda_1(W)}$.}
To describe the convergence, we introduce $\lambda_2(\tilde W) = 1 - \frac{\lambda_{M-1}(W)}{\lambda_1(W)} = 1 - \frac{1}{\chi(W)} = 1 - \frac{1}{\chi}$.

The next definition is necessary to describe a certain class of distributed algorithms, for which we will obtain lower bounds. We use a definition quite similar to \cite{arjevani2015communication, scaman2017optimal}. 

\begin{definition}\label{app:proc} Let \ans{us} introduce a procedure with two parameters $T$ and $K$, which we call {\tt Black-Box Procedure}$(T,K)$ \ans{or {\tt BBP}$(T,K)$}. Each worker $m$ has its own local memories $\mathcal{M}^x_{m}$ and $\mathcal{M}^y_{m}$ for the $x$- and $y$-variables, respectively--with  initialization $\mathcal{M}_{m}^x = \mathcal{M}_{m}^y= \{0\}$.  $\mathcal{M}_{m}^x$ and $\mathcal{M}_{m}^x$ are updated as follows.
   
    $\bullet$ \textbf{Local computation:} \ans{By each local computation} the $m$th device can independently sample a random variable $\xi_m$ \ans{from the distribution $\mathcal{D}_m$} and adds to its $\mathcal{M}^x_{m}$ and $\mathcal{M}^y_{m}$ a finite number of points $x,y$, satisfying 
    \begin{equation}\begin{aligned}\label{app:oracle-opt-step}
        x \in \text{span} \big\{x',\nabla_x f_{m}(x'',y'', \xi_m) \big\}, ~~~
        y \in \text{span} \big\{y',\nabla_y f_{m}(x'',y'', \xi_m)\big\}
    \end{aligned}\end{equation}
    for given $x', x'' \in \mathcal{M}^x_{m}$ and  $y', y'' \in \mathcal{M}^y_{m}$. \ans{We also assume that devices make projections for free, i.e., $m$th can add to $\mathcal{M}^x_{m}$ and $\mathcal{M}^y_{m}$ a finite number of points $x,y$, satisfying }
    \begin{equation}\begin{aligned}\label{app:oracle-proj}
        x \in \text{span} \big\{\text{proj}_{\X}(x') \big\}, ~~~
        y \in \text{span} \big\{\text{proj}_{\Y}(y')\big\}~~~ \text{for}~~ x'\in \mathcal{M}^x_{m} ~~\text{and}~~  y'\in \mathcal{M}^y_{m}.
    \end{aligned}\end{equation}
    
    $\bullet$ \textbf{Communication:} Based upon communication rounds  among neighbouring nodes, $\mathcal{M}^x_{m}$ and $\mathcal{M}^y_{m}$ are updated according to
    \begin{equation}\label{app:oracle-comm}
        \mathcal{M}^x_{m} := \text{span}\left\{\bigcup_{(i,m) \in \mathcal{E}} \mathcal{M}^x_{i} \right\}, \quad 
        \mathcal{M}^{y}_{m} := \text{span}\left\{\bigcup_{(i,m) \in \mathcal{E}} \mathcal{M}^y_{i} \right\}.
    \end{equation}

    $\bullet$ \textbf{Output:} 
    The final global output is calculated as: 
    \begin{align*}
        \hat x \in \text{span}\left\{\bigcup_{m=1}^M \mathcal{M}^x_{m} \right\},~~\hat y \in \text{span}\left\{\bigcup_{m=1}^M \mathcal{M}^y_{m} \right\}.
    \end{align*}
    We assume that each node makes no more than $T$ local iterations (for simplicity, that exactly $T$) during the operation of the algorithm. The number of communication rounds is also limited to a certain number of $K < T$.
\end{definition}

\section{Lower bounds}\label{sec:lb}

Following the classical results on obtaining lower bounds, it is sufficient to give an example of a <<bad>> function \cite{nesterov2013introductory}, and the <<bad>> partitioning of this function between nodes \cite{scaman2017optimal}.
We start with the <<bad>> function. \ans{The first important point in constructing of our <<bad>> function is that it consists of two independent parts: deterministic and stochastic.} Consider $f_m(x,y) = f^{deter}_m(x^{deter},y) + f^{stoch}(x^{stoch}),$
where the vectors $x^{deter}$ and $x^{stoch}$ together give the vector $x = \ans{(x^{deter}, x^{stoch})}$. At the same time we have access to $F_m(x,y,\xi) = F^{deter}_m(x^{deter},y) + \nabla f^{stoch}(x^{stoch}, \xi)$. This means that we have a deterministic oracle for $f^{deter}_m$ and a stochastic one -- for $f^{stoch}$. Such $f_m$ helps to rewrite the original problem \eqref{distr}  as follows:
\begin{equation}
    \label{m1}
    \min_{x^{deter} \in \mathcal{X}^{deter}} \max_{y \in \Y} \frac{1}{M} \sum\limits_{m=1}^M f^{deter}_m(x^{deter},y) + \min_{x^{stoch} \in \mathcal{X}^{stoch}} f^{stoch}(x^{stoch}).
\end{equation}
Therefore, we separately prove the estimates for each of the problems, and then combine. 


\subsection{Deterministic lower bounds}

In this part, we provide lower bounds for the centralized (Theorem \ref{th1}) and decentralized (Theorem \ref{th2}) cases.

\begin{theorem} \label{th1}
For any $L > \mu >0$ and any $\Delta \in \mathbb{N}$, there exists a distributed saddle point problem satisfying Assumptions \ref{ass:as1g} and \ref{ass:as2g} on $\mathcal{X} \times \mathcal{Y} = \mathbb{R}^n \times \mathbb{R}^n$ (where $n$ is sufficiently large) with $x^*, y^* \neq 0$ \ans{over a fixed network with a diameter $\Delta$}, such that for any output $\hat x, \hat y$ of any procedure satisfying Definition \ref{app:proc}, the following estimate hold:
\begin{equation*}
   \|\hat x - x^*\|^2 + \|\hat y - y^*\|^2 = \Omega\left(\exp\left( - \frac{4\mu}{L - \mu} \cdot \frac{K}{\Delta} \right) \| y^0 - y^*\|^2\right).
\end{equation*}
\end{theorem}


\begin{theorem} \label{th2}
For any $L > \mu >0$ and any $\chi \geq 1$, there exists a distributed saddle point problem satisfying Assumptions \ref{ass:as1g} and \ref{ass:as2g} on $\mathcal{X} \times \mathcal{Y} = \mathbb{R}^n \times \mathbb{R}^n$ (where $n$ is sufficiently large) with $x^*, y^* \neq 0$ \ans{over a fixed network characterized by a gossip matrix (Definition \ref{def:goss}) with a condition number $\chi$}, such that for any output $\hat x, \hat y$ of any procedure satisfying Definition \ref{app:proc},  the following estimate hold:
\begin{equation*}
    \|\hat x - x^*\|^2 + \|\hat y - y^*\|^2 = \Omega\left(\exp\left( - \frac{32\mu}{L - \mu} \cdot \frac{K}{\sqrt{\chi}} \right) \| y^0 - y^*\|^2\right).
\end{equation*}
\end{theorem}

\textbf{Convex-concave case.}
Note that in the convex-concave case the problem is usually considered on a bounded set (see Assumption \ref{as:as4}). \ans{Moreover, the convergence criterion
for algorithms is formulated in terms of the following gap function:}
\begin{equation}
    \label{gap}
    \ans{\mathrm{gap}(z) := } \mathrm{gap}(x,y) := \max_{y' \in \mathcal{Y}} f(x, y') - \min_{x' \in \mathcal{X}} f(x', y).
\end{equation}
Therefore, the lower bounds are also needed in terms of \eqref{gap}. Following the inequality 6 of \cite{zhang2019lower}, we can rewrite the estimates from Theorems \ref{th1} and \ref{th2} as follows
\begin{equation}
    \label{eq:temppppp}
    \mathrm{gap}(x,y) \geq \frac{\mu}{2} \|x - x^* \|^2 + \frac{\mu}{2} \|y - y^* \|^2.
\end{equation}
In Table 1, the lower bounds for the convex-concave case are already diclared. Let us give an intuition how these estimates can be obtained from Theorems \ref{th1} and \ref{th2}. 

\add{
One way to get the lower estimates is to prove by contradiction and assume that the estimate given in Table 1 is not valid: there exists an algorithm that converges better than the given bound, namely for some numerical constant $C$ the output of this method after $k$ communications satisfies
\begin{equation}
    \label{eq:tempppp}
    \mathrm{gap}(\hat x^k, \hat y^k) < \frac{C L \Delta (\| x^0 - x^* \|^2 + \| y^0 - y^* \|^2)}{k}.
\end{equation}
Using \eqref{eq:temppppp}, we have
$$
\|\hat x^k - x^* \|^2 + \|\hat y^k - y^* \|^2 < \frac{2 C L \Delta (\| x^0 - x^* \|^2 + \| y^0 - y^* \|^2)}{\mu k}.
$$
It follows that for some $k < \frac{2 C L \Delta}{\mu}$ one can guarantee that
$
(\|\hat x^k - x^* \|^2 + \|\hat y^k - y^* \|^2) < (\| x^0 - x^* \|^2 + \| y^0 - y^* \|^2) / 2.
$
Using the restarts idea, specifically running the algorithm for $k$ communications several times from the output of the previous run, after
\begin{align*}
K 
&= k \cdot \log_2 \frac{\| x^0 - x^*\|^2 + \| y^0 - y^*\|^2}{\varepsilon} 
\\
&<  \frac{2 C L \Delta}{\mu} \cdot \log_2 \frac{\| x^0 - x^*\|^2 + \| y^0 - y^*\|^2}{\varepsilon} \quad \text{communications}
\end{align*}
one can find some $\varepsilon$-solution of the strongly convex-strongly concave problem. But this contradicts the results of Theorem \ref{th1}, which states that 
$$
K = \Omega\left( \frac{L\Delta}{\mu} \log_2 \frac{\| y^0 - y^*\|^2}{\varepsilon} \right).
$$
Similar considerations can be made for Theorem \ref{th2}. But this reasoning has a drawback due to the fact that in the lower bounds from Table 1 (and further in the upper bounds from Section \ref{sec:oa}), $\Omega_z$ appears, while in \eqref{eq:tempppp} we used $(\| x^0 - x^* \|^2 + \| y^0 - y^* \|^2)$. Therefore, this reasoning is not completely correct.

Another way to obtain the result is to consider the same line as in the proofs of Theorems \ref{th1} and \ref{th2}, but replace $\mu$ with $\frac{\varepsilon}{\Omega^2_z}$. For any $\mu > 0$, a strongly convex-strongly concave function is also convex-concave. Therefore, this substitution of $\mu$ can serve as an example of a convex-concave function.
}



\subsection{Stochastic lower bounds}

\ans{Due to our choice of $f_m$ from \eqref{m1}, one can note that in order to obtain lower stochastic bounds, we consider the minimization problem rather than the SPP, and moreover, this problem is the same for each node. Therefore, it is suggested that we are in an ideal situation when each node communicates to each and collectively minimizes the same function. Then the total number of stochastic oracle calls is $MT$. It may seem that such a formulation simplifies the problem and may not yield the most advanced lower bounds. But in the next section, we will give upper bounds that coincide with the lower bounds, and this will verify that considering such an idealistic setting is sufficient.}

Let us formulate two theorems for the convex and strongly convex cases of $f^{\text{stoch}}$.

\begin{theorem} \label{th3}
For any $L > \mu >0$, there exists a stochastic minimization problem with $L$-smooth and  $\mu$-strongly convex function (i.e., satisfying Assumptions \ref{ass:as1g} and \ref{ass:as2g}),  such that for any output $\hat x$ of any {\tt BBP}$(T,K)$ (Definition \ref{app:proc}) with $M$ workers, one can obtain the following estimate:
\begin{equation*}
    \add{\mathbb{E}\left[\|\hat x - x^*\|^2\right]} =  \Omega\left(\frac{\sigma^2}{MT\mu^2}\right).
\end{equation*}
 \end{theorem}
 
\begin{theorem} \label{th4}
For any $L >0$ and any  $M,T \in \mathbb{N}$, there exists a stochastic minimization problem with $L$-smooth and convex function (i.e., satisfying Assumptions \ref{ass:as1g} and \ref{ass:as2c}) on a bounded $\mathcal{X}^{stoch}$ \ans{with a diameter $\Omega_z$ (i.e., satisfying Assumption \ref{as:as4})}, such that for any output $\hat x$ of any {\tt BBP}$(T,K)$ (Definition \ref{app:proc}) with $M$ workers, one can obtain the following estimate:
\begin{equation*}
\EE\left[f^{\text{stoch}}(\hat x)  - f^{\text{stoch}}(x^*)\right] =  \Omega\left(\frac{\sigma \Omega_z}{\sqrt{MT}}\right).
\end{equation*}
 \end{theorem}

\subsection{Connection of lower bounds}

The connecting of deterministic and stochastic bounds follows from \eqref{m1}. The results for the centralized and decentralized cases are shown in Table 1. 
See Appendix \ref{ap:lb} for complete proof of this part. To verify the tightness of our lower bounds, the next section designs algorithms that reach such bounds.

\section{Optimal algorithms}\label{sec:oa}

This section focuses on theoretically near-optimal algorithms. It is easy to check that our algorithms satisfy the {\tt BBP} definition. 

\subsection{Centralized case}

We design our algorithm based on MiniBatch SGD and Extra Step. For this algorithm we introduce $r$ as a maximum distance from nodes to server. It is easy to note that $r$ is upper bounded by the diameter $\Delta$.
\begin{algorithm} [th]
	\caption{Centralized Extra Step Method}
	\label{alg1}
	\begin{algorithmic}
\STATE
\noindent {\bf Parameters:}  Stepsize $\gamma \leq \frac{1}{4L}$; communication rounds $K$, number of local steps $T$.\\
\noindent {\bf Initialization:} Choose  $(x^0,y^0)=z^0\in \mathcal{Z}$, $k = \left\lfloor \frac{K}{r} \right\rfloor$ and batch size $b = \left\lfloor\frac{T}{2k}\right\rfloor$.
\FOR {$t=0,1, 2, \ldots, k-1$ }
\STATE Generate batch $\xi^t_m$ on each machine independently
\STATE Each machine $m$  computes  $g^t_m = \frac{1}{b}\sum\limits_{i=1}^b F_m(z^t, \xi^{t,i}_m)$   ~and sends  $g^t_m$ to server
 \STATE Server  computes  $z^{t+1/2} = \text{proj}_{\mathcal{Z}}(z^t -  \frac{\gamma}{M}\sum\limits_{m=1}^M g^t_m)$ ~and then sends $z^{t+1/2}$ to machines
 \STATE Generate batch $\xi^{t+1/2}_m$ on each machine independently
 \STATE Each machine $m$  computes  $g^{t+1/2}_m = \frac{1}{b}\sum\limits_{i=1}^b F_m(z^{t+1/2}, \xi^{t+1/2,i}_m)$ ~and sends  $g^{t+1/2}_m$ to server
 \STATE Server  computes $z^{t+1} = \text{proj}_{\mathcal{Z}}(z^t -  \frac{\gamma}{M}\sum\limits_{m=1}^M g^{t+1/2}_m)$ ~and then sends $z^{t+1}$ to machines
\ENDFOR
	\end{algorithmic}
\end{algorithm}

\begin{theorem}\label{th5}
Let $\{ z^t\}_{t \geq 0}$ denote the iterates of Algorithm~\ref{alg1} for solving the problem \eqref{distr}. Let Assumptions \ref{ass:as1g}, \ref{as:as3} be satisfied. Then if $\gamma \leq \frac{1}{4L}$, we have the following estimates in

$\bullet$ the $\mu$-strongly convex-strongly concave case (Assumption \ref{ass:as2g}):
\begin{equation*}
 \E[\| z^{k} - z^* \|^2] = \mathcal{\tilde O}\left(\| z^{0} - z^* \|^2 \exp\left( -\frac{\mu K}{4L\Delta}\right)+\frac{\sigma^2}{\mu^2 M T}\right),
 \end{equation*}
 
$\bullet$ the convex-concave case (Assumptions \ref{ass:as2c} and \ref{as:as4}):
\begin{equation*}
    \E[\mathrm{gap}(z^{k}_{avg})] = \mathcal{O}\left(\frac{L \Omega_z^2 \Delta}{K} + \frac{\sigma \Omega_z}{\sqrt{MT}} \right),
\end{equation*}

$\bullet$ the non-convex-non-concave case (Assumptions \ref{ass:as2n} and \ref{as:as4}):
\begin{equation*}
    \E\left[ \frac{1}{k}\sum\limits_{t=0}^{k-1}\|F(z^t)\|^2\right] = \mathcal{O}\left(\frac{L^2 \Omega_z^2 \Delta}{K} + \frac{\sigma^2 K}{MT\Delta} \right),
\end{equation*}
where $z^{k}_{avg} = \frac{1}{k} \sum\limits_{t=0}^{k-1} z^{t+1/2}$.
\end{theorem}

\subsection{Decentralized case} 

The idea of Algorithm \ref{alg2} combines three things: Extra Step,
accelerated consensus -- {\tt FastMix} (see Algorithm \ref{alg3} in Appendix \ref{aoa} or \cite{liu2011accelerated,ye2020multi}) and the right size of batches. 
\begin{algorithm} [th]
	\caption{Decentralized Extra Step Method}
	\label{alg2}
	\begin{algorithmic}
\STATE
\noindent {\bf Parameters:}  Stepsize $\gamma \leq \frac{1}{4L}$; communication rounds $K$, number of local calls $T$, number of {\tt FastMix} steps $P$.\\
\noindent {\bf Initialization:} Choose  $(x^0,y^0)=z^0\in \mathcal{Z}$, $z^0_m = z^0$, $k = \left\lfloor \frac{K}{P} \right\rfloor$ and batch size $b = \left\lfloor\frac{T}{2k}\right\rfloor$.
\FOR {$t=0,1, 2, \ldots, k-1$ }
\STATE Generate batch $\xi^t_m$ on each machine independently
 \STATE Each machine $m$ computes   $\hat z_m^{t+1/2} = z_m^{t} - \gamma \cdot \frac{1}{b}\sum\limits_{i=1}^b F_m(z^t_m, \xi^{t,i}_m)$
 \STATE  Communication:  $\tilde z^{t+1/2}_1, \ldots,  \tilde z^{t+1/2}_M$ ={\tt FastMix}$(\hat z^{t+1/2}_1, \ldots, \hat z^{t+1/2}_M, P)$
\STATE Each machine $m$ computes $z^{t+1/2}_m = \text{proj}_{\mathcal{Z}}(\tilde z^{t+1/2}_m)$
\STATE Generate batch $\xi^{t+1/2}_m$ on each machine independently
 \STATE Each machine $m$ compute  $\hat z_m^{t+1} = z_m^{t} - \gamma \cdot \frac{1}{b}\sum\limits_{i=1}^b F_m(z^{t+1/2}_m, \xi^{t+1/2,i}_m)$
 \STATE Communication:  $\tilde z^{t+1}_1, \ldots,  \tilde z^{t+1}_M$ ={\tt FastMix}$(\hat z^{t+1}_1, \ldots, \hat z^{t+1}_M, P)$
 \STATE Each machine $m$ compute  $z^{t+1}_m = \text{proj}_{\mathcal{Z}}(\tilde z^{t+1}_m)$
\ENDFOR
	\end{algorithmic}
\end{algorithm}

\begin{theorem} \label{th6}
Let $\{ z_m^t\}_{t \geq 0}$ denote the iterates of Algorithm~\ref{alg2} for solving the problem \eqref{distr}. Let Assumptions \ref{ass:as1g}, \ref{as:as1l}, \ref{as:as3} be satisfied. Then if $\gamma \leq \frac{1}{4L}$ and $P = \mathcal{O}\left(\sqrt{\chi} \log \frac{1}{\varepsilon}\right)$, we have the following estimates in

$\bullet$ the $\mu$-strongly convex-strongly concave case (Assumption \ref{ass:as2g}):
\begin{eqnarray*}
 \E[\| \bar z^{k} -z^* \|^2] = \mathcal{\tilde O}\left( \| z^{0} - z^* \|^2 \exp\left( -\frac{\mu K}{8L\sqrt{\chi}} \right) + \frac{\sigma^2}{\mu^2 M T} \right) ,
 \end{eqnarray*}

$\bullet$ the convex-concave case (Assumptions \ref{ass:as2c} and \ref{as:as4}):
\begin{equation*}
    \E[\mathrm{gap}
    (\bar z^{k}_{avg})] = \mathcal{\tilde O}\left(\frac{L \Omega_z^2 \sqrt{\chi}}{K} + \frac{\sigma \Omega_z}{\sqrt{MT}} \right),
\end{equation*}

$\bullet$ the non-convex-non-concave case (Assumptions \ref{ass:as2n} and \ref{as:as4}):
\begin{equation*}
    \E\left[ \frac{1}{k}\sum\limits_{t=0}^{k-1}\|F(\bar z^t)\|^2\right] = \mathcal{\tilde O}\left(\frac{L^2 \Omega_z^2 \sqrt{\chi}}{K} + \frac{\sigma^2 K}{MT\sqrt{\chi} } \right),
\end{equation*}
where $\bar z^{t} = \frac{1}{M} \sum\limits_{m=1}^M z_m^{t}$ and $\bar z^{k+1}_{avg} = \frac{1}{Mk} \sum\limits_{t=0}^{k-1} \sum\limits_{m=1}^M z_m^{t+1/2}$.
\end{theorem}

\textbf{Discussions. } Let us make some comments on our Algorithms:

$\bullet$ It is easy to see that our Algorithms are near-optimal \ans{in the strongly convex-strongly concave and convex-concave cases} -- see Table 1 for details. However, there are several practical drawbacks of these Algorithms. The first is related to the fact that in Algorithm \ref{alg2} we have to take multi consensus steps at each iteration. This approach does not always pay off in practice. On the other hand, the optimal decentralized algorithms for minimization problems also use {\tt FastMix} -- see literature review in
\cite{song2021optimal}. Second, if $T \gg K$, we collect a very large batch at each iteration, in practice such batches do not make sense. Therefore, the idea arises to use these local computations of gradients more efficiently, e.g. by doing local steps. This brings us to Section \ref{sec:na}.

$\bullet$ One can note that in the non-convex-non-concave case, we do not guarantee the convergence if $T \approx K$. However, the method converges sublinearly when $\sigma = 0$. In this case, we cover the deterministic results of \cite{liu2019decentralizedprox}. In the stochastic case ($\sigma > 0$), convergence is also not guaranteed in \cite{liu2019decentralized,barazandeh2021solving}. Therefore, we cover and even overlap their analysis, since they consider only the homogeneous case ($f_m = f$).

\section{New local algorithm}\label{sec:na}

In this section, we work on sets $\mathcal{X} = \mathbb{R}^{n_x}$ and $\mathcal{Y} = \mathbb{R}^{n_y}$. Additionally, we introduce the following assumption:

\begin{customass}{5} \label{as:as5}
The values of the local operator are considered sufficiently close to the value of the mean operator, i.e. for all $z \in \mathcal{Z}$
    \begin{equation} 
    \label{as4}\|F_m(z) - F(z) \| \leq D.
    \end{equation}
\end{customass}

This assumption is often called $D$ - heterogeneity.

Our algorithm is a combination of Local SGD and  Extra Step. One can note that such an algorithm is {\tt BBP}$(T,K)$.
\begin{algorithm} [th]
	\caption{Extra Step Local SGD}
	\label{alg4}
	\begin{algorithmic}
\STATE
\noindent {\bf Parameters:}  stepsize $\gamma \leq \frac{1}{21HL_{\max}}$; number of local steps $T$, sets $I$ of communications steps ($|I| = K$).\\
\noindent {\bf Initialization:} Choose  $(x^0,y^0)=z^0\in \mathcal{Z}$, for all $m$, $z^0_m = z^0$ and $\hat z = z^0$.
\FOR {$t=0,1, 2, \ldots, T-1$ }
\STATE Generate random variable $\xi^t_m$ on each machine independently
 \STATE Each machine $m$ computes  \hspace{0.1cm} $z^{t+1/2}_m = z^t_m - \gamma F_m(z^t_m, \xi^t_m)$
 \STATE Generate random variable $\xi^{t+1/2}_m$ on each machine independently
 \STATE Each machine $m$ computes  \hspace{0.1cm} $z^{t+1}_m = z^t_m - \gamma F_m(z^{t+1/2}_m, \xi^{t+1/2}_m)$
 \STATE \textbf{if} $t \in I$  \textbf{do}
 \STATE \hspace{0.4cm}  Each machine sends $z^{t+1}_m$ to server
\STATE \hspace{0.4cm} Server computes $\hat z = \frac{1}{M}\sum\limits_{m=1}^M z^{t+1}_m$,~ sends $\hat z$ to machines
 \STATE \hspace{0.4cm}  Each machine gets $\hat z$ and sets $z^{t+1}_m = \hat z$ 
\ENDFOR
\STATE 
\noindent {\bf Output:} $\hat z$.
	\end{algorithmic}
\end{algorithm}


\begin{theorem} \label{th7}
Let $\{ z^t_m\}_{t \geq 0}$ denote the iterates of Algorithm~\ref{alg4} for solving the problem \eqref{distr}. Let Assumptions \ref{as:as1l}, \ref{as:as3} and \ref{as:as5} be satisfied. Also let $H = \max_p |k_{p+1} - k_p|$ be a maximum distance between moments of communication ($k_p \in I$). Then we have the following estimates in 

$\bullet$ the $\mu$-strongly convex-strongly concave case (Assumption \ref{ass:as2g}) with $\gamma \leq \frac{1}{21HL_{\max}}$:
\begin{equation*}
 \E[\|\bar z^{T} - z^* \|^2 ] = \mathcal{\tilde O}\left(\|z^0 - z^* \|^2 \cdot \exp\left( - \frac{\mu T}{42 H L_{\max}}\right) + \frac{\sigma^2}{\mu^2 M T} + \frac{L_{\max}^2 H}{\mu^4   T^2}  \left( H D^2 + \sigma^2 \right)\right),
 \end{equation*}
 
$\bullet$ the non-convex-non-concave case (Assumption \ref{ass:as2n} and with assumption that for all $t$, $\|\bar z^t\| \leq \Omega$) with $\gamma \leq \frac{1}{4L_{\max}}$:
\begin{align*}
\E\left[ \frac{1}{T}\sum\limits_{t=0}^{T-1}\|F(\bar z^t)\|^2\right] &= \mathcal{O} \Bigg(\frac{L_{\max}^2 \Omega^2}{T} + \frac{\left[H L_{\max} \Omega \left( HD^2  + \sigma^2 \right) \right]^{2/3}}{T^{1/3}} \\ &\hspace{1cm}+\frac{\sigma^2}{M} + L_{\max} \Omega \sqrt{ H \left( HD^2  + \sigma^2 \right) } \Bigg),
\end{align*}
where $\bar z^{t} = \frac{1}{M} \sum\limits_{m=1}^M z_m^{t}$.
\end{theorem}


\textbf{Discussions. } Let us add some remarks about obtained results:

$\bullet$ Compared to Algorithm \ref{alg1}, Algorithm \ref{alg4} gives worse convergence guarantees. Why then Algorithm \ref{alg4} is needed? For practical reasons. Local SGD or FedAvg is a fairly well-known and popular federated learning concept. We extend this concept to min-max problems, including non-convex-non-concave ones.
In particular, the theory states that for Algorithm \ref{alg1} step $\gamma \sim \frac{1}{L_{\max}}$, and for Algorithm \ref{alg1} $\gamma \sim \frac{1}{HL_{\max}}$, but in practice one can use the same steps (learning rates) for both Algorithms. It seems natural that Algorithm \ref{alg4} can outperform Algorithm \ref{alg1} in some regimes, simply because it takes more steps (see Section \ref{sec:e}).

$\bullet$ As noted in Section \ref{sec:contr}, there are two other methods of the Local SGD type for SPPs \cite{deng2021local,hou2021efficient}. However, these methods use Descent-Ascent instead of Extra Step as a base. 
Also, the stepsize of these methods is confusing, even in the strongly convex-strongly concave case, it is proposed to take $\gamma = \frac{\mu}{ H L_{\max}^2}$, which in practice is a very small number and provides a very slow convergence of the methods.

\section{Experiments}\label{sec:e}

\subsection{Bilinear problem} \label{sec:be}

Let us start with an experiment on the bilinear problem:
\begin{equation}
    \label{bilin}
    \minmax_{{\tiny x,y\in [-1;1]^n}}\frac{1}{M} \sum\limits_{m=1}^M \left(x^T A_m y + b^T_m x + c_m^T y\right),
\end{equation}
where $n = 100$, $M = 100$, 
matrices $A_m\succ 0$ are randomly generated with $\lambda_{\max} = 1000$ (then $L = 1000$). Coordinates $b_m, c_m$ are generated uniformly on $[-1000;1000]$. Moreover, we add unbiased \ans{Gaussian} noise with $\sigma^2 = 10000$ to the gradients. Starting point is zero. 

\begin{figure}[h]
\begin{minipage}{0.33\textwidth}
  \centering
\includegraphics[width =  \textwidth ]{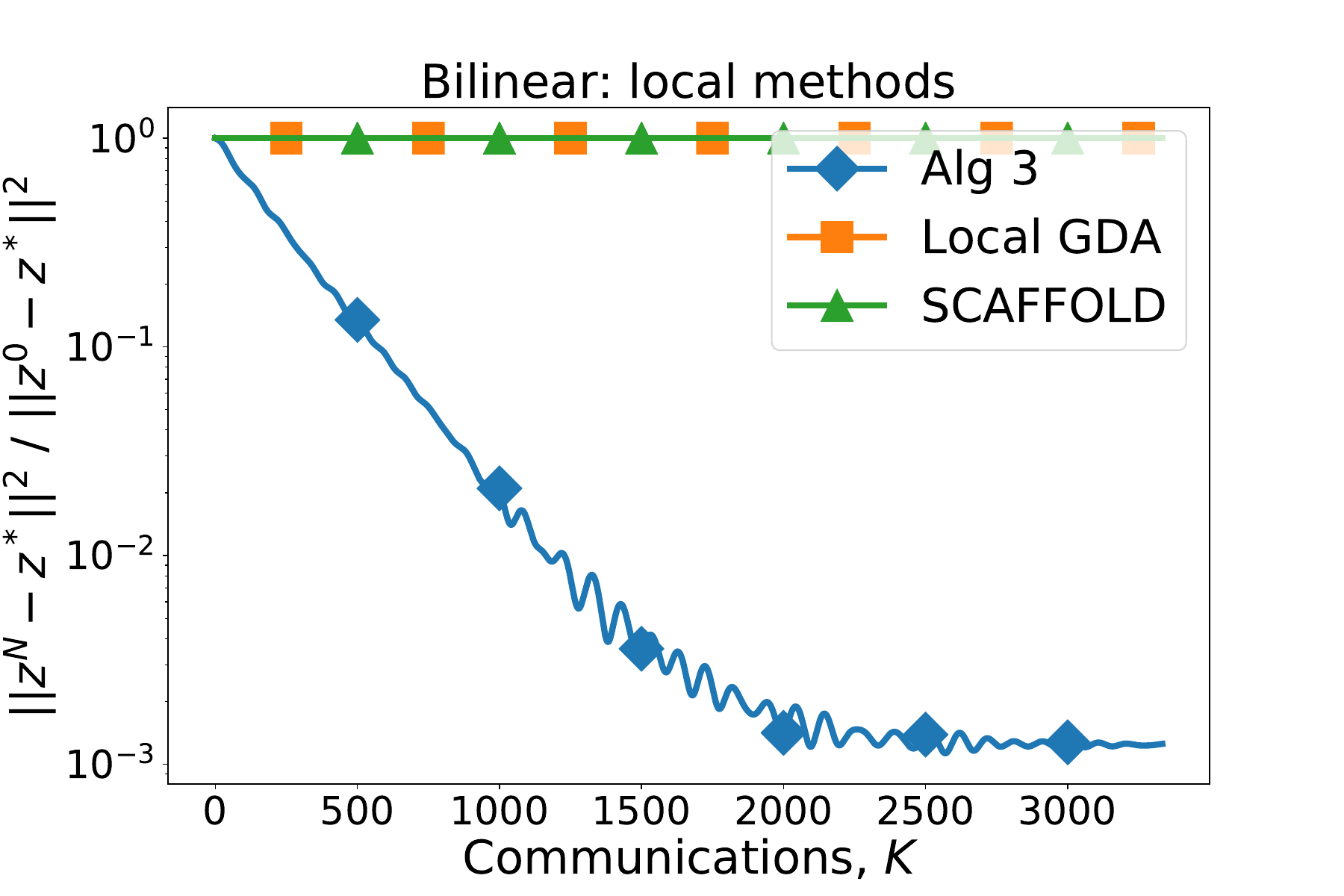}
\end{minipage}%
\begin{minipage}{0.33\textwidth}
  \centering
\includegraphics[width =  \textwidth ]{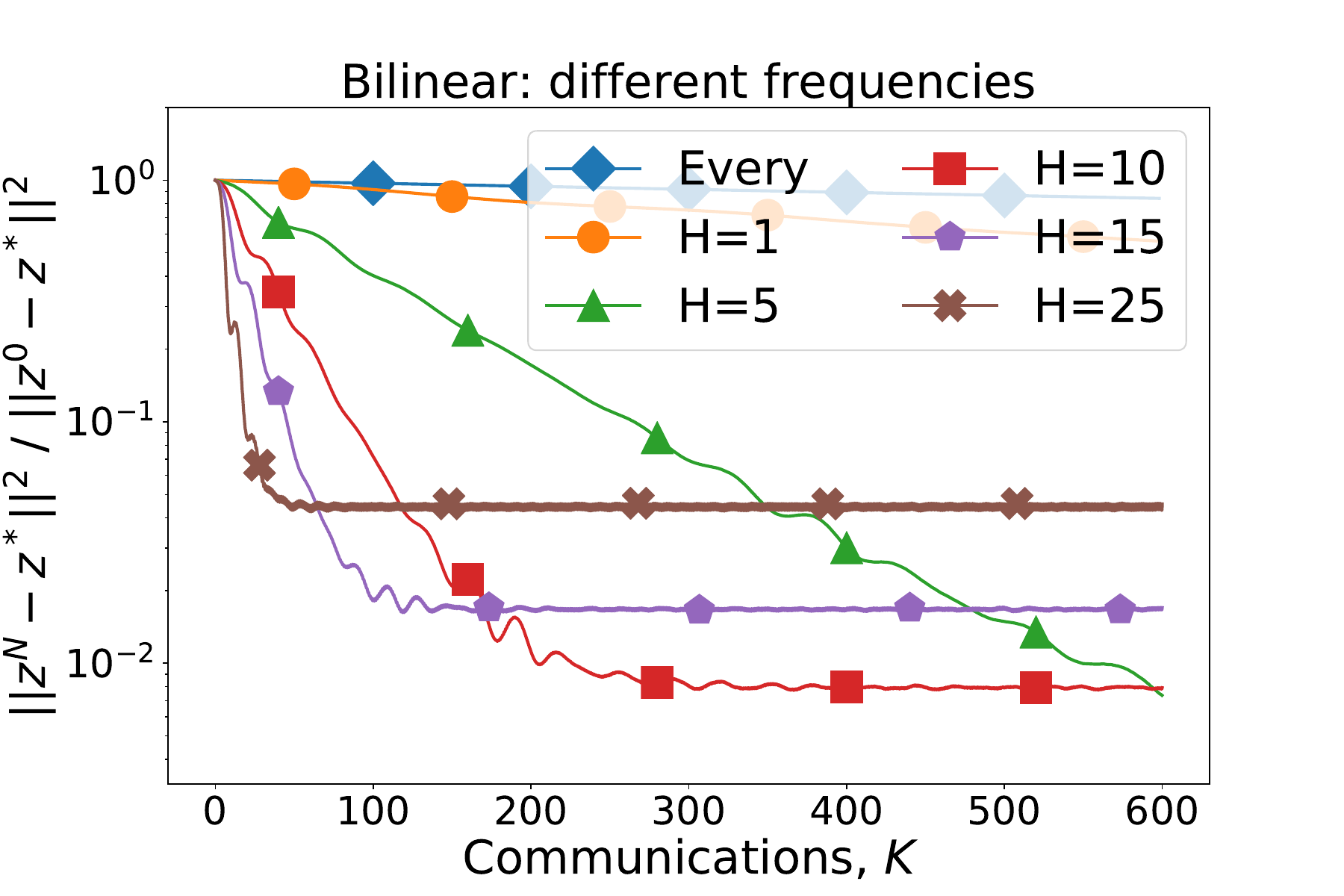}
\end{minipage}%
\begin{minipage}{0.33\textwidth}
  \centering
\includegraphics[width =  \textwidth ]{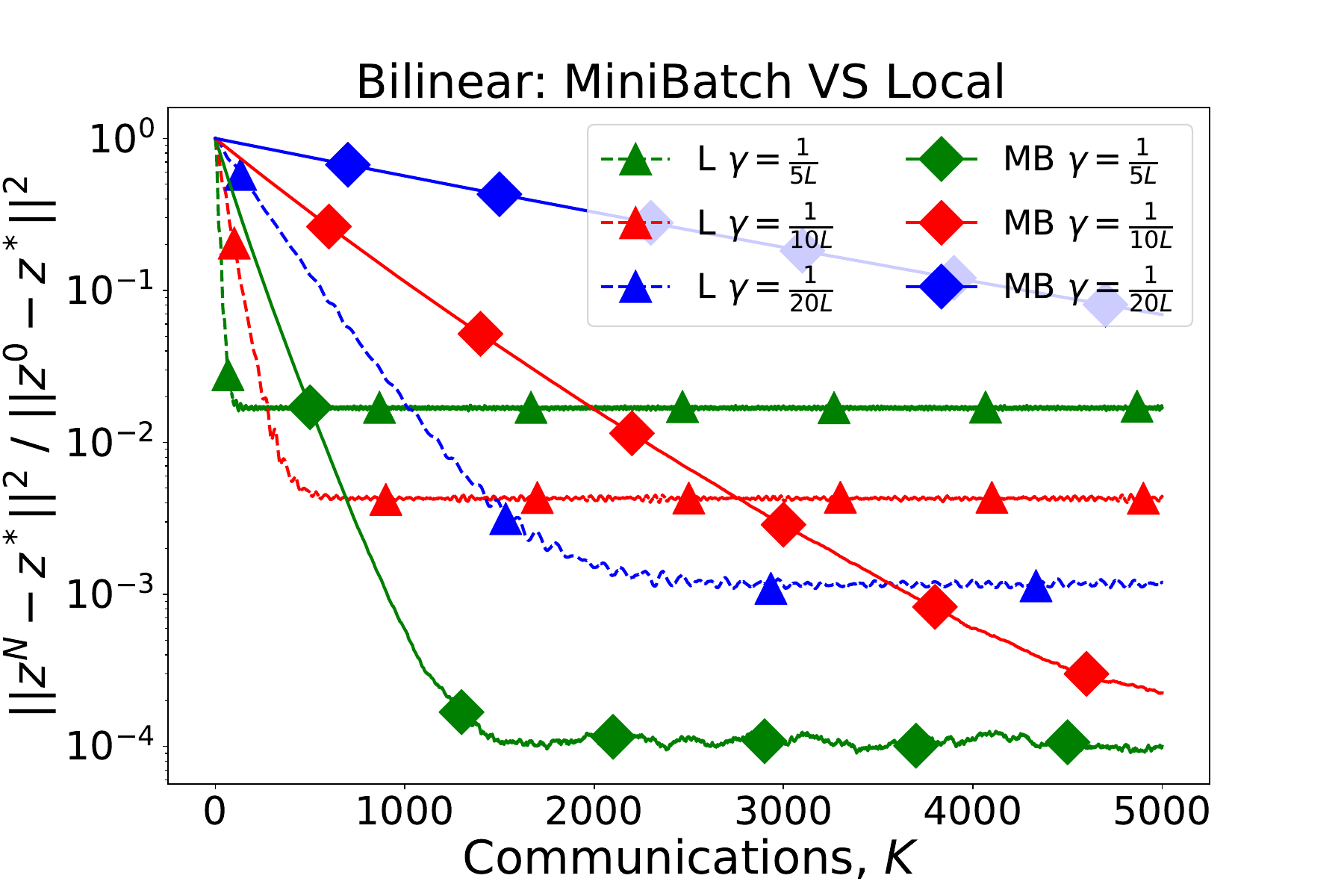}
\end{minipage}%
\\
\begin{minipage}{0.33\textwidth}
  \centering
(a)
\end{minipage}%
\begin{minipage}{0.33\textwidth}
  \centering
(b)
\end{minipage}%
\begin{minipage}{0.33\textwidth}
  \centering
(c)
\end{minipage}%
\caption{(a) Comparison of Algorithm \ref{alg4} and \cite{deng2021local,hou2021efficient} with $H = 3$ and tuned steps;
(b) Comparison of Algorithm \ref{alg4} with different communication frequencies $H$, as well as Algorithm \ref{alg1} with batch size 1 (blue line -- "Every") for \eqref{bilin}; 
(c) Comparison of Algorithm \ref{alg4} (L) with  communication frequencies $H = 3$ and Algorithm \ref{alg1} (MB) with batch size 6 for \eqref{bilin}.}
\label{fig:toy1}
\end{figure}

The purpose of the first experiment is to compare our local method (Algorithm \ref{alg4}) with the local approaches from papers \cite{deng2021local,hou2021efficient}. For all methods $H = 3$, and the step is chosen for best convergence. See Figure \ref{fig:toy1} (a) for the results. Note that our Algorithm \ref{alg4} outperforms the competitors. Moreover, the methods from the papers \cite{deng2021local,hou2021efficient} do not converge at all with any steps $\gamma$. As mentined above (Section \ref{sec:contr}), this is due to the fact that these methods are based on Descent-Ascent.

The next experiment is aimed at comparing Algorithm \ref{alg4} with different communication frequencies $H$. We take $\gamma = \frac{1}{15L}$.
From the point of view of communications (Figure \ref{fig:toy1} (b)), we get a standard result for local methods: less frequent communications, the faster convergence (in communications), but worse solution accuracy. This is due to fluctuations during local iterations, which lead away from the solution of the global solution. 

In the third experiment, we want to vary the step and compare Algorithm \ref{alg4} with a frequency of 3 and Algorithm \ref{alg1} with a batch of 6 (such parameters give that there are 6 local calls for one communication for both Algorithms). This problem statement is interesting because Algorithm \ref{alg1} is optimal, but Algorithm \ref{alg4} is not, but it can be better in practice. We see (Figure \ref{fig:toy1} (c)) that the local method wins in rate, but loses in extreme accuracy.

\subsection{Federated GAN}



\hspace{0.45cm} $\bullet$ \textbf{Model, data, optimizer. } A very popular enhancement of GANs is Conditional GAN, originally proposed in \cite{cgan}. It allows to direct the generation process by introducing class labels. 
We use a more complex Deep Convolutional GAN \cite{dcgan} with adjustments allowing to condition the output by class labels. We consider the CIFAR-10 \cite{cifar10} and split the dataset into $4$ parts. For each part, we select 2 majors classes that forms $30\%$ of the data, while the rest of the data split is uniformly filled by the other classes. As optimizers we use Algorithm \ref{alg4} and Local Adam \cite{kingma2014adam} -- a variation of Algorithm \ref{alg4}, but where the local gradient steps are replaced with Adam updates.

$\bullet$ \textbf{Setting. } Here we would like to consider the experiment of federated learning. Communication is a strong bottleneck of the federated setting, since the data is the local data of the users on their devices, and they may simply not be online to transmite information. Therefore, our goal is to reduce communication, which requires local methods. Then we want to compare how our optimizers work with a different number of local steps. In particular, we try to communicate once in an epoch, once in 5 epochs and once in 10 epochs. It is interesting to check how the frequency of communication affects the quality of the training.

\begin{figure}[h!]
\begin{minipage}{0.5\textwidth}
  \centering
\includegraphics[width =  0.9\textwidth ]{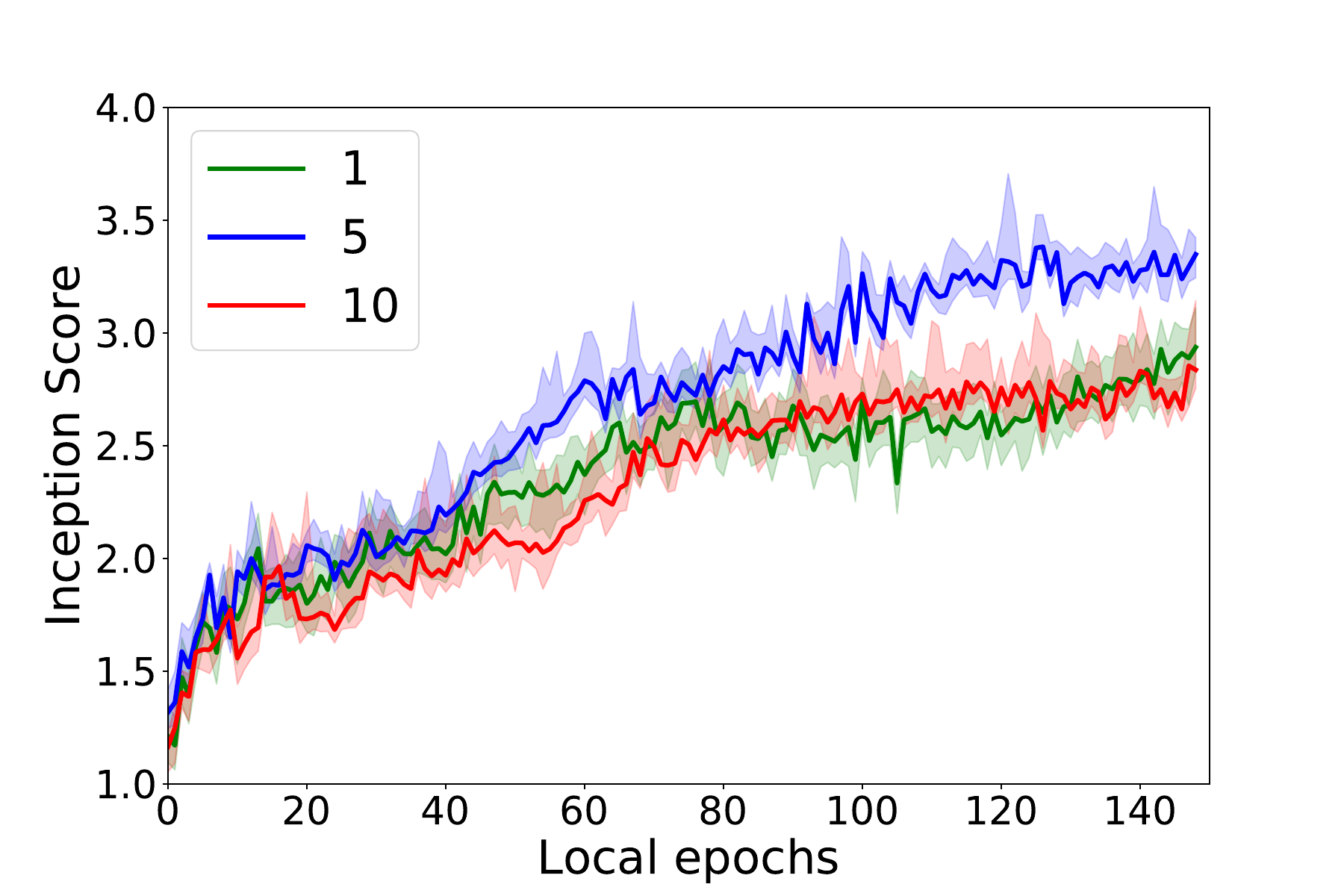}
\end{minipage}%
\begin{minipage}{0.5\textwidth}
  \centering
\includegraphics[width =  0.9\textwidth ]{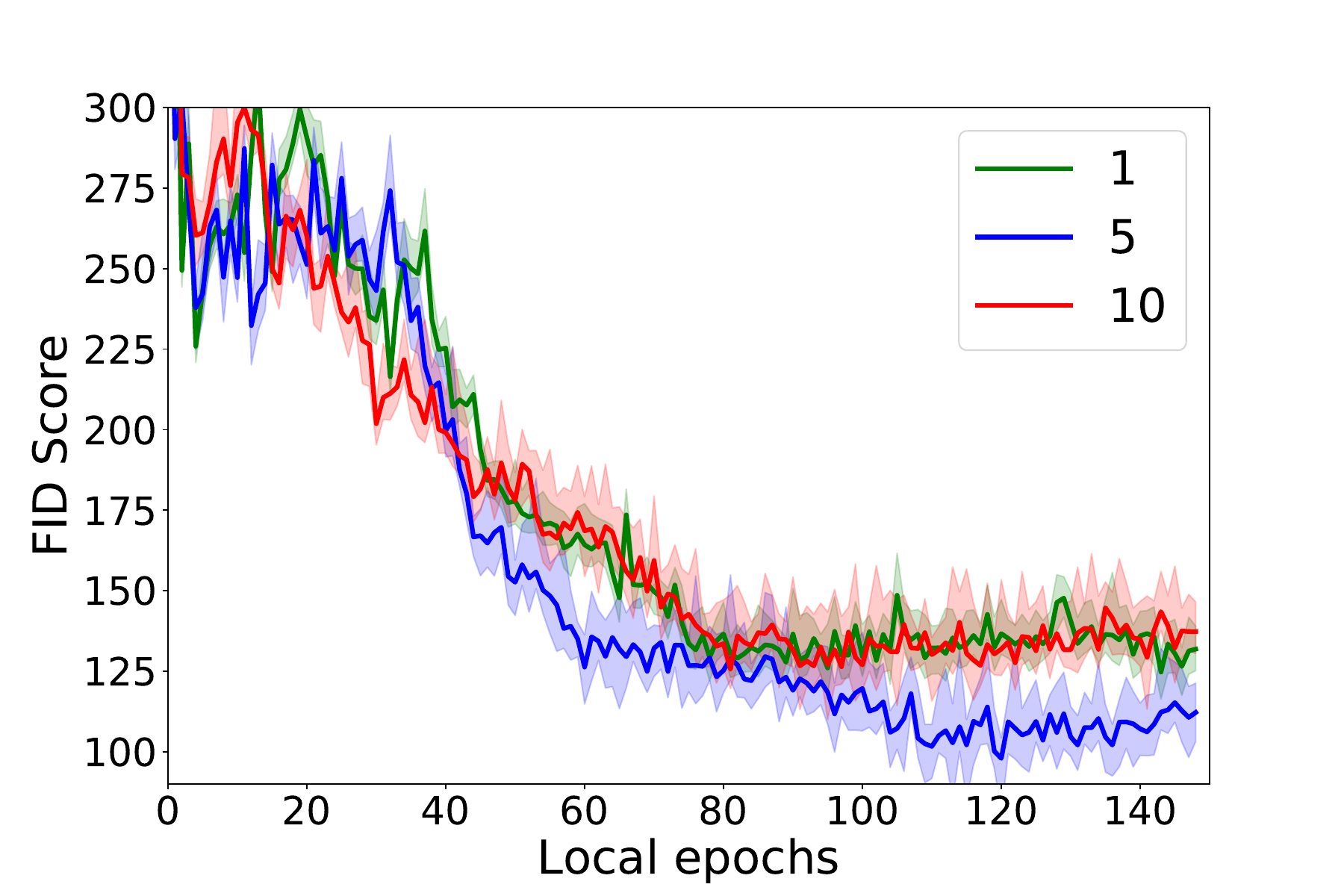}
\end{minipage}%
\begin{minipage}{0.5\textwidth}
  \centering
\end{minipage}%
\begin{minipage}{0.5\textwidth}
  \centering
\end{minipage}%
\caption{Comparison of three distances between communications in Local Adam in DCGAN distributed learning on CIFAR-10. We compare the FID Score and the Inception Score in terms of the local epochs number. 
The experiment was repeated 3 times on different data random splitting -- the maximum and minimum deviations are shown on the plots.
}
\label{fig:gans1}
\end{figure}

\begin{figure}[h!]
\begin{minipage}{0.33\textwidth}
  \centering
\includegraphics[width =  0.9\textwidth ]{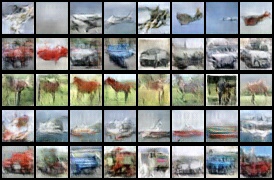}
\end{minipage}%
\begin{minipage}{0.33\textwidth}
  \centering
\includegraphics[width =  0.9\textwidth ]{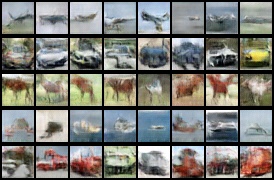}
\end{minipage}%
\begin{minipage}{0.33\textwidth}
  \centering
\includegraphics[width =  0.9\textwidth ]{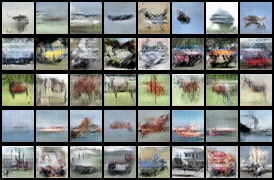}
\end{minipage}%
\\
\begin{minipage}{0.33\textwidth}
  \centering
  (a) $1$
\end{minipage}%
\begin{minipage}{0.33\textwidth}
  \centering
  (b) $5$
\end{minipage}%
\begin{minipage}{0.33\textwidth}
  \centering
  (c) $10$
\end{minipage}%
\caption{Pictures generated by DSGAN trained distributed on different distance between communications: (a)  1, (b) 5, (c) 10 epochs.}
\label{fig:gans2}
\end{figure}

$\bullet$ \textbf{Results. } Based on the results of experiments on bilinear problems (Section \ref{sec:be}), it was expected that methods which connect to the server less frequently (but do the same number of local epochs) would outperform their competitors in terms of communication budget. This trend is observed in Figures \ref{fig:gans1} and \ref{fig:gans2} -- methods making fewer communications do not lose in terms of FID and IS. On the other hand, the increasing distance between communications can have a significant impact on the quality of the training, especially in the last epochs. Therefore, we recommend using local methods with long gaps between communications only in the early stages of training, then it is worthwhile to communicate more and more frequently.

For more experiments with Algorithm \ref{alg4} and Local Adam on MNIST see Appendix \ref{add_exp}.

\section*{Acknowledgments}

The work on new versions of the paper was done in the Laboratory of Federated Learning Problems of the ISP RAS (Supported by Grant App. No. 2 to Agreement No. 075-03-2024-214).

\bibliography{refs}
\bibliographystyle{abbrv}

\clearpage
\appendix

\clearpage

\part*{Supplementary Material}

\section{General facts and technical lemmas}\label{sec:gf_tl}

\ans{\begin{lemma}
For an arbitrary integer $n\ge 1$ and arbitrary set of vectors $a_1,\ldots,a_n \in \mathbb{R}^d$ we have
\begin{equation}
    \left\|\sum\limits_{i=1}^n a_i\right\|^2 \le n\sum\limits_{i=1}^n \|a_i\|^2.\label{eq:squared_sum}
\end{equation}
\end{lemma}}

\begin{lemma}
Suppose given a convex closed set $\mathcal{Z}$, then the operator of the Euclidean projection onto this set is non-expansive, i.e. for all $z, z' \in \mathcal{Z}$,
\begin{equation}
    \label{proj}
    \| \mathrm{proj}_{\mathcal{Z}}(z) - \mathrm{proj}_{\mathcal{Z}}(z') \| \leq \|z - z' \|.
\end{equation}
\end{lemma}

\section{Proof of Theorems from Section  \ref{sec:lb}} \label{ap:lb}

As mentioned in the main part of the paper we consider the following model of functions:
\begin{equation}
\label{A1}
f_m(x,y) = f^{deter}_m (x^{deter},y) + f^{stoch}(x^{stoch}).
\end{equation}
Note that the function $f^{deter}_m$ uses the vector $x^{deter}$, and the function $f^{stoch}$ uses another vector $x^{stoch}$. The variables in the vectors $x^{deter}$ and $x^{stoch}$ do not intersect, but together $x^{deter}$ and $x^{stoch}$ form a complete vector $x$, for example, according to the following rule: $x_{2k-1} = x^{deter}_k$ and $x_{2k} = x^{stoch}_k$ for $k = 1,2 \ldots$. At the same time, for $f^{deter}_m$, we have access to $\nabla_x f^{deter}_m(x,y)$, $\nabla_y f^{deter}_m(x^{deter},y)$, and for $f^{stoch}$, to stochastic realizations $\nabla_x f^{stoch}_m(x^{stoch}, \xi)$ that satisfy Assumption \ref{as:as3}. 
Moreover, $f^{deter}_m$ are different for each device, but $f^{stoch}$ is the same.

We take <<bad>> functions with even $n_{x^{stoch}}= n_{x^{deter}} = n_y = n$. Moreover, $n$ must be taken large enough, as stated in Theorems. 

\subsection{Deterministic lower bounds}

We start with deterministic lower bounds. 
Our example builds on a splitting of the <<bad>> function for the non-distributed case from \cite{zhang2019lower}. 
Next, we give an example of the functions $f^{deter}_m (x^{deter},y)$ and their location on the nodes.
To simplify the notation, we use $f_m (x,y)$  instead of $f^{deter}_m (x^{deter},y)$ in this subsection. \add{Moreover, in the deterministic lower bounds, we consider the unconstrained problem over $\X \times \Y = \mathbb{R}^n \times \mathbb{R}^n$, hence the projection operators from \eqref{app:oracle-proj} are identical: $\text{proj}_{\X}(x) = x$, $\text{proj}_{\Y}(y) = y$. It means that we can simplify Definition \ref{app:proc} and remove \eqref{app:oracle-proj} from it.}
Next, we introduce some auxiliary arrangements of functions on the nodes, prove some facts for them, and then present the final <<bad>> examples and prove the lower bounds. 

Let $B \subset \mathcal{V}$ be a subset of the nodes of $G$. For $d \in \mathcal{N}$ we define $B_d = \{v \in \mathcal{V} ~:~ d(B,v) \geq d\}$, where $d(B,v)$ is a distance between the set $B$ and the node $v$ \ans{(the smallest number of edges between the vertex $v$ and the vertices from $B$)}. Then we construct the following  arrangement of bilinearly functions on nodes:
\begin{equation}
\label{t2}
f_m (x,y) = 
\begin{cases}
f_1 (x,y) = \frac{M}{2 |B_d|}\cdot\frac{L}{2} x^T A_1 y + \frac{\mu}{2}\|x\|^2 - \frac{\mu}{2}\|y\|^2 + \frac{M}{2 |B_d|}\cdot \frac{L^2}{2\mu}e_1^T y, & m \in B_d\\
f_2 (x,y) =\frac{M}{2 |B|} \cdot\frac{L}{2} x^T A_2 y +\frac{\mu}{2}\|x\|^2 - \frac{\mu}{2}\|y\|^2, & m \in B\\
f_3 (x,y) = \frac{\mu}{2}\|x\|^2 - \frac{\mu}{2}\|y\|^2, & \text{otherwise}
\end{cases}.
\end{equation}
where $e_1 = (1,0 \ldots, 0)$ and
\begin{equation*}
A_1 = \left(
\begin{array}{cccccccc}
1&0 & & & & & &  \\
&1 &-2 & & & & &  \\
& &1 &0 & & & & \\
& & &1 &-2 & & & \\
& & & &\ldots &\ldots & & \\
& & & & &1  &-2   & \\
& & &   & & &1 &0 \\
& & &  & & & &1 \\
\end{array}
\right), ~~
A_2 = \left(
\begin{array}{cccccccc}
1&-2 & & & & & &  \\
&1 &0 & & & & &  \\
& &1 &-2 & & & & \\
& & &1 &0 & & & \\
& & & &\ldots &\ldots & & \\
& & & & &1  &0   & \\
& & &   & & &1 &-2 \\
& & &  & & & &1 \\
\end{array}
\right).
\end{equation*}
In most cases, we want the simplest case with $|B| = |B_d| = 1$.

\begin{lemma} \label{l2}
Let the problem~\ans{\eqref{distr}}+\eqref{t2} be solved by any method  that satisfies Definition \ref{app:proc}. Then after $K$ communication rounds,  only the first $\left\lfloor \frac{K}{d} \right\rfloor$ coordinates of the   global output can be non-zero while  the rest of the $n-\left\lfloor \frac{K}{d} \right\rfloor$ coordinates are strictly equal to zero. 
\end{lemma}

\textbf{Proof:}
We begin introducing some notation for our proof:
\begin{equation*}
    E_{0} := \{ 0\}, \quad E_{k} := \text{span} \{ e_1, \ldots, e_k\}.
\end{equation*}
Note that, the initialization from Definition \eqref{app:proc} gives $\mathcal{M}^x_{m} = E_0$, $\mathcal{M}^y_{m} = E_0$.

Suppose that, for some $m$,  $\mathcal{M}^x_{m} = E_k$ and $\mathcal{M}^y_{m} = E_k$, at some given iteration. Let us analyze how $\mathcal{M}^x_{m}, \mathcal{M}^y_{m}$ can change by performing only local computations. 

Firstly, we consider the case when $k$ odd. After one local update, we have the following: 

$\bullet$ For  machines $m$  which own $f_1$, it holds
\begin{equation}\begin{aligned}
\label{update_lower1}
        x \in \text{span} \big\{&e_1~,~ x'~,~A_{1} y'\big\} = E_k,\\
        y \in \text{span} \big\{&e_1~,~y'~,~A_1^T x'\big\} = E_k,
\end{aligned}\end{equation}
for given $x' \in \mathcal{M}^x_{m}$ and  $y' \in \mathcal{M}^y_{m}$. Since $A_1$ has a block diagonal structure, after one local computation, we have  $\mathcal{M}^x_{m} = E_k$ and $\mathcal{M}^y_{m} = E_k$. The situation does not change, no matter how many local computations one  does.


$\bullet$ For machines $m$ which own $f_2$, it holds
\begin{equation*}\begin{aligned}
        x \in \text{span} \big\{&x'~,~A_2 y'\big\} = E_{k+1},\\
        y  \in \text{span} \big\{&y'~,~A_2^T x'\big\} = E_{k+1}, 
\end{aligned}\end{equation*}
for given $x' \in \mathcal{M}^x_{m}$ and  $y' \in \mathcal{M}^y_{m}$. It means that, after local computations (at least one local computation), one has  $\mathcal{M}^x_{m} = E_{k+1}$ and $\mathcal{M}^y_{m} = E_{k+1}$. Therefore,  machines with function $f_2$ can progress by one new non-zero coordinate. \ans{The situation with even $k$ is opposite, the devices with $f_1$ can increase the number of non-zero coordinates by exactly $1$ from $k$ to $k+1$, but the machines with $f_2$ do not progress from local computations.}

This means that we constantly have to transfer progress from the group of machines with $f_1$ to the group of machines with $f_2$ and back. Initially, all devices have zero coordinates.  Furthermore, after at least one local computation, the machines with $f_1$ can receive the first non-zero coordinate using $e_1$ in the gradients for $y$ (but only the first, not the second), and the rest of the devices are left with all zeros. Next, we pass the first non-zero coordinate to machines with $f_2$. This requires  $d$ communication rounds. \ans{The devices with $f_2$ now give a progress on one more non-zero coordinate (the second) according to the reasoning above, this update is passed to the machines with $f_1$ and there give another progress.} Then the process continues in the same way. It remains to note that for this update in the number of non-zero coordinates, we need at least one local calculation for each non-zero coordinate. Note that the local computation budget is sufficient ($T > K$ -- see Definition \ref{app:proc}). This  completes the proof.
\EndProof

Consider the problem with the global objective function:
\begin{align}
    \label{t144}
    f(x,y) &:= \frac{1}{M} \sum\limits_{m=1}^M f_m(x,y) \nonumber\\
    &= \frac{1}{M} \left( |B_d| \cdot f_1(x,y) + |B| \cdot f_2(x,y) + (M - |B_d| - |B|) \cdot f_3(x,y)\right) \nonumber\\
    &= \frac{L}{2} x^T A y + \frac{\mu}{2}\|x\|^2 - \frac{\mu}{2}\|y\|^2 + \frac{L^2}{4\mu} e_1^T y, ~~~\text{with}~~~A = \frac{1}{2}(A_1 + A_2)
\end{align}

With the fact that $\| A\| \leq 2$, one can easy verify that \eqref{t144} satisfies Assumptions \ref{ass:as1g} and \ref{ass:as2g}.

The previous lemma gives an idea of what the solution obtained using procedures that satisfy Definition \ref{app:proc}. The next lemma is already to the approximate solution of the problem \eqref{distr} + \eqref{t144} and how it is closed to the real solution.

\begin{lemma}[Lemma 3.3 from \cite{zhang2019lower}]\label{lem2}
Let $\alpha = \frac{4\mu^2}{L^2}$ and $q = \frac{1}{2}\left(2 + \alpha - \sqrt{\alpha^2 + 4\alpha} \right) \in (0;1)$ be the smallest root of $q^2 - (2 + \alpha) q + 1 = 0$, and let \add{us} introduce approximation $\bar y^*$ as follows
\begin{equation}
\label{eq:approx}
    \bar y^*_i = \frac{q^i}{1-q}.
\end{equation}
\add{Then the error between this approximation and the exact solution} of \eqref{distr} + \eqref{t144} can be bounded
\begin{equation*}
    \|\bar y^* - y^*\| \leq \frac{q^{n+1}}{\alpha(1-q)}.
\end{equation*}
\end{lemma}
\textbf{Proof:} \ans{For the problem \eqref{distr}+\eqref{t144}, we can write down the optimality condition $\nabla_x f(x^*, y^*) = 0$, express $x^*$ through $y^*$, substitute it in $f$ and obtain the following optimization problem \add{in} $y$:}
\begin{equation*}
    g(y) = -\frac{1}{2}y^T \left(\frac{L^2}{4\mu}A^T A + \mu I \right)y + \frac{L^2}{4\mu} e_1^T y,
\end{equation*}
where one can easy found 
\begin{equation*}
A^T A = \left(
\begin{array}{cccccccc}
1&-1 & & & & & &  \\
-1&2 &-1 & & & & &  \\
&-1 &2 & -1 & & & & \\
& & -1&2 &-1 & & & \\
& & &-1 &2 &-1 & & \\
& & & & &\ldots & & \\
& & & & &-1 &2 &-1 \\
& & & & & &-1 &2 \\
\end{array}
\right).
\end{equation*}
The optimality $\nabla g(y^*) = 0$ gives
\begin{equation*}
    \left(\frac{L^2}{4\mu}A^T A + \mu I \right)y^* = \frac{L^2}{4\mu} e_1,
\end{equation*}
or
\begin{equation*}
    \left(A^T A + \alpha I \right)y^* = e_1.
\end{equation*}
Let us write in the form of a set of equations:
\begin{equation*}
\left\{
\begin{array}{l}
(1+\alpha)y_1^* - y_2^* = 1 \\
-y_1^* + (2 + \alpha) y^*_2 - y^*_3 = 0\\
\ldots \\
-y_{n-2}^* + (2 + \alpha) y^*_{n-1} - y^*_n = 0\\
-y^*_{n-1} + (2+\alpha)y^*_n = 0
\end{array}
\right .
\end{equation*}
Note that the approximation \eqref{eq:approx} satisfies the following set of equations:
\begin{equation*}
\left\{
\begin{array}{l}
(1+\alpha)\bar y_1^* - \bar y_2^* = 1 \\
-\bar y_1^* + (2 + \alpha) \bar y^*_2 - \bar y^*_3 = 0\\
\ldots \\
-\bar y_{n-2}^* + (2 + \alpha) \bar y^*_{n-1} - \bar y^*_n = 0\\
-\bar y^*_{n-1} + (2+\alpha) \bar y^*_n = \frac{q^{n+1}}{1-q}
\end{array}
\right .
\end{equation*}
or in the short form:
\begin{equation*}
    \left(A^T A + \alpha I \right)\bar y^* = e_1 + \frac{q^{n+1}}{1-q}e_n.
\end{equation*}
Then the difference between the approximation and the true solution is
\begin{equation*}
    \bar y^* - y^* = \left(A^T A + \alpha I \right)^{-1}\frac{q^{n+1}}{1-q}e_n,
\end{equation*}
With the fact that $\alpha^{-1} I \succeq \left(A^T A + \alpha I \right)^{-1} \succ 0$, it implies the statement of Lemma.
\EndProof

Now we formulate a key lemma (similar to Lemma 3.4 from \cite{zhang2019lower}).

\begin{lemma} \label{l234}
For any pairs $T, K$ ($T > K$) one can found the distributed saddle point problem in the form \eqref{distr}+\eqref{t2}+\eqref{t144} with $B_d \neq \varnothing$ and the size $n \geq \max \left\{ 2 \log_q \left( \frac{\alpha}{4\sqrt{2}}\right), 2K\right\}$, where $\alpha = \frac{4\mu^2}{L^2}$ and $q = \frac{1}{2}\left(2 + \alpha - \sqrt{\alpha^2 + 4\alpha} \right) \in (0;1)$, such that any output $\hat x, \hat y$ produced by  any {\tt BBP}$(T,K)$ satisfying Definition \ref{app:proc} after $K$ communications rounds and $T$ local computations, \add{satisfies} the following estimate:
\begin{equation*}
    \|\hat x - x^*\|^2 + \|\hat y - y^*\|^2 \geq q^{\frac{2K}{d}} \frac{\| y^0 - y^*\|^2}{16}.
\end{equation*}
\end{lemma}
\ans{\textbf{Proof:} 
Lemma \ref{l2} states that after $K$ ($K < T$) communications only $k = \left\lfloor \frac{K}{d} \right\rfloor$ coordinates in the output $\hat y$ can be non-zero. Therefore, by the definition of $\bar y^*$ from \eqref{eq:approx}, by $k \leq K \leq \frac{n}{2}$ and with $q < 1$, we have
\begin{align}
    \label{eq:ttt444}
    \|\hat y - \bar y^*\| &\geq \sqrt{\sum\limits_{j=k+1}^n  (\bar y^*_j)^2} = \frac{q^k}{1-q} \sqrt{q^2 + q^4 + \ldots + q^{2(n-k)}} \notag\\
    &\geq \frac{q^k}{\sqrt{2}(1-q)} \sqrt{q^2 + q^4 + \ldots + q^{2n}} = \frac{q^k}{\sqrt{2}} \| \bar y^*\| = \frac{q^k}{\sqrt{2}} \| y^0 - \bar y^*\|.
\end{align}
With Lemma \ref{lem2}, we can guarantee that 
$$
\| \bar y^* - y^*\| \leq \frac{q^{n+1}}{\alpha(1-q)} \leq \frac{q^{\frac{n}{2}}}{\alpha} \cdot q^k \cdot \frac{q}{1-q}.
$$
Here we also used that $n \geq 2K \geq 2k$ and $q < 1$. Next, we take into account that $n \geq 2 \log_q \left( \frac{\alpha}{4\sqrt{2}}\right)$ and get
\begin{equation}
    \label{eq:ttt445}
    \| \bar y^* - y^*\| \leq \frac{1}{4 \sqrt{2}} \cdot q^k \cdot \frac{q}{1-q} \leq \frac{1}{4 \sqrt{2}} \cdot q^k \cdot \| y^0  - \bar y^*\|,
\end{equation}
where we also noticed from Lemma \ref{lem2} that $\| y^0 - \bar y^* \| = \| \bar y^* \| \geq \frac{q}{1-q}$. Combining \eqref{eq:ttt444} and \eqref{eq:ttt445}, we obtain
\begin{align*}
    \|\hat x - x^*\|^2 + \|\hat y - y^*\|^2 &\geq \|\hat y - y^*\|^2
    \\
    &\geq  \left(\|\hat y - \bar y^*\| - \|\bar y^* - y^* \|\right)^2
    \\
    &\geq \|\hat y - \bar y^*\|^2 - 2 \|\bar y^* - y^* \| \cdot \|\hat y - \bar y^*\|
    \\
    &\geq \|\hat y - \bar y^*\|^2 - \frac{2 q^k}{4 \sqrt{2}} \| y^0  - \bar y^*\| \cdot \|\hat y - \bar y^*\|.
\end{align*}
We need to minimize the quadratic function: $\phi(t) = t^2 -  \frac{2 q^k}{4 \sqrt{2}} \| y^0  - \bar y^*\| t$, for $t \geq \frac{q^k}{\sqrt{2}} \| y^0 - \bar y^*\|$. It is easy to see that $t^* = \frac{q^k}{\sqrt{2}} \| y^0 - \bar y^*\|$, then
\begin{equation}
    \label{eq:ttt447}
    \|\hat x - x^*\|^2 + \|\hat y - y^*\|^2 \geq \frac{q^{2k}}{4} \| y^0  - \bar y^*\|^2.
\end{equation}
It remains to note that
\begin{equation}
    \label{eq:ttt446}
    \| y^0 - y^* \| \leq \| y^0 - \bar y^* \| + \| \bar y^* - y^* \| \leq \left(1 + \frac{1}{4 \sqrt{2}} \cdot q^k\right) \cdot \| y^0  - \bar y^*\| \leq 2 \| y^0  - \bar y^*\|.
\end{equation}
Here we used \eqref{eq:ttt445} and $q<1$. Substituting \eqref{eq:ttt446} into \eqref{eq:ttt447}, we have
\begin{equation*}
    \|\hat x - x^*\|^2 + \|\hat y - y^*\|^2 \geq \frac{q^{2k}}{16} \| y^0 - y^*\|^2 = q^{2\left\lfloor \frac{K}{d} \right\rfloor } \frac{\| y^0 - y^*\|^2}{16} \geq q^{\frac{2K}{d}} \cdot \frac{\| y^0 - y^*\|^2}{16}.
\end{equation*}
}
\EndProof

Building on the above preliminary results, we are now ready to prove our complexity lower bounds as stated in Theorems \ref{th1} and \ref{th2}.

\subsubsection{Centralized case}

\ans{
\begin{theorem}[Theorem \ref{th1}]
For any $L > \mu >0$ and any connected graph with diameter $\Delta$, there exists a distributed saddle point problem on $\mathcal{X} \times \mathcal{Y} = \mathbb{R}^n \times \mathbb{R}^n$ with $x^*, y^* \neq 0$ over a fixed network, such that the following statements hold:
\begin{itemize}
    \item the diameter of the network is equal to $\Delta$,
    \item $f = \frac{1}{M} \sum\limits_{m=1}^M f_m$ is $L$-smooth, $\mu$-strongly convex-strongly concave ,
    \item size $n \geq \max \left\{ 2 \log_q \left( \frac{\alpha}{4\sqrt{2}}\right), 2K\right\}$, where $\alpha = \frac{4\mu^2}{L^2}$ and $q = \frac{1}{2}\left(2 + \alpha - \sqrt{\alpha^2 + 4\alpha} \right) \in (0;1)$,
    \item for any output $\hat x, \hat y$ of any {\tt BBP}$(T,K)$ (Definition \ref{app:proc}), the following \add{estimate} hold:
\begin{equation*}
    \|\hat x - x^*\|^2 + \|\hat y - y^*\|^2 = \Omega\left(\exp\left( - \frac{4 \mu}{L - \mu} \cdot \frac{K}{\Delta} \right) \| y^0 - y^*\|^2\right).
\end{equation*}
\end{itemize}
\end{theorem}
}

\textbf{Proof:} It suffices to consider a linear graph on $\Delta + 1$ vertices $\{v_1, \ldots, v_{\Delta+1}\}$ and apply Lemma \ref{l234} for the problem \eqref{distr}+\eqref{t2}+\eqref{t144} with $B = \{v_1\}$ and $d = \Delta$. Then
\begin{equation*}
    \left(\frac{1}{q}\right)^{\frac{2K}{\Delta}} \geq  \frac{\| y^0 - y^*\|^2}{16(\|\hat x - x^*\|^2 + \|\hat y - y^*\|^2)}.
\end{equation*}
Taking the logarithm of the two parts of the inequality, we get
\begin{equation*}
    \frac{2K}{\Delta}  \geq  \ln\left(\frac{\| y^0 - y^*\|^2}{16(\|\hat x - x^*\|^2 + \|\hat y - y^*\|^2)}\right) \frac{1}{\ln(q^{-1})}.
\end{equation*}
Next, we work with
\begin{align*}
    \frac{1}{\ln (q^{-1})} &= \frac{1}{\ln (1+ (1-q)/q))} \geq \frac{q}{1-q} = \frac{1 + \frac{2 \mu^2}{L^2} - 2\sqrt{\frac{\mu^2}{L^2} + \left(\frac{\mu^2}{L^2}\right)^2}}{2\sqrt{\frac{\mu^2}{L^2} + \left(\frac{\mu^2}{L^2}\right)^2} - \frac{2\mu^2}{L^2}} \nonumber\\ 
    &= \frac{2\sqrt{\frac{\mu^2}{L^2} + \left(\frac{\mu^2}{L^2}\right)^2} - \frac{2\mu^2}{L^2}}{\frac{4\mu^2}{L^2}}  \nonumber\\ 
    &= \frac{1}{2}\sqrt{\frac{L^2}{\mu^2} + 1} - \frac{1}{2}.
\end{align*}
Finally, one can obtain
\begin{equation*}
    \frac{2K}{\Delta}  \geq  \ln\left(\frac{\| y^0 - y^*\|^2}{16(\|\hat x - x^*\|^2 + \|\hat y - y^*\|^2)}\right) \cdot \frac{1}{2}\left(\frac{L}{\mu} - 1\right),
\end{equation*}
and 
\begin{equation*}
    \exp\left(\frac{4\mu}{L - \mu}\frac{K}{\Delta}\right)  \geq  \frac{\| y^0 - y^*\|^2}{16(\|\hat x - x^*\|^2 + \|\hat y - y^*\|^2)},
\end{equation*}
which completes the proof.
\EndProof

\subsubsection{Decentralized case}

\ans{
\begin{theorem}[Theorem \ref{th2}]
For any $L > \mu >0$ and any connected graph with diameter $\Delta$, there exists a distributed saddle point problem on $\mathcal{X} \times \mathcal{Y} = \mathbb{R}^n \times \mathbb{R}^n$ with $x^*, y^* \neq 0$ over a fixed network characterized by a gossip matrix, such that the following statements hold:
\begin{itemize}
    \item the gossip matrix $W$ have the condition number $\chi$,
    \item $f = \frac{1}{M} \sum\limits_{m=1}^M f_m$ is $L$-smooth, $\mu$-strongly convex-strongly concave,
    \item size $n \geq \max \left\{ 2 \log_q \left( \frac{\alpha}{4\sqrt{2}}\right), 2K\right\}$, where $\alpha = \frac{4 \mu^2}{L^2}$ and $q = \frac{1}{2}\left(2 + \alpha - \sqrt{\alpha^2 + 4\alpha} \right)$,
    \item for any output $\hat x, \hat y$ of any {\tt BBP}$(T,K)$ (Definition \ref{app:proc}),  the following \add{estimate} hold:
\begin{equation*}
    \|\hat x - x^*\|^2 + \|\hat y - y^*\|^2 = \Omega\left(\exp\left( - \frac{32\mu}{L - \mu} \cdot \frac{K}{\sqrt{\chi}} \right) \| y^0 - y^*\|^2\right).
\end{equation*}
\end{itemize}
\end{theorem}
}

\textbf{Proof:}
The proof follow similar steps as in the  proof of  \cite[Theorem 2]{scaman2017optimal}. 
Let $\gamma_M = \frac{1 - \cos \frac{\pi}{M}}{1 + \cos \frac{\pi}{M}}$ be a  decreasing sequence of positive numbers. Since $\gamma_2 = 1$ and $\lim_m \gamma_M = 0$, there exists $M \geq 2$ such that $\gamma_M \geq \tfrac{1}{\chi} > \gamma_{M+1}$.

$\bullet$ If $M \geq 3$, let us consider as a network a linear graph $\mathcal{G}$ of size $M$ with vertexes $v_1, \ldots v_M$, and weighted with $w_{1,2} = 1 - a$, $w_{i,i+1} = 1$ for $i \geq 2$. We apply Lemma \ref{l234} for problem \eqref{distr}+\eqref{t2}+\eqref{t144} with $B = \{v_1\}$ and $d = M-1$, then we have $B_d = \{v_M\}$. Hence,
\begin{equation*}
    \|\hat x - x^*\|^2 + \|\hat y - y^*\|^2 \geq q^{\frac{2K}{d} } \frac{\| y^0 - y^*\|^2}{16}.
\end{equation*}
We consider $W_a$ is the Laplacian of the weighted graph $\mathcal G$. $W_a$ satisfies Definition \ref{def:goss}. 
One can note that with $a = 0$, $\tfrac{1}{\chi(W_a)} = \gamma_M$, with $a = 1$, we have $\tfrac{1}{\chi(W_a)} = 0$ (since the network is disconnected). Hence, there exists $a \in (0;1]$ such that $\tfrac{1}{\chi(W_a)} = \chi$. Then $\tfrac{1}{\chi} \geq \gamma_{M+1} \ans{= \frac{1 - \cos \frac{\pi}{M+1}}{1 + \cos \frac{\pi}{M+1}}} \geq \frac{2}{(M+1)^2}$, and $M \geq \sqrt{2 \chi} - 1 \geq \frac{\sqrt{\chi}}{4}$. Finally, since $M \geq 3$, we get $d = M -1 \geq \tfrac{M}{2} \geq  \tfrac{\sqrt{\chi}}{8}$. Hence,
\begin{equation*}
    \|\hat x - x^*\|^2 + \|\hat y - y^*\|^2 \geq q^\frac{16K }{\sqrt{\chi}} \frac{\| y^0 - y^*\|^2}{16}.
\end{equation*}
Similarly to the proof of the previous theorem
\begin{equation}
    \label{r509}
    \exp\left(\frac{32\mu}{L - \mu}\frac{K}{\sqrt{\chi}}\right)  \geq  \frac{\| y^0 - y^*\|^2}{16(\|\hat x - x^*\|^2 + \|\hat y - y^*\|^2)}.
\end{equation}

$\bullet$ If $M = 2$, we construct a fully connected network with 3 nodes with weight $w_{1,3} = a \in [0;1]$. Let $W_a$ is the Laplacian. If $a = 0$, then the network is a linear graph and $\rho(W_a) = \gamma_3 =\frac{1}{3}$. Hence, there exists $a \in [0;1]$ such that $\chi(W_a) = \chi$. Finally, $B = \{v_1\}$, $B_d = \{v_3\}$ and $d \geq 1 \geq \frac{\sqrt{\chi}}{2}$. Whence, it follows that in this case \eqref{r509} is also valid.
\EndProof
\subsection{Stochastic lower bounds}

\subsubsection{Strongly convex case}

We consider the following simple problem with function $f: \mathbb{R} \to \mathbb{R}$:
\begin{equation}
    \label{stoch_str}
    \min_{x \in \mathbb{R}} f(x) = \frac{\mu}{2}(x - x^0)^2,
\end{equation}
where we do not know the constant $x^0 \neq 0$. $f(x)$ is a $\mu$-strongly convex and $\mu$-smooth function. We minimize this function by using stochastic first order oracle 
\begin{equation*}
    \nabla f(x, \xi) = \mu(x + \xi - x^0), ~\text{where}~ \xi \in \mathcal{N}\left(0, \frac{\sigma^2}{\mu^2}\right).
\end{equation*}
One can note that $\EE[\nabla f(x, \xi)] = \mu(x - x^0) = \nabla f(x)$, and $\EE\left[|\nabla f(x, \xi) - \nabla f(x)|^2\right] = \EE\left[\mu^2|\xi|^2\right] = \sigma^2$.
We use some {\tt BBP}$(T,K)$ (Definition \ref{app:proc}), which calls the stochastic oracle $N = MT$ times in some set of points $\{x_i\}_{i=1}^N$, for these points oracle returns $y_i = \mu( x_i - x^0 + \xi_i)$, where all  $\xi_i \in \mathcal{N}(0, \sigma^2/\mu^2)$ and independent. Using $x_i, y_i$, one can compute point $z_i = x_i - y_i / \mu = x^0 - \xi_i \in \mathcal{N}(x^0, \sigma^2/\mu^2)$ and independent.
Hence, the original problem \eqref{stoch_str} and the working of any {\tt BBP} are easy to 
reformulate in the following way: after $N$ calls of the oracle we have set of pairs $\{(x_i, z_i)\}_{i=1}^N$, where $z_i \in \mathcal{N}(x^0, \sigma^2/\mu^2)$ and independent. By these pairs we need to estimate the unknown constant $x^0$. One can do it by MLE:
$$x^{\text{MLE}}_N = \frac{1}{N} \sum\limits_{i=1}^N z_i, ~~~ x^{\text{MLE}}_N \in \mathcal{N}\left(x^0, \frac{\sigma^2}{N\mu^2}\right).$$
Then
$$\EE\left[\|x^{\text{MLE}}_N  - x^*\|^2\right] = \EE\left[|x^{\text{MLE}}_N  - x^0|^2\right] = \text{Var}\left[x^{\text{MLE}}_N\right] = \frac{\sigma^2}{N\mu^2},$$
or
$$\EE\left[f(x^{\text{MLE}}_N)  - f(x^*)\right] = \frac{\mu}{2}\EE\left[|x^{\text{MLE}}_N  - x^0|^2\right] = \frac{\mu}{2} \text{Var}\left[x^{\text{MLE}}_N\right] = \frac{\sigma^2}{2 N\mu}.$$
We need to show that the estimate obtained with the MLE is the best in terms of $N$. For this we need the classical statistical fact \cite{10.1214/aoms/1177729840}:
\ans{\begin{lemma}
    The unique estimator that is minimax for the quadratic loss function is the MLE.
\end{lemma}}
Then we have the following theorem:
\begin{theorem}[Theorem \ref{th3}]
For any $L > \mu >0$, there exists a stochastic minimization problem with $L$-smooth and  $\mu$-strongly convex function (i.e., satisfying Assumptions \ref{ass:as1g} and \ref{ass:as2g}),  such that for any output $\hat x$ of any {\tt BBP}$(T,K)$ (Definition \ref{app:proc}) with $M$ workers, one can obtain the following estimate:
\begin{equation*}
    \add{\mathbb{E}\left[\|\hat x - x^*\|^2\right]} =  \Omega\left(\frac{\sigma^2}{MT\mu^2}\right).
\end{equation*}
\end{theorem}

\subsubsection{Convex case}

For the convex case, we work with
\begin{equation}
\label{stoch_conv}
\min_{x \in [-\frac{\Omega_x}{2}, \frac{\Omega_x}{2}]} \frac{\tilde \varepsilon}{\Omega_x} x ,
\end{equation}
where $\tilde \varepsilon$ can only take two values $\varepsilon$ or $-\varepsilon$ with some positive $\varepsilon$. 
Of course, we do not know which of the two values $\tilde \varepsilon$ takes. We can assume, for example, that $\tilde \varepsilon$ is randomly chosen with equal probability at the beginning. It is easy to check that \eqref{stoch_conv} is convex and $L$-smooth for any $L$ and $\varepsilon$. 
The first order stochastic oracle returns $\nabla f(x, \xi) = \xi \in \mathcal{N}(\tilde \varepsilon/\Omega_x, \sigma^2)$.  One can note that $\EE[\nabla f(x, \xi)] = \tilde \varepsilon/\Omega_x = \nabla f(x)$, and $\EE\left[|\nabla f(x, \xi) - \nabla f(x)|^2\right] = \sigma^2$.
We use some procedure {\tt BBP}$(T,K)$ (Definition \ref{app:proc}), which calls  the oracle $N = MT$ times in some set of points $\{x_i\}_{i=1}^N$. For these points, the oracle returns $\xi_i$, where all  $\xi_i \in \mathcal{N}(\tilde \varepsilon, \sigma^2)$ and independent. Note that we can say in advance that $x^* = \frac{\Omega_x}{2}$ if $\tilde \varepsilon = -\varepsilon$ and $x^* = -\frac{\Omega_x}{2}$ if $\tilde \varepsilon = \varepsilon$. We have a rather simple task, from independent samples $\{ \xi_i\}_{i=1}^N \in \mathcal{N}(\tilde \varepsilon/\Omega_x, \sigma^2)$ , we need to determine $\tilde \varepsilon$ from two equally probable hypotheses $H_1: \tilde \varepsilon = \varepsilon$ or $H_2: \tilde \varepsilon = -\varepsilon$. For these problems the likelihood ratio criterion can be used:
\begin{equation}
    \label{crit}
    \delta(\xi_1, \ldots, \xi_N) = 
    \begin{cases}
   H_1, & T(\xi_1,\ldots,\xi_N) < c \\
   H_2, & T(\xi_1,\ldots,\xi_N) \geq c \\
   \end{cases},~~~ T(\xi_1, \ldots, \xi_N) = \frac{f_{H_2}(\xi_1,\ldots,\xi_N)}{f_{H_1}(\xi_1,\ldots,\xi_N)},
\end{equation}
where $f_{H}$ is a density function of a random vector $\xi_1,\ldots,\xi_N$ with distribution from the hypothesis $H$. The Neyman–Pearson lemma yields
\begin{lemma}
There is a constant $c$ for which the likelihood-ratio criterion \eqref{crit} is
\begin{itemize}
    \item minmax criterion. The number $c$ should be chosen so that the type I error and the type II error were the same;
    \item Bayesian criterion for given prior probabilities $r$ and $s$. The number $c$ is chosen equal to the ratio $r/s$. 
\end{itemize}
\end{lemma}
Due to the symmetry of the hypotheses with respect to zero, as well as the fact that the prior probabilities can be considered equal to 1/2, we have that $c = 1$ for minmax and Bayesian criterions. By simple transformations we can rewrite \eqref{crit}:
\begin{equation*}
    \delta(\xi_1, \ldots, \xi_N) = 
    \begin{cases}
   H_1, & \sum\limits_{i=1}^N \xi_i > 0 \\
   H_2, & \sum\limits_{i=1}^N \xi_i \leq 0 \\
   \end{cases},~~~~~~~ \hat x_N = 
    \begin{cases}
   -\frac{\Omega_x}{2}, & \sum\limits_{i=1}^N \xi_i > 0 \\
   \frac{\Omega_x}{2}, & \sum\limits_{i=1}^N \xi_i \leq 0 \\
   \end{cases}.
\end{equation*}
This criterion is more than natural. Neyman–Pearson lemma says it is optimal.
Next we analyse error of this criterion (we will consider only case with $\tilde \varepsilon = \varepsilon$, the other case one can parse similarly):
$$\EE\left[f(\hat x_N)  - f(x^*)\right] = \EE\left[\frac{\varepsilon}{\Omega_x} \hat x_N  + \frac{\varepsilon}{2}\right] = \varepsilon \cdot \PP\left\{\sum\limits_{i=1}^N \xi_i \leq 0\right\} = \varepsilon \cdot \PP\left\{S_N \leq 0\right\},$$
where $S_N = \sum\limits_{i=1}^N \xi_i \in \mathcal{N}(\varepsilon N /\Omega_x, \sigma^2 N)$, then $\frac{S_N - \varepsilon N /\Omega_x}{\sigma \sqrt{N}} \in \mathcal{N}(0, 1)$. Finally, we get
\begin{align*}
\EE\left[f(\hat x_N)  - f(x^*)\right] &=  \varepsilon \PP\left\{\frac{S_N - \varepsilon N /\Omega_x}{\sigma \sqrt{N}} \leq -\frac{\varepsilon\sqrt{N}}{\Omega_x \sigma}\right\}
\\
&= \varepsilon \PP\left\{\ans{- \frac{S_N - \varepsilon N /\Omega_x}{\sigma \sqrt{N}} } \geq \frac{\varepsilon\sqrt{N}}{\Omega_x \sigma}\right\}
\\
&\geq \varepsilon \cdot \frac{1}{3t} \exp\left(-\frac{t^2}{2}\right) \cdot \left( 1 - \frac{1}{t^2} \right).
\end{align*}
In last inequality we define $t = \frac{\varepsilon\sqrt{N}}{\Omega_x \sigma}$ and use lower bound for tail of standard normal distribution.
With $\varepsilon = \frac{2\Omega_x \sigma}{\sqrt{N}}$, we have $t = 2$ and then
$$\EE\left[f(\hat x_N)  - f(x^*)\right] \geq  \frac{\varepsilon}{4t} \exp\left(-2\right)\geq \frac{1}{4 \exp(2)} \cdot \frac{\sigma \Omega_x}{\sqrt{N}}.$$

Hence, we get the next theorem:
\begin{theorem} [Theorem \ref{th4}]
For any $L >0$ and any  $M,T \in \mathbb{N}$, there exists a stochastic minimization problem with $L$-smooth and convex function (i.e., satisfying Assumptions \ref{ass:as1g} and \ref{ass:as2c}) on a bounded $\mathcal{X}^{stoch}$ with a diameter $\Omega_z$ (i.e., satisfying Assumption \ref{as:as4}), such that for any output $\hat x$ of any {\tt BBP}$(T,K)$ (Definition \ref{app:proc}) with $M$ workers, one can obtain the following estimate:
$$\EE\left[f(\hat x)  - f(x^*)\right] =  \Omega\left(\frac{\sigma \Omega_x}{\sqrt{MT}}\right).$$
\end{theorem}

\clearpage

\section{Proof of Theorems from Section  \ref{sec:oa}}
\label{aoa}

\subsection{Centralized case}

We start our proof with the following lemma:

\begin{lemma} 
\label{l1}
Let $z,y \in \mathbb{R}^n$ and $\mathcal{Z} \subset \mathbb{R}^n$ be a convex closed set. We set $z^+ = \text{proj}_{\mathcal{Z}}(z - y)$, then for all $u \in \mathcal{Z}$:
$$\|z^+ - u\|^2 \leq \|z - u \|^2 - 2 \langle y, z^+ - u\rangle - \|z^+ - z \|^2.$$
\end{lemma}
\textbf{Proof:} For all $u \in \mathcal{Z}$ we have $\langle z^+ -(z - y), z^+ - u\rangle \leq 0$. Then
 \begin{align*}
 \|z^+ - u \|^2 &= \|z^+ - z + z - u \|^2 \\
&= \|z-u \|^2 + 2 \langle z^+ - z, z - u \rangle + \|z^+ -z \|^2 \\
&= \|z-u \|^2 + 2 \langle z^+ - z, z^+ - u \rangle - \|z^+ - z \|^2 \\
&= \|z-u \|^2 + 2 \langle z^+ - (z - y), z^+ - u \rangle - 2 \langle y, z^+ - u \rangle - \|z^+ - z \|^2 \\
&\leq \|z - u\|^2 - 2 \langle y, z^+ - u \rangle - \|z^+ - z \|^2.
 \end{align*}
\EndProof

Before proof the main theorems, we add the following notation:
$$\bar g^t = \frac{1}{M}\sum\limits_{m=1}^M g^t_m,~~~~~~\bar g^{t+1/2} = \frac{1}{M}\sum\limits_{m=1}^M g^{t+1/2}_m.$$

\subsubsection{Strongly convex-strongly concave problems}

\begin{theorem}[Theorem \ref{th5}] \label{1th1}
Let $\{ z^t\}_{t \geq 0}$ denote the iterates of Algorithm~\ref{alg1} for solving the problem \eqref{distr}. Let Assumptions \ref{ass:as1g}, \ref{ass:as2g} and \ref{as:as3} be satisfied. Then if $\gamma \leq \frac{1}{4L}$, we have the following estimate:
\begin{equation*}
 \E\left[\| z^{k} - z^* \|^2\right]  = 
 \mathcal{O}\left(\| z^{0} - z^* \|^2 \exp\left( - \frac{\mu }{4L} \cdot\frac{K}{\Delta} \right) + \frac{\sigma^2}{\mu^2 M T}\right) .
 \end{equation*}
\end{theorem}

\textbf{Proof:} Applying the previous Lemma with $z^+ = z^{t+1}$,  $z= z^{t}$, $u = z$ and $y = \gamma \bar g^{t+1/2}$, we get
\begin{equation*}
 \|z^{t+1} - z \|^2 \leq \| z^t - z \|^2 - 2 \gamma \langle \bar g^{t+1/2},  z^{t+1} - z \rangle - \| z^{t+1} -  z^t \|^2,
 \end{equation*}
and with $z^+ =  z^{t+1/2}$,  $z= z^{t}$, $u = z^{t+1}$, $y = \gamma \bar g^{t}$:
\begin{equation*}
 \| z^{t+1/2} -  z^{t+1} \|^2 \leq \| z^t - z^{t+1} \|^2 - 2 \gamma \langle \bar g^{t},  z^{t+1/2} -  z^{t+1} \rangle - \| z^{t+1/2} -  z^t \|^2.
 \end{equation*}
Next, we sum up the two previous equalities
 \begin{align*}
 \| z^{t+1} - z \|^2 + \| z^{t+1/2} -  z^{t+1} \|^2 \leq& \| z^t - z \|^2 - \| z^{t+1/2} -  z^t \|^2 \\
 & - 2 \gamma \langle \bar g^{t+1/2},  z^{t+1} - z \rangle - 2 \gamma \langle \bar g^{t}, z^{t+1/2} -  z^{t+1} \rangle.
 \end{align*}
A small rearrangement gives
\begin{align}
\label{t1}
 \| z^{t+1} - z \|^2 &+ \| z^{t+1/2} -  z^{t+1} \|^2 
 \nonumber\\
 \leq& \| z^t - z \|^2  - \| z^{t+1/2} -  z^t \|^2 \nonumber\\ 
 &- 2 \gamma \langle \bar g^{t+1/2},  z^{t+1/2} - z \rangle  + 2 \gamma \langle \bar g^{t+1/2} - \bar g^{t},  z^{t+1/2} -  z^{t+1} \rangle \nonumber\\
 \leq& \| z^t - z \|^2  - \| z^{t+1/2} -  z^t \|^2 \nonumber\\ 
 & - 2 \gamma \langle \bar g^{t+1/2},  z^{t+1/2} - z \rangle  + \gamma^2 \|\bar g^{t+1/2} - \bar g^{t}\|^2 + \|  z^{t+1/2} -  z^{t+1}\|^2.
 \end{align}
Then we substitute $z = z^*$  and take the total expectation of both sides of the equation
\begin{align}
\label{1}
 \E\left[\| z^{t+1} - z^* \|^2\right]  \leq& \E\left[\| z^t - z^* \|^2\right] - \E\left[\| z^{t+1/2} -  z^t \|^2\right] \notag\\
 &- 2 \gamma \E\left[\langle \bar g^{t+1/2},  z^{t+1/2} - z^* \rangle\right] + \gamma^2 \E\left[\|\bar g^{t+1/2} - \bar g^{t}\|^2\right].  
 \end{align}
Let us work with $\E\left[\|\bar g^{t+1/2} - \bar g^{t}\|^2\right]$:
\begin{eqnarray*}
\E\left[\|\bar g^{t+1/2} - \bar g^{t}\|^2\right]
&=& \E\left[\|\bar g^{t+1/2} - F(z^{t+1/2}) + F(z^{t}) - \bar g^{t} + F(z^{t+1/2}) - F(z^{t}) \|^2\right] \\
&\overset{\eqref{eq:squared_sum}}{\leq}& 3\E\left[\|\bar g^{t+1/2} - F(z^{t+1/2}) \|^2\right] + 3\E\left[\| F(z^{t}) - \bar g^{t} \|^2\right] \\
&&+ 3\E\left[\| F(z^{t+1/2}) - F(z^{t}) \|^2\right] \\
&\overset{\eqref{as1g}}{\leq}& 3\E\left[\left\| \frac{1}{bM} \sum\limits_{m=1}^M \sum\limits_{i=1}^b (F_m(z^{t+1/2}, \xi^{t+1/2,i}_m) -  F_m(z^{t+1/2}))\right\|^2\right] \\
&&+ 3\E\left[\left\| \frac{1}{bM} \sum\limits_{m=1}^M \sum\limits_{i=1}^b (F_m(z^t, \xi^{t,i}_m) -  F_m(z^t))\right\|^2\right] \\
&&+ 3L^2\E\left[\| z^{t+1/2} - z^{t} \|^2\right] \\
&=& \frac{3}{(bM)^2}\E\left[\left\| \sum\limits_{m=1}^M \sum\limits_{i=1}^b (F_m(z^{t+1/2}, \xi^{t+1/2,i}_m) -  F_m(z^{t+1/2}))\right\|^2\right] \\
&&+ \frac{3}{(bM)^2}\E\left[\left\| \sum\limits_{m=1}^M \sum\limits_{i=1}^b (F_m(z^t, \xi^{t,i}_m) -  F_m(z^t))\right\|^2\right] \\
&&+ 3L^2\E\left[\| z^{t+1/2} - z^{t} \|^2\right].
\end{eqnarray*}
Using that all $\{\xi^{t,i}_m\}_{i=1, m=1}^{b,M}$ and $\{\xi^{t+1/2,i}_m\}_{i=1, m=1}^{b,M}$ are independent, we get
\begin{eqnarray}
\label{t432}
\E\left[\|\bar g^{t+1/2} - \bar g^{t}\|^2\right] &\leq& \frac{3}{(bM)^2} \sum\limits_{m=1}^M \sum\limits_{i=1}^b \E\left[\left\|  F_m(z^{t+1/2}, \xi^{t+1/2,i}_m) -  F_m(z^{t+1/2})\right\|^2\right] \nonumber\\
&&+ \frac{3}{(bM)^2} \sum\limits_{m=1}^M \sum\limits_{i=1}^b \E\left[\left\|  F_m(z^t, \xi^{t,i}_m) -  F_m(z^t)\right\|^2\right] \nonumber\\
&&+ 3L^2\E\left[\| z^{t+1/2} - z^{t} \|^2\right]\nonumber\\
&\overset{\eqref{as3}}{\leq}& 3L^2\E\left[\| z^{t+1/2} - z^{t} \|^2\right] + \frac{6 \sigma^2}{bM}.
\end{eqnarray}
Next, we estimate $\E\left[\langle \bar g^{t+1/2},  z^{t+1/2} - z^* \rangle\right]$. To begin with, we use the independence of all $\xi$, as well as the unbiasedness of $\bar g^{t+1/2}$ with respect to the conditional expectation by the random variables $\{\xi^{t+1/2,i}_m\}_{i=1, m=1}^{b,M}$:
\begin{align}
\label{t505}
\E\left[\langle \bar g^{t+1/2},  z^{t+1/2} - z^* \rangle\right] 
&= \E\left[\E_{\{\xi^{t+1/2,i}_m\}_{i=1, m=1}^{b,M}}\left[\langle \bar g^{t+1/2},  z^{t+1/2} - z^* \rangle\right]\right] \nonumber\\
&= \E\left[\langle \E_{\{\xi^{t+1/2,i}_m\}_{i=1, m=1}^{b,M}}\left[\bar g^{t+1/2}\right],  z^{t+1/2} - z^* \rangle\right]
\nonumber\\
&= \E\left[\langle F(z^{t+1/2}),  z^{t+1/2} - z^* \rangle\right].
\end{align}
By the property of the solution $z^*$, we get
\begin{align*}
\E\left[\langle \bar g^{t+1/2},  z^{t+1/2} - z^* \rangle\right] &\geq \E\left[\langle F(z^{t+1/2}) -F(z^*),  z^{t+1/2} - z \rangle\right] \\
&\geq  \mu\E\left[\|z^{t+1/2} - z^*\|^2\right].
\end{align*}
Let us use a simple fact $\|z^{t+1/2} - z^*\|^2   \geq \frac{1}{2}\|z^t - z^*\|^2 - \|z^{t+1/2} - z^t\|^2$, then
\begin{align}
\label{t4321}
\E\left[\langle \bar g^{t+1/2},  z^{t+1/2} - z^* \rangle\right] \geq  \frac{\mu}{2}\E\left[\|z^t - z^*\|^2 \right] - \mu \E\left[\|z^{t+1/2} - z^t\|^2 \right].
\end{align}
Combining three inequalities: \eqref{1} with $z = z^*$, \eqref{t432}, \eqref{t4321}, we obtain:
\begin{align*}
 \E\left[\| z^{t+1} - z^* \|^2\right]  &\leq (1 - \mu \gamma)\E\left[\| z^t - z^* \|^2\right] \\
 &\hspace{0.4cm}+(2\mu\gamma + 3\gamma^2 L^2 -1) \E\left[\| z^{t+1/2} -  z^t \|^2\right] + \frac{6 \sigma^2 \gamma^2}{bM}.  
 \end{align*}
In Algorithm \ref{alg1} the step $\gamma \leq \frac{1}{4L}$, then 
\begin{align*}
 \E\left[\| z^{t+1} - z^* \|^2\right]  &\leq (1 - \mu \gamma)\E\left[\| z^t - z^* \|^2\right]  + \frac{6 \sigma^2 \gamma^2}{bM}.
 \end{align*}
 Let us run the recursion from $0$ to $k-1$:
 \begin{align*}
 \E\left[\| z^{k} - z^* \|^2\right]  &\leq (1 - \mu \gamma)^k\E\left[\| z^0 - z^* \|^2\right]  + \frac{6 \sigma^2 \gamma}{\mu bM}.
 \end{align*}
Then we 
carefully choose $\gamma = \min\left\{\frac{1}{4L}; \frac{\ln(\max\{2; b M \mu^2\|z^0 - z^*\|^2 k / 6\sigma^2\})}{\mu k}\right\}$ and get (for more details one can see \cite{stich2019unified})
 \begin{align*}
 \E\left[\| z^{k+1} - z^* \|^2\right]  = 
 \mathcal{\tilde O}\left(\| z^{0} - z^* \|^2 \exp\left( - \frac{\mu k}{4L} \right) + \frac{\sigma^2}{\mu^2 bM k}\right) .
 \end{align*}
Substitute the batch size $b$ and the number of iterations $k$ from the description of Algorithm \ref{alg1}:
\begin{align*}
 \E\left[\| z^{k+1} - z^* \|^2\right]  = 
 \mathcal{\tilde O}\left(\| z^{0} - z^* \|^2 \exp\left( - \frac{\mu }{4L} \cdot\frac{K}{r} \right) + \frac{\sigma^2}{\mu^2 M T}\right) .
 \end{align*}
Finally, we remember that $r \leq \Delta$ and finish the proof.
\EndProof

\subsubsection{Convex-concave problems}
\begin{theorem}[Theorem \ref{th5}] \label{t13}
Let $\{ z^t\}_{t \geq 0}$ denote the iterates of Algorithm~\ref{alg1} for solving the problem \eqref{distr}. Let Assumptions \ref{ass:as1g}, \ref{ass:as2c}, \ref{as:as3} and \ref{as:as4} be satisfied. Then if $\gamma \leq \frac{1}{4L}$, we have the following estimate:
\begin{equation*}
    \E[\mathrm{gap}(z^{k}_{avg})] = \mathcal{O}\left(\frac{L \Omega_z^2 \Delta}{K} + \frac{\sigma \Omega_z}{\sqrt{MT}} \right).
\end{equation*}
\end{theorem}

\textbf{Proof:} 
We have already shown some of the necessary estimates, namely, we need to use \eqref{t1} with some small rearrangement
\begin{align*}
2 \gamma \langle F(z^{t+1/2}),  z^{t+1/2} - z \rangle  \leq& \| z^t - z \|^2 - \| z^{t+1} - z \|^2   - \| z^{t+1/2} -  z^t \|^2 \nonumber\\ 
 &+ 2 \gamma \langle F(z^{t+1/2}) - \bar g^{t+1/2},  z^{t+1/2} - z \rangle + \gamma^2 \|\bar g^{t+1/2} - \bar g^{t}\|^2.
 \end{align*}
 Next, we sum over all $t$ from $0$ to $k-1$
\begin{align}
\label{temp7}
\frac{1}{k}\sum\limits_{t=0}^{k-1}\langle F(z^{t+1/2}),  z^{t+1/2} - z \rangle 
\leq& \frac{\| z^0 - z \|^2 - \| z^{k+1} - z \|^2}{2\gamma k} \nonumber\\
&+ \frac{1}{k}\sum\limits_{t=0}^{k-1} \langle F(z^{t+1/2}) - \bar g^{t+1/2},  z^{t+1/2} - z \rangle \nonumber\\
&+ \frac{1}{2\gamma k}\sum\limits_{t=0}^{k-1} \gamma^2 \|\bar g^{t+1/2} - \bar g^{t}\|^2 - \| z^{t+1/2} -  z^t \|^2.
 \end{align}
Then by $x^{k}_{avg} = \frac{1}{k}\sum_{t=0}^{k-1} x^{t+1/2}$ and $y^{k}_{avg} = \frac{1}{k}\sum_{t=0}^{k-1} y^{t+1/2}$, the Jensen's inequality and convexity-concavity of $f$:
\begin{align*}
    \mathrm{gap}(z^{k}_{avg})
    \leq& \max\limits_{y' \in \mathcal{Y}} f\left(\frac{1}{k} \left(\sum^{k-1}_{t=0} x^{t+1/2} \right), y'\right) - \min\limits_{x' \in \mathcal{X}} f\left(x', \frac{1}{k} \left(\sum^{k-1}_{t=0} y^{t+1/2} \right)\right) 
    \nonumber \\
    \leq& \max\limits_{y' \in \mathcal{Y}} \frac{1}{k} \sum^{k-1}_{t=0} f(x^{t+1/2}, y')  - \min\limits_{x' \in \mathcal{X}} \frac{1}{k} \sum^{k-1}_{t=0} f(x', y^{t+1/2}).
\end{align*}
Given the fact of linear independence of $x'$ and $y'$:
\begin{equation*}
    \mathrm{gap}(z^{k}_{avg}) \leq \max\limits_{(x', y') \in \mathcal{Z}}\frac{1}{k} \sum^{k-1}_{t=0}  \left(f(x^{t+1/2}, y')  - f(x', y^{t+1/2}) \right).
\end{equation*}
Using convexity and concavity of the function $f$:
\begin{align}
\label{temp8}
    \mathrm{gap}(z^{k}_{avg}) &\leq  \max\limits_{(x', y') \in \mathcal{Z}}\frac{1}{k} \sum^{k-1}_{t=0}  \left(f(x^{t+1/2}, y')  - f(x', y^{t+1/2}) \right)  \nonumber \\
    &=  \max\limits_{(x', y') \in \mathcal{Z}} \frac{1}{k} \sum^{k-1}_{t=0} \Big(f(x^{t+1/2}, y') -f(x^{t+1/2}, y^{t+1/2})  \nonumber \\
    &\hspace{3cm}+ f(x^{t+1/2}, y^{t+1/2}) - f(x', y^{t+1/2}) \Big) \nonumber \\
    &\leq \max\limits_{(x', y') \in \mathcal{Z}} \frac{1}{k} \sum^{k-1}_{t=0} \Big(\langle \nabla_y f (x^{t+1/2}, y^{t+1/2}), y'-y^{t+1/2} \rangle 
    \nonumber \\
    &\hspace{3cm}
    + \langle \nabla_x f (x^{t+1/2}, y^{t+1/2}), x^{t+1/2}-x' \rangle \Big) \nonumber \\
    &\leq \max\limits_{z \in \mathcal{Z}} \frac{1}{k} \sum^{k-1}_{t=0} \langle  F(z^{t+1/2}), z^{t+1/2} - z\rangle.
\end{align}
Together with \eqref{temp8}, \eqref{temp7} gives (additionally, we take the full expectation)
\begin{eqnarray*}
\E[\mathrm{gap}(z^{k}_{avg})] &\leq&  \E\left[ \max\limits_{z \in \mathcal{Z}}\frac{\| z^0 - z \|^2 -\| z^{k} - z \|^2}{2\gamma k}\right] \nonumber\\ 
&&+ \frac{1}{k}  \E\left[ \max\limits_{z \in \mathcal{Z}}\sum\limits_{t=0}^{k-1} \langle F(z^{t+1/2}) - \bar g^{t+1/2},  z^{t+1/2} - z \rangle\right] \nonumber\\
&&+ \frac{1}{2\gamma k} \E\left[\sum\limits_{t=0}^{k-1} \gamma^2 \|\bar g^{t+1/2} - \bar g^{t}\|^2 - \| z^{t+1/2} -  z^t \|^2\right] \nonumber\\
&\overset{\eqref{as5},\eqref{t432}}{\leq}& \frac{\Omega_z^2}{2\gamma k}+ \frac{1}{k}  \E\left[ \max\limits_{z \in \mathcal{Z}}\sum\limits_{t=0}^{k-1} \langle F(z^{t+1/2}) - \bar g^{t+1/2},  z^{t+1/2} - z \rangle\right] \nonumber\\
&&+ \frac{1}{2\gamma k} \E\left[\sum\limits_{t=0}^{k-1}  3\gamma^2L^2\| z^{t+1/2} - z^{t} \|^2 + \frac{6 \gamma^2\sigma^2}{bM} - \| z^{t+1/2} -  z^t \|^2\right].
 \end{eqnarray*}
 With $\gamma = \frac{1}{4L}$, we get
 \begin{align}
 \label{o303}
\E[\mathrm{gap}(z^{k}_{avg})] \leq  \frac{\Omega_z^2}{2\gamma k}+ \frac{1}{k}  \E\left[ \max\limits_{z \in \mathcal{Z}}\sum\limits_{t=0}^{k-1} \langle F(z^{t+1/2}) - \bar g^{t+1/2},  z^{t+1/2} - z \rangle\right] +\frac{3 \gamma\sigma^2}{bM}.
 \end{align}
To finish the proof we need to estimate $\E\left[ \max\limits_{z \in \mathcal{Z}}\sum\limits_{t=0}^{k-1} \langle F(z^{t+1/2}) - \bar g^{t+1/2},  z^{t+1/2} - z \rangle\right]$. Let us define the sequence $v$: $v^0 = z^{1/2}$, $v^{t+1} = \text{proj}_{\mathcal{Z}}(v^t-\gamma \delta_t)$ with $\delta^t = F(z^{t+1/2}) - \bar g^{t+1/2}$. Then we have
\begin{equation}
    \label{t4441}
    \sum\limits_{t=0}^{k-1} \langle \delta^t, z^{t+1/2} - z \rangle = \sum\limits_{t=0}^{k-1} \langle \delta^t, z^{t+1/2} - v^t \rangle + 
    \sum\limits_{t=0}^{k-1} \langle \delta^t,  v^t - z \rangle . 
\end{equation}
By the definition of $v^{t+1}$, we have for all $z \in \mathcal{Z}$
\begin{equation*}
    \langle v^{t+1} - v^{t} +\gamma\delta^t, z - v^{t+1} \rangle \geq 0.
\end{equation*}
Rewriting this inequality, we get
\begin{align*}
    \langle \gamma\delta^t, v^t  - z \rangle \leq& \langle \gamma\delta^t, v^t - v^{t+1} \rangle  + \langle v^{t+1} - v^t, z - v^{t+1} \rangle \nonumber\\
\leq& \langle \gamma\delta^t, v^t - v^{t+1} \rangle + \frac{1}{2}\|v^t - z\|^2 -  \frac{1}{2}\|v^{t+1} - z\|^2 - \frac{1}{2}\| v^t - v^{t+1}\|^2
\nonumber\\
\leq& \frac{\gamma^2}{2} \|\delta^t\|^2  + \frac{1}{2}\|v^t - v^{t+1}\|^2 + \frac{1}{2}\|v^t - z\|^2 \\
&-  \frac{1}{2}\|v^{t+1} - z\|^2 - \frac{1}{2}\| v^t - v^{t+1}\|^2
\nonumber\\
=& \frac{\gamma^2}{2} \|\delta^t\|^2  + \frac{1}{2}\|v^t - z\|^2 -  \frac{1}{2}\|v^{t+1} - z\|^2 .
\end{align*}
With \eqref{t4441}, it gives
\begin{eqnarray*}
    \sum\limits_{t=0}^{k-1} \langle \delta^t, z^{t+1/2} - z \rangle &\leq& \sum\limits_{t=0}^{k-1} \langle \delta^t, z^{t+1/2} - v^t \rangle \\
    &&+ 
    \frac{1}{\gamma}\sum\limits_{t=0}^{k-1} \left(\frac{\gamma^2}{2} \|\delta^t\|^2  + \frac{1}{2}\|v^t - z\|^2 -  \frac{1}{2}\|v^{t+1} - z\|^2 \right) \nonumber\\
    &\leq& \sum\limits_{t=0}^{k-1} \langle \delta^t, z^{t+1/2} - v^t \rangle + 
    \frac{\gamma}{2}\sum\limits_{t=0}^{k-1} \|\delta^t\|^2 + \frac{1}{2\gamma}\|v^0 - z\|^2 
    \nonumber\\
    &\leq& \sum\limits_{t=0}^{k-1} \langle \delta^t, z^{t+1/2} - v^t \rangle + 
    \frac{\gamma}{2}\sum\limits_{t=0}^{k-1} \|\delta^t\|^2 + \frac{\Omega_z^2}{2\gamma}, 
\end{eqnarray*}
\ans{where in the last inequality we used Assumption \ref{as:as4} with $v^0, z \in \mathcal{Z}$.} The right side is independent of $z$, then
\begin{align}
    \label{t561}
     \max_{z \in \mathcal{Z}} \sum\limits_{t=0}^{k-1} \langle \delta^t, z^{t+1/2} - z \rangle \leq& \sum\limits_{t=0}^{k-1} \langle \delta^t, z^{t+1/2} - v^t \rangle \nonumber\\
     &+ 
    \frac{\gamma}{2}\sum\limits_{t=0}^{k-1} \|F(z^{t+1/2}) - \bar g^{t+1/2}\|^2 + \frac{\Omega_z^2}{2\gamma}.  
\end{align}
Taking the full expectation and using independence $v^t - z^{t+1/2} $, $\{\xi^{t+1/2,i}_m\}_{i=1, m=1}^{b,M}$, we get
\begin{eqnarray*}
     &&\hspace{-2cm}\EE\left[ \max_{z \in \mathcal{Z}} \sum\limits_{t=0}^{k-1} \langle \delta^t, z^{t+1/2} - z \rangle\right] \\
     &\leq& \EE\left[\sum\limits_{t=0}^{k-1} \langle \delta^t, z^{t+1/2} - v^t \rangle\right] + 
    \frac{\gamma}{2}\sum\limits_{t=0}^{k-1} \EE\left[\|F(z^{t+1/2}) - \bar g^{t+1/2}\|^2 \right]+ \frac{\Omega_z^2}{2\gamma} \nonumber\\
    &=& \EE\left[\sum\limits_{t=0}^{k-1} \langle \EE_{\{\xi^{t+1/2,i}_m\}_{i=1, m=1}^{b,M}}\left[F(z^{t+1/2}) - \bar g^{t+1/2}\right], z^{t+1/2} - v^t \rangle\right] \nonumber\\
    &&+ 
    \frac{\gamma}{2}\sum\limits_{t=0}^{k-1} \EE\left[\|F(z^{t+1/2}) - \bar g^{t+1/2}\|^2 \right]+ \frac{\Omega_z^2}{2\gamma}
    \nonumber\\
    &=& \frac{\gamma}{2}\sum\limits_{t=0}^{k-1} \EE\left[\|F(z^{t+1/2}) - \bar g^{t+1/2}\|^2 \right]+ \frac{\Omega_z^2}{2\gamma} \nonumber\\
    &\overset{\eqref{t432}}{\leq}& \frac{\gamma k}{2} \cdot \frac{3 \sigma^2}{bM} + \frac{\Omega_z^2}{2\gamma}.  
\end{eqnarray*}
Then we can finish \eqref{o303} and get
\begin{equation*}
\E[\mathrm{gap}(z^{k}_{avg})] \leq  \frac{\Omega_z^2}{\gamma k}+ \frac{\gamma}{2} \cdot \frac{5 \sigma^2}{bM}.
 \end{equation*}
Let $\gamma = \min\left\{\frac{1}{4L}; \Omega_z\sqrt{\frac{2bM}{5k\sigma^2}}\right\}$, then
\begin{equation*}
    \E[\mathrm{gap}(z^{k}_{avg})] = \mathcal{O}\left(\frac{L \Omega_z^2}{k} + \frac{\sigma \Omega_z}{\sqrt{bMk}} \right).
\end{equation*}
Substitute the batch size $b$ and the number of iterations $k$ from the description of Algorithm \ref{alg1} with $r \leq \Delta$:
\begin{equation*}
    \E[\mathrm{gap}(z^{k}_{avg})] = \mathcal{O}\left(\frac{L \Omega_z^2 \Delta}{K} + \frac{\sigma \Omega_z}{\sqrt{MT}} \right).
\end{equation*}
\EndProof

\subsubsection{Non-convex-non-concave problems}
\begin{theorem}[Theorem \ref{th5}]\label{t131}
Let $\{ z^t\}_{t \geq 0}$ denote the iterates of Algorithm~\ref{alg1} for solving the problem \eqref{distr}. Let Assumptions \ref{ass:as1g}, \ref{ass:as2n}, \ref{as:as3} and \ref{as:as4} be satisfied. Then if $\gamma \leq \frac{1}{4L}$, we have the following estimate:
\begin{equation*}
    \E\left[ \frac{1}{k}\sum\limits_{t=0}^{k-1}\|F(z^t)\|^2\right] = \mathcal{O}\left(\frac{L^2 \Omega_z^2 \Delta}{K} + \frac{\sigma^2 K}{MT\Delta} \right).
\end{equation*}
\end{theorem}

\textbf{Proof:} We start proof with combining \eqref{1} ($z = z^*$), \eqref{t432} and \eqref{t505} 
\begin{align*}
 \E\left[\| z^{t+1} - z^* \|^2\right]  \leq& \E\left[\| z^t - z^* \|^2\right] - \E\left[\| z^{t+1/2} -  z^t \|^2\right] \notag\\
 &- 2 \gamma \E\left[\langle F(z^{t+1/2}),  z^{t+1/2} - z^* \rangle\right] \\
 &+  3 \gamma^2 L^2\E\left[\| z^{t+1/2} - z^{t} \|^2\right] + \frac{6 \gamma^2 \sigma^2}{bM}.  
 \end{align*}
 Using the Minty assumption \eqref{as2m}, we obtain
 \begin{align*}
 \E\left[\| z^{t+1} - z^* \|^2\right]  &\leq \E\left[\| z^t - z^* \|^2\right] - (1 - 3 \gamma^2 L^2)\E\left[\| z^{t+1/2} -  z^t \|^2\right]  + \frac{6 \gamma^2 \sigma^2}{bM} \notag\\
 &= \E\left[\| z^t - z^* \|^2\right] - \gamma^2 (1 - 3 \gamma^2 L^2) \E\left[\left\| \bar g^t \right\|^2\right]  + \frac{6 \gamma^2 \sigma^2}{bM}.  
 \end{align*}
 With $\gamma \leq \frac{1}{4L}$, we get
  \begin{align*}
 \E\left[\| z^{t+1} - z^* \|^2\right]  &\leq \E\left[\| z^t - z^* \|^2\right] - (1 - 3 \gamma^2 L^2)\E\left[\| z^{t+1/2} -  z^t \|^2\right]  + \frac{6 \gamma^2 \sigma^2}{bM} \notag\\
 &= \E\left[\| z^t - z^* \|^2\right] - \frac{3\gamma^2}{4}\E\left[\left\| \bar g^t \right\|^2\right]  + \frac{6 \gamma^2 \sigma^2}{bM}.  
 \end{align*}
 The fact: $-\| \bar g^t\|^2 \leq -\frac{2}{3} \| F(z^t)\|^2 + 2\| \bar g^t - F(z^t)\|^2 $, gives
 \begin{equation*}
 \E\left[\| z^{t+1} - z^* \|^2\right]  \leq  \E\left[\| z^t - z^* \|^2\right] - \frac{\gamma^2}{2}\E\left[\left\| F(z^t) \right\|^2\right] +  2\gamma^2\| \bar g^t - F(z^t)\|^2 + \frac{6 \gamma^2 \sigma^2}{bM}.  
 \end{equation*}
The term $\| \bar g^t - F(z^t)\|^2$ was estimated, when we deduced \eqref{t432}. Then
 \begin{equation*}
 \frac{\gamma^2}{2}\E\left[\left\| F(z^t) \right\|^2\right]  \leq  \E\left[\| z^t - z^* \|^2\right] - \E\left[\| z^{t+1} - z^* \|^2\right] +  \frac{8 \gamma^2 \sigma^2}{bM}.  
 \end{equation*}
Summing over all $t$ from $0$ to $k-1$:
\begin{equation*}
\E\left[\frac{1}{k} \sum\limits_{t=0}^{k-1}\left\|  F(z^t) \right\|^2\right]  \leq \frac{2\E\left[\| z^0 - z^* \|^2\right]}{\gamma^2 k} +  \frac{16 \sigma^2}{bM}.  
 \end{equation*}
Next we substitute $\gamma = \frac{1}{4L}$, $k$, $b$ and finish the proof. 
\EndProof

\subsection{Decentralized case}

First of all, we present the missing Algorithm \ref{alg3}:

\begin{algorithm} [th]
	\caption{FastMix}
	\label{alg3}
	\begin{algorithmic}
\STATE
\noindent {\bf Parameters:} Vectors $z_1, ..., z_M$, communic. rounds $P$.\\
\STATE \noindent {\bf Initialization:}
Construct matrix $\textbf{z}$ with rows $z^T_1, ..., z^T_M$,
\STATE choose $\textbf{z}^{-1}=\textbf{z}$, $\textbf{z}^0 = \textbf{z}$, $\eta = \frac{1 - \sqrt{1 - \lambda_2^2(\tilde W)}}{1 + \sqrt{1 - \lambda_2^2(\tilde W)}}$.
\FOR {$h=0,1, 2, \ldots, P-1$ } 
\STATE $\textbf{z}^{h+1} = (1 + \eta)\tilde W \textbf{z}^{h} - \eta \textbf{z}^{h-1}$,
\ENDFOR
\STATE
\noindent {\bf Output:} rows $z_1, ..., z_M$ of  $\textbf{z}^{P}$ .
	\end{algorithmic}
\end{algorithm}

We introduce the following notation:
$$z^{t} = \frac{1}{M} \sum\limits_{m=1}^M  z^{t}_m,~~~~  z^{t+1/2} = \frac{1}{M} \sum\limits_{m=1}^M  z^{t+1/2}_m,
$$
$$\hat z^{t} = \frac{1}{M} \sum\limits_{m=1}^M \hat z^{t}_m,~~~~ \hat z^{t+1/2} = \frac{1}{M} \sum\limits_{m=1}^M \hat z^{t+1/2}_m, ~~~~~\tilde z^{t} = \frac{1}{M} \sum\limits_{m=1}^M \tilde z^{t}_m,~~~~ \tilde z^{t+1/2} = \frac{1}{M} \sum\limits_{m=1}^M \tilde z^{t+1/2}_m,$$
$$
\bar g^{t} = \frac{1}{M} \sum\limits_{m=1}^M  g^{t}_m \ans{= \frac{1}{M} \sum\limits_{m=1}^M \left[\frac{1}{b}\sum\limits_{i=1}^b F_m(z^t_m, \xi^{t,i}_m)\right]},
$$
$$
\bar g^{t+1/2} = \frac{1}{M} \sum\limits_{m=1}^M  g^{t+1/2}_m \ans{= \frac{1}{M} \sum\limits_{m=1}^M \left[ \frac{1}{b}\sum\limits_{i=1}^b F_m(z^{t+1/2}_m, \xi^{t+1/2,i}_m) \right]}.
$$
Next, we introduce the convergence of  FastMix \cite{liu2011accelerated,ye2020multi}:

\begin{lemma} \label{l8} Let $\{\tilde z^{t+1}_m\}^M_{m=1}$ be the output of Algorithm \ref{alg3} with the input $\{\hat z^{t+1}_m\}^M_{m=1}$. Then it holds that
  $$\frac{1}{M} \sum\limits_{m=1}^M\|\tilde z^{t+1}_m - \tilde z^{t+1}\|^2 \leq \left(1 - \frac{1}{\sqrt{\chi}}\right)^{2P} \left(\frac{1}{M} \sum\limits_{m=1}^M\|\hat z^{t+1}_m - \hat z^{t+1}\|^2\right)~~~~\text{and}~~~~\hat z^t = \tilde z^t.$$
\end{lemma}
Let after $P$ iterations we get $\varepsilon_0$-accuracy of consensus, i.e.
\begin{equation}
\label{delt}
    \tilde z^{t}_m - \tilde z^{t} = \delta^t_m, ~~~~~~ \|\delta^t_m\| \leq \varepsilon_0,~~~~~~~~ \tilde z^{t+1/2}_m - \tilde z^{t+1/2} = \delta^{t+1/2}_m, ~~~~~~ \|\delta^{t+1/2}_m\| \leq \varepsilon_0.
\end{equation}
Then let us estimate the number of iterations $P$ to achieve such $\varepsilon_0$ (how to choose this parameter we will talk later) accuracy:

\begin{corollary} \label{cor45}
To achieve $\varepsilon_0$-accuracy in terms of \eqref{delt} we need to take $P$:

$\bullet$ in the convex-concave (Assumptions \ref{ass:as2c} and \ref{as:as4}) and non-convex-non-concave (Assumptions \ref{ass:as2n} and \ref{as:as4}) cases
$$P = \mathcal{O}\left( \sqrt{\chi} \log\left( 1 + \frac{\Omega_z^2 + \frac{Q^2 + \sigma^2/b}{L^2_{\max}}}{\varepsilon_0^2} \right)\right),$$

$\bullet$ in the strongly convex-strongly concave case (Assumption \ref{ass:as2g})
$$P = \mathcal{O}\left( \sqrt{\chi} \log\left( 1 + \frac{\|z^0 - z^*\|^2 + \frac{Q^2 + \sigma^2/b}{L^2_{\max}}}{\varepsilon_0^2} \right)\right),$$
where $Q^2 = \frac{1}{M} \sum\limits_{m=1}^M \|F_m(z^{*})\|^2$.
\end{corollary}

\textbf{Proof: } The proof is in a rough estimate of $\frac{1}{M} \sum\limits_{m=1}^M\|\hat z^{t+1}_m - \hat z^{t+1}\|^2$:

\begin{align*}
\frac{1}{M} \sum\limits_{m=1}^M\|\hat z^{t+1}_m - \hat z^{t+1}\|^2 =& \frac{1}{M} \sum\limits_{m=1}^M\|z_m^{t} - \gamma g^{t+1/2}_m - z^{t} + \gamma \bar g^{t+1/2}\|^2 \\
\leq& \frac{2}{M} \sum\limits_{m=1}^M\|z_m^{t} - z^{t} \|^2 + \frac{2\gamma^2}{M} \sum\limits_{m=1}^M\|g^{t+1/2}_m - \bar g^{t+1/2}\|^2
\\
\leq& \frac{2}{M} \sum\limits_{m=1}^M\|\text{proj}_{\mathcal{Z}}(\tilde z_m^{t}) - \frac{1}{M} \sum\limits_{i=1}^M\text{proj}_{\mathcal{Z}}(\tilde z_i^{t}) \|^2 \\
&+ \frac{2\gamma^2}{M} \sum\limits_{m=1}^M\|g^{t+1/2}_m\|^2
\end{align*}
In the last inequality we use the property: $ \frac{1}{M}\sum\limits_{m=1}^M\|g^{t+1/2}_m - \bar g^{t+1/2}\|^2 \ans{= \frac{1}{M}\sum\limits_{m=1}^M\|g^{t+1/2}_m - \frac{1}{M}\sum\limits_{m=1}^M g^{t+1/2}_m \|^2} \leq \frac{1}{M} \sum\limits_{m=1}^M\|g^{t+1/2}_m\|^2$. Then we take the full expectation and get
\begin{eqnarray*}
\mathbb{E}\left[\frac{1}{M} \sum\limits_{m=1}^M\|\hat z^{t+1}_m - \hat z^{t+1}\|^2 \right] &\overset{\eqref{eq:squared_sum}}{\leq}& \frac{4}{M} \mathbb{E}\left[ \sum\limits_{m=1}^M\|\text{proj}_{\mathcal{Z}}(\tilde z_m^{t}) - \text{proj}_{\mathcal{Z}}(\tilde z^{t}) \|^2\right] \\
&&+ 4\mathbb{E}\left[\|\text{proj}_{\mathcal{Z}}(\tilde z^{t}) - \frac{1}{M} \sum\limits_{i=1}^M\text{proj}_{\mathcal{Z}}(\tilde z_i^{t}) \|^2\right] \\
&&+ \frac{2\gamma^2}{M} \mathbb{E}\left[\sum\limits_{m=1}^M\|g^{t+1/2}_m\|^2\right] \\
&\overset{\eqref{proj},\eqref{eq:squared_sum}}{\leq}& \frac{8}{M} \mathbb{E}\left[\sum\limits_{m=1}^M\|\tilde z_m^{t} - \tilde z^{t}\|^2\right] + \frac{2\gamma^2}{M} \mathbb{E}\left[\sum\limits_{m=1}^M\|g^{t+1/2}_m\|^2 \right]
\\
&\overset{\eqref{delt},\eqref{as4}}{\leq}& 8\varepsilon^2_0 + \frac{4\gamma^2}{M} \mathbb{E}\left[ \sum\limits_{m=1}^M \|F_m(z^{t+1/2})\|^2 \right] + \frac{4\gamma^2\sigma^2}{b}
\\
&\leq& 8\varepsilon^2_0 + \frac{8\gamma^2}{M} \mathbb{E}\left[\sum\limits_{m=1}^M\|F_m(z^{t+1/2}) - F_m(z^{*})\|^2\right] \\ 
&&+ \frac{8\gamma^2}{M} \sum\limits_{m=1}^M\|F_m(z^{*})\|^2 + \frac{4\gamma^2\sigma^2}{b}
\\
&\overset{\eqref{as1l}}{\leq}& 8\varepsilon^2_0 + 8\gamma^2 L^2_{\max} \mathbb{E}\left[\|z^{t+1/2} - z^{*}\|^2 \right]\\
&&+ \frac{8\gamma^2}{M} \sum\limits_{m=1}^M\|F_m(z^{*})\|^2 + \frac{4\gamma^2\sigma^2}{b}
\end{eqnarray*}
The proof of the theorem follows from $\gamma \leq \frac{1}{4 L_{\max}}$ and the fact that in the convex-concave and non-convex-non-concave cases we can bounded $\|z^{t+1/2} - z^{*}\| \leq \Omega_z$, in the strongly convex-strongly concave cases -- $\ans{\mathbb{E}\left[\|z^{t+1/2} - z^{*}\|^2\right] \leq \|z^{0} - z^{*}\|^2}$.

\EndProof

We are now ready to prove the main theorems. Note we can rewrite one step of the algorithm as follows:
\begin{align*}
z^{t+1/2} &= \frac{1}{M} \sum\limits_{m=1}^M z^{t+1/2}_m = \frac{1}{M} \sum\limits_{m=1}^M \text{proj}_{\mathcal{Z}} (\tilde z^{t+1/2} + \delta^{t+1/2}_m) \nonumber\\
&= \text{proj}_{\mathcal{Z}} (\tilde z^{t+1/2}) + \frac{1}{M} \sum\limits_{m=1}^M \text{proj}_{\mathcal{Z}} (\tilde z^{t+1/2} + \delta^{t+1/2}_m) - \text{proj}_{\mathcal{Z}} (\tilde z^{t+1/2})
\nonumber\\
&\ans{= \text{proj}_{\mathcal{Z}} \left(\frac{1}{M}\sum\limits_{m=1}^M \tilde z^{t+1/2}_m \right) + \frac{1}{M} \sum\limits_{m=1}^M \text{proj}_{\mathcal{Z}} (\tilde z^{t+1/2} + \delta^{t+1/2}_m) - \text{proj}_{\mathcal{Z}} (\tilde z^{t+1/2}).}
\end{align*}
\ans{Next, we use that the mixing procedure (in particular FastMix) does not change the sum of the local vectors. In particular, $\frac{1}{M}\sum\limits_{m=1}^M \tilde z^{t+1/2}_m = \frac{1}{M}\sum\limits_{m=1}^M \hat z^{t+1/2}_m $, it gives}
\begin{align*}
z^{t+1/2} &= \ans{\text{proj}_{\mathcal{Z}} \left(\frac{1}{M}\sum\limits_{m=1}^M \hat z^{t+1/2}_m \right) + \frac{1}{M} \sum\limits_{m=1}^M \text{proj}_{\mathcal{Z}} (\tilde z^{t+1/2} + \delta^{t+1/2}_m) - \text{proj}_{\mathcal{Z}} (\tilde z^{t+1/2})}
\nonumber\\
&= \text{proj}_{\mathcal{Z}} \left(\frac{1}{M}\sum\limits_{m=1}^M z^t_m - \gamma g^t_m\right) + \Delta^t
= \text{proj}_{\mathcal{Z}} \left(z^t - \gamma \bar g^t\right) + \Delta^t.
\end{align*}
Here we added one more notation: $\Delta^{t+1/2} := \frac{1}{M} \sum\limits_{m=1}^M \text{proj}_{\mathcal{Z}} (\tilde z^{t+1/2} + \delta^{t+1/2}_m) - \text{proj}_{\mathcal{Z}} (\tilde z^{t+1/2})$ and $\Delta^{t} := \frac{1}{M} \sum\limits_{m=1}^M \text{proj}_{\mathcal{Z}} (\tilde z^{t} + \delta^{t}_m) - \text{proj}_{\mathcal{Z}} (\tilde z^{t})$. It is easy to see that $\|\Delta^{t+1/2} \| \leq \varepsilon_0$ and $\|\Delta^{t} \| \leq \varepsilon_0$.
We see that the step of the algorithm is very similar to the step of Algorithm \ref{alg1}, but with imprecise projection onto a set. Let us prove the following lemma:
\begin{lemma} 
\label{l24}
Let $\mathcal{Z} \subset \mathbb{R}^n$ be a convex compact set, \ans{$z \in \mathcal{Z}$} and $y,\Delta \in \mathbb{R}^n$. We set $z^+ = \text{proj}_{\mathcal{Z}}(z - y) + \Delta$, then for all $u \in \mathcal{Z}$:
$$\|z-u \|^2 + 2\|\Delta\| \cdot \| z^+ - u\| + 4 \|\Delta\| \cdot \| y  \| - 2 \langle y, z^+ - u \rangle - \|z^+ - z \|^2.$$
\end{lemma}
\textbf{Proof:} Let $r = \text{proj}_{\mathcal{Z}}(z - y) $. For all $u \in \mathcal{Z}$ we have $\langle r-(z - y), r - u\rangle \leq 0$. Then
 \begin{eqnarray*}
 \|z^+ - u \|^2 &=& \|z^+ - z + z - u \|^2 \\
&=& \|z-u \|^2 + 2 \langle z^+ - z, z - u \rangle + \|z^+ -z \|^2 \\
&=& \|z-u \|^2 + 2 \langle z^+ - z, z^+ - u \rangle - \|z^+ - z \|^2 \\
&=& \|z-u \|^2 + 2 \langle z^+ - (z - y), z^+ - u \rangle - 2 \langle y, z^+ - u \rangle - \|z^+ - z \|^2 \\
&=& \|z-u \|^2 + 2 \langle r - (z - y), r - u \rangle + 2 \langle \Delta, r - u \rangle + 2 \langle z^+ - (z - y), \Delta \rangle \\
&&- 2 \langle y, z^+ - u \rangle - \|z^+ - z \|^2 \\
&\leq& \|z-u \|^2 + 2 \langle \Delta, z^+ - u \rangle + 2 \langle \Delta, r - (z - y)  \rangle \\
&&- 2 \langle y, z^+ - u \rangle - \|z^+ - z \|^2 \\
&\leq& \|z-u \|^2 + 2\|\Delta\| \cdot \| z^+ - u\| + 2 \|\Delta\| \cdot \|\text{proj}_{\mathcal{Z}}(z - y) - \text{proj}_{\mathcal{Z}}(z)\|  \\
&&+ 2 \|\Delta\| \cdot \| y  \| - 2 \langle y, z^+ - u \rangle - \|z^+ - z \|^2
\\
&\overset{\eqref{proj}}{\leq}& \|z-u \|^2 + 2\|\Delta\| \cdot \| z^+ - u\| + 4 \|\Delta\| \cdot \| y  \| 
\\&&- 2 \langle y, z^+ - u \rangle - \|z^+ - z \|^2.
 \end{eqnarray*}
\EndProof

\subsubsection{Convex-concave problems}

\begin{theorem}[Theorem \ref{th6}]
Let $\{ z_m^t\}_{t \geq 0}$ denote the iterates of Algorithm~\ref{alg2} for solving problem \eqref{distr}. Let Assumptions \ref{ass:as1g}, \ref{as:as1l}, \ref{ass:as2n}, \ref{as:as3} 
 and \ref{as:as4}  be satisfied. Then if $\gamma \leq \frac{1}{4L}$ and $P = \mathcal{O}\left(\sqrt{\chi} \log \frac{1}{\varepsilon}\right)$, we have the following estimate
\begin{equation*}
    \E[\mathrm{gap}(\bar z^{k}_{avg})] = \mathcal{\tilde O}\left(\frac{L \Omega_z^2 \sqrt{\chi}}{K} + \frac{\sigma \Omega_z}{\sqrt{MT}} \right).
\end{equation*}
\end{theorem}

\textbf{Proof:} The same way as in Theorem \ref{1th1} one can get
\begin{align}
\label{1d}
 \| z^{t+1} - z \|^2  &\leq \| z^t - z \|^2 - \| z^{t+1/2} -  z^t \|^2 - 2 \gamma \langle  \bar g^{t+1/2},  z^{t+1/2} - z \rangle \notag\\
 &\hspace{0.4cm} + \gamma^2 \| \bar g^{t+1/2} -  \bar g^{t}\|^2 +4 \|\Delta^{t+1/2} \|\cdot\|z^{t+1} - z\| + 4\E\left[\|\Delta^{t+1/2}\| \cdot \|\gamma \bar g^{t+1/2} \| \right] 
 \notag\\
 &\hspace{0.4cm}+4\|\Delta^t \|\cdot\|z^{t+1/2} - z^{t+1}\| + 4\|\Delta^t\| \cdot \|\gamma \bar g^{t} \| \notag\\
 &\leq \| z^t - z \|^2 - \| z^{t+1/2} -  z^t \|^2 \notag\\
 &\hspace{0.4cm}- 2 \gamma \langle  \bar g^{t+1/2},  z^{t+1/2} - z \rangle + \gamma^2 \| \bar g^{t+1/2} -  \bar g^{t}\|^2 \notag\\
 &\hspace{0.4cm}+4\varepsilon_0 \|z^{t+1} - z\| + 4\varepsilon_0 \gamma\| \bar g^{t+1/2} \| \notag\\
 &\hspace{0.4cm}+4 \varepsilon_0\|z^{t+1/2} - z^{t+1}\| + 4\varepsilon_0 \gamma\|g^{t} \|.  
 \end{align}
Here we use $\| \Delta^t \|, \| \Delta^{t+1/2} \| \leq \varepsilon_0$ and the triangle inequality. Next we use estimate on $\mathrm{gap}$ \eqref{temp8} and taking full expectation:
\begin{align}
2 \gamma k\cdot \E[\mathrm{gap}(\bar z^{k}_{avg})]
&\leq 2 \gamma \E \left[\max_{z \in \mathcal{Z}} \sum\limits_{t=0}^{k-1}\langle  F(z^{t+1/2}),  z^{t+1/2} - z \rangle\right] \notag\\
 &\leq \Omega_z^2  - \sum\limits_{t=0}^{k-1}\E\left[\| z^{t+1/2} -  z^t \|^2\right] \notag\\
 &\hspace{0.4cm}+ 2 \gamma \E\left[ \max_{z \in \mathcal{Z}} \sum\limits_{t=0}^{k-1} \langle  F(z^{t+1/2}) - \bar g^{t+1/2},  z^{t+1/2} - z \rangle\right] \notag\\
 &\hspace{0.4cm}+ \gamma^2 \sum\limits_{t=0}^{k-1} \E\left[\| \bar g^{t+1/2} -  \bar g^{t}\|^2\right]\notag\\ &\hspace{0.4cm} +4\varepsilon_0 \sum\limits_{t=0}^{k-1} \E\left[ \max_{z \in \mathcal{Z}} \|z^{t+1} - z\|\right] + 4\varepsilon_0  \gamma \sum\limits_{t=0}^{k-1}\E\left[\| \bar g^{t+1/2} \| \right] \notag\\
 &\hspace{0.4cm}+4 \varepsilon_0 \sum\limits_{t=0}^{k-1} \E\left[\|z^{t+1/2} - z^{t+1}\|\right] + 4\varepsilon_0 \gamma \sum\limits_{t=0}^{k-1}\E\left[\|\bar g^{t} \|\right].  
 \label{1d4}
 \end{align}
Let us work with $\E\left[\|\bar g^{t+1/2} - \bar g^{t}\|^2\right]$:
\begin{eqnarray*}
\E\left[\| g^{t+1/2} - g^{t}\|^2\right] &=& \E\Bigg[\| \bar g^{t+1/2} - \frac{1}{M}\sum\limits_{m=1}^M F_m(z^{t+1/2}_m) + \frac{1}{M}\sum\limits_{m=1}^M F_m(z^{t+1/2}_m) \\
&&- F(z^{t+1/2}) + F(z^{t}) - \frac{1}{M}\sum\limits_{m=1}^M F_m(z^{t}_m) + \frac{1}{M}\sum\limits_{m=1}^M F_m(z^{t}_m) - \bar g^{t} \\
&&+ F(z^{t+1/2}) - F(z^{t}) \|^2\Bigg] \\
&\overset{\eqref{eq:squared_sum}}{\leq}& 
5\E\left[\left\| \frac{1}{bM} \sum\limits_{m=1}^M \sum\limits_{i=1}^b (F_m(z^{t+1/2}_m, \xi^{t+1/2,i}_m) - F_m(z^{t+1/2}_m)) \right\|^2\right] \\
&&+ 5\E\left[\left\| \frac{1}{M}\sum\limits_{m=1}^M F_m(z^{t+1/2}_m) - F(z^{t+1/2}) \right\|^2\right] \\
&& +5\E\left[\left\| F(z^{t}) - \frac{1}{M}\sum\limits_{m=1}^M F_m(z^{t}_m) \right\|^2\right] \\
&&+5\E\left[\left\| \frac{1}{bM} \sum\limits_{m=1}^M \sum\limits_{i=1}^b (F_m(z^t_m, \xi^{t,i}_m) -  F_m(z^t_m)) \right\|^2\right] \\
&&+5\E\left[\left\| F(z^{t+1/2}) - F(z^{t}) \right\|^2\right].
\end{eqnarray*}
Using that all $\{\xi^{t,i}_m\}_{i=1, m=1}^{b,M}$ and $\{\xi^{t+1/2,i}_m\}_{i=1, m=1}^{b,M}$ are independent, we get
\begin{eqnarray}
\label{t432dd}
&&\hspace{-2cm}\E\left[\| \bar g^{t+1/2} - \bar g^{t}\|^2\right] 
\nonumber\\
&\overset{\eqref{as1g}}{\leq}& \frac{5}{(bM)^2} \sum\limits_{m=1}^M \sum\limits_{i=1}^b \E\left[\left\|  F_m(z^{t+1/2}_m, \xi^{t+1/2,i}_m) -  F_m(z^{t+1/2}_m)\right\|^2\right] \nonumber\\
&&+ \frac{5}{(bM)^2} \sum\limits_{m=1}^M \sum\limits_{i=1}^b \E\left[\left\|  F_m(z^t_m, \xi^{t,i}_m) -  F_m(z^t_m)\right\|^2\right] \nonumber\\
&&+ 5L^2\E\left[\| z^{t+1/2} - z^{t} \|^2\right] \nonumber\\
&&+ 5\E\left[\left\| \frac{1}{M}\sum\limits_{m=1}^M (F_m(z^{t+1/2}_m) - F_m(z^{t+1/2})) \right\|^2\right] \nonumber\\
&&+ 5\E\left[\left\| \frac{1}{M}\sum\limits_{m=1}^M (F_m(z^{t}_m) - F_m(z^{t})) \right\|^2\right]\nonumber\\
&\overset{\eqref{as1l},\eqref{as3}, \eqref{eq:squared_sum}}{\leq}& 5L^2\E\left[\| z^{t+1/2} - z^{t} \|^2\right] + \frac{10 \sigma^2}{bM} \nonumber\\
&&+ 5L_{\max}^2\E\Bigg[\frac{1}{M}\sum\limits_{m=1}^M\Bigg\|   \text{proj}_{\mathcal{Z}} (\tilde z^{t+1/2} + \delta^{t+1/2}_m) 
\nonumber\\
&&\hspace{4cm}- \frac{1}{M} \sum\limits_{j=1}^M \text{proj}_{\mathcal{Z}} (\tilde z^{t+1/2} + \delta^{t+1/2}_j)\Bigg\|^2\Bigg] 
\nonumber\\
&&+ 5L_{\max}^2\E\left[\frac{1}{M}\sum\limits_{m=1}^M\left\|   \text{proj}_{\mathcal{Z}} (\tilde z^{t} + \delta^{t}_m) - \frac{1}{M} \sum\limits_{j=1}^M \text{proj}_{\mathcal{Z}} (\tilde z^{t} + \delta^{t}_j)\right\|^2\right]
\nonumber \\
&\overset{}{\leq}& 5L^2\E\left[\| z^{t+1/2} - z^{t} \|^2\right] + \frac{10 \sigma^2}{bM} \nonumber\\
&&+ 10L_{\max}^2\E\left[\frac{1}{M}\sum\limits_{m=1}^M\left\|   \text{proj}_{\mathcal{Z}} (\tilde z^{t+1/2} + \delta^{t+1/2}_m) -\text{proj}_{\mathcal{Z}} (\tilde z^{t+1/2})\right\|^2\right] 
\nonumber\\
&&+ 10L_{\max}^2\E\left[\frac{1}{M}\sum\limits_{j=1}^M\left\|   (\text{proj}_{\mathcal{Z}} (\tilde z^{t+1/2} + \delta^{t+1/2}_j) - \text{proj}_{\mathcal{Z}} (\tilde z^{t+1/2}))\right\|^2\right] 
\nonumber\\
&&+ 10L_{\max}^2\E\left[\frac{1}{M}\sum\limits_{m=1}^M\left\|   \text{proj}_{\mathcal{Z}} (\tilde z^{t} + \delta^{t}_m) -\text{proj}_{\mathcal{Z}} (\tilde z^{t})\right\|^2\right] 
\nonumber\\
&&+ 10L_{\max}^2\E\left[\frac{1}{M}\sum\limits_{j=1}^M\left\|   (\text{proj}_{\mathcal{Z}} (\tilde z^{t} + \delta^{t}_j) - \text{proj}_{\mathcal{Z}} (\tilde z^{t}))\right\|^2\right] 
\nonumber\\
&\overset{\eqref{delt}}{\leq}& 5L^2\E\left[\| z^{t+1/2} - z^{t} \|^2\right] + \frac{10 \sigma^2}{bM} + 40 L_{\max}^2 \varepsilon_0^2.
\end{eqnarray}
Next we estimate $\E\left[ \max_{z \in \mathcal{Z}} \sum\limits_{t=0}^{k-1} \langle  F(z^{t+1/2}) - \bar g^{t+1/2},  z^{t+1/2} - z \rangle\right]$. To begin with, we use the same approach as in \eqref{t4441}, \eqref{t561} with sequence $v$: $v^0 = z^{1/2}$, $v^{t+1} = \text{proj}_{\mathcal{Z}}(v^t-\gamma (F(z^{t+1/2}) - \bar g^{t+1/2}))$ and get
\begin{align*}
\max_{z \in \mathcal{Z}} \sum\limits_{t=0}^{k-1} \langle F(z^{t+1/2}) - \bar g^{t+1/2}, z^{t+1/2} - z \rangle \leq& \sum\limits_{t=0}^{k-1} \langle F(z^{t+1/2}) - g^{t+1/2}, z^{t+1/2} - v^t \rangle \nonumber\\ 
&+     \frac{\gamma}{2}\sum\limits_{t=0}^{k-1} \|F(z^{t+1/2}) - \bar g^{t+1/2}\|^2 + \frac{\Omega_z^2}{2\gamma}. 
\end{align*}
To begin with, we use the independence of all $\xi$, as well as the unbiasedness of $ \bar g^{t+1/2}$ with respect to the conditional m.o. by random variables $\{\xi^{t+1/2,i}_m\}_{i=1, m=1}^{b,M}$:
\begin{eqnarray}
&&\hspace{-1cm}\E\Bigg[\max_{z \in \mathcal{Z}} \sum\limits_{t=0}^{k-1}  \langle  F(z^{t+1/2}) - \bar g^{t+1/2},  z^{t+1/2} - z \rangle\Bigg] \nonumber\\
&\leq& \sum\limits_{t=0}^{k-1} \E\left[\E_{\{\xi^{t+1/2,i}_m\}_{i=1, m=1}^{b,M}}\left[\langle  F(z^{t+1/2}) - \bar g^{t+1/2},  z^{t+1/2} - v^t \rangle\right]\right] \nonumber\\
&&+     \frac{\gamma}{2}\sum\limits_{t=0}^{k-1} \E\left[\|F(z^{t+1/2}) - \bar g^{t+1/2}\|^2\right] + \frac{\Omega_z^2}{2\gamma} \nonumber\\
&=& \sum\limits_{t=0}^{k-1} \E\left[\langle \E_{\{\xi^{t+1/2,i}_m\}_{i=1, m=1}^{b,M}}\left[ F(z^{t+1/2}) - \bar g^{t+1/2}\right],  z^{t+1/2} - v^t \rangle\right] \nonumber\\
&&+     \frac{\gamma}{2}\sum\limits_{t=0}^{k-1} \E\left[\|F(z^{t+1/2}) - \bar g^{t+1/2}\|^2\right] + \frac{\Omega_z^2}{2\gamma}
\nonumber\\
&=& \sum\limits_{t=0}^{k-1}\E\left[\langle \frac{1}{M} \sum\limits_{m=1}^M (F_m(z^{t+1/2}) - F_m(z_m^{t+1/2})),  z^{t+1/2} - v^t \rangle\right]  \nonumber\\
&&+     \frac{\gamma}{2}\sum\limits_{t=0}^{k-1} \E\left[\|F(z^{t+1/2}) - \bar g^{t+1/2}\|^2\right] + \frac{\Omega_z^2}{2\gamma}
\nonumber\\
&\leq& \sum\limits_{t=0}^{k-1}\E\left[\left\| \frac{1}{M} \sum\limits_{m=1}^M (F_m(z_m^{t+1/2}) - F_m(z^{t+1/2}))\right\| \cdot \|  z^{t+1/2} - v^t \|\right]
\nonumber\\
&&+     \frac{\gamma}{2}\sum\limits_{t=0}^{k-1} \E\left[\|F(z^{t+1/2}) - \bar g^{t+1/2}\|^2\right] + \frac{\Omega_z^2}{2\gamma}
\nonumber\\
&\overset{\eqref{as1l}}{\leq}& \sum\limits_{t=0}^{k-1}\E\left[\left(\frac{L_{\max}}{M} \sum\limits_{m=1}^M \left\|  z_m^{t+1/2} - z^{t+1/2}\right\|\right) \cdot \|  z^{t+1/2} - v^t \|\right]
\nonumber\\
&&+     \frac{\gamma}{2}\sum\limits_{t=0}^{k-1} \E\left[\|F(z^{t+1/2}) - \bar g^{t+1/2}\|^2\right] + \frac{\Omega_z^2}{2\gamma}
\nonumber\\
&\leq&  \sum\limits_{t=0}^{k-1}\E\Bigg[\Bigg(\frac{L_{\max}}{M} \sum\limits_{m=1}^M \Big\|   \text{proj}_{\mathcal{Z}} (\tilde z^{t+1/2} + \delta^{t+1/2}_m) 
\nonumber\\
&&\hspace{4cm} - \frac{1}{M} \sum\limits_{j=1}^M \text{proj}_{\mathcal{Z}} (\tilde z^{t+1/2} + \delta^{t+1/2}_j)\Bigg\|\Bigg) \cdot \|  z^{t+1/2} - v^t \|\Big] \nonumber\\
&&+     \frac{\gamma}{2}\sum\limits_{t=0}^{k-1} \E\left[\|F(z^{t+1/2}) - \bar g^{t+1/2}\|^2\right] + \frac{\Omega_z^2}{2\gamma}
\nonumber \\
&\leq& \sum\limits_{t=0}^{k-1}\E\left[\left(\frac{L_{\max}}{M} \sum\limits_{m=1}^M \left\|   \text{proj}_{\mathcal{Z}} (\tilde z^{t+1/2} + \delta^{t+1/2}_m) - \text{proj}_{\mathcal{Z}} (\tilde z^{t+1/2})\right\|\right) \cdot \|  z^{t+1/2} - v^t \|\right]
\nonumber\\
\label{t505d}
&&+\E\left[\left(\frac{L_{\max}}{M} \sum\limits_{j=1}^M \left\| (\text{proj}_{\mathcal{Z}} (\tilde z^{t+1/2} + \delta^{t+1/2}_j) - \text{proj}_{\mathcal{Z}} (\tilde z^{t+1/2} ))\right\|\right) \cdot \|  z^{t+1/2} - v^t \|\right]
\nonumber\\
&&+     \frac{\gamma}{2}\sum\limits_{t=0}^{k-1} \E\left[\|F(z^{t+1/2}) - \bar g^{t+1/2}\|^2\right] + \frac{\Omega_z^2}{2\gamma}
\nonumber\\
&\overset{\eqref{delt}}{\leq}& 2 L_{\max} \varepsilon_0 \sum\limits_{t=0}^{k-1}\E\left[\|  z^{t+1/2} - v^t \|\right] +     \frac{\gamma}{2}\sum\limits_{t=0}^{k-1} \E\left[\|F(z^{t+1/2}) - \bar g^{t+1/2}\|^2\right] + \frac{\Omega_z^2}{2\gamma}\nonumber\\
&\leq& 2 L_{\max} \varepsilon_0 k \Omega_z +     \frac{\gamma}{2}\sum\limits_{t=0}^{k-1} \E\left[\|F(z^{t+1/2}) - \bar g^{t+1/2}\|^2\right] + \frac{\Omega_z^2}{2\gamma}
.
\end{eqnarray}
Next we combine \eqref{1d4}, \eqref{t432dd} and \eqref{t505d}
\begin{align*}
2\gamma k\E[\mathrm{gap}(\bar z^{k}_{avg})] \leq& 2\Omega_z^2 +(5L^2 \gamma^2 -1)\sum\limits_{t=0}^{k-1} \E\left[\| z^{t+1/2} -  z^t \|^2\right] \notag\\
 &+ 4 \gamma L_{\max} \varepsilon_0 k \Omega_z +     \gamma^2\sum\limits_{t=0}^{k-1} \E\left[\|F(z^{t+1/2}) - \bar g^{t+1/2}\|^2\right] + \gamma^2 \frac{10 k \sigma^2}{bM} \notag\\
 & + 40 \gamma^2 k L_{\max}^2 \varepsilon_0^2 +4\varepsilon_0 \sum\limits_{t=0}^{k-1} \E\left[ \max_{z \in \mathcal{Z}} \|z^{t+1} - z\|\right] \notag\\
 & + 4\varepsilon_0  \gamma \sum\limits_{t=0}^{k-1}\E\left[\| \bar g^{t+1/2} \| \right] +4 \varepsilon_0 \sum\limits_{t=0}^{k-1} \E\left[\|z^{t+1/2} - z^{t+1}\|\right] \notag\\
 & + 4\varepsilon_0 \gamma \sum\limits_{t=0}^{k-1}\E\left[\| \bar g^{t} \|\right].
 \end{align*}
Then we use $\gamma \leq \frac{1}{4L}$ and Assumption \ref{as:as4}:
\begin{align}
\label{tempd1}
2\gamma k\E[\mathrm{gap}(\bar z^{k}_{avg})] 
 &\leq 2\Omega_z^2 + \gamma^2\sum\limits_{t=0}^{k-1} \E\left[\|F(z^{t+1/2}) - \bar g^{t+1/2}\|^2\right] \notag\\
 &\hspace{0.4cm}+ \gamma^2 \frac{10 k \sigma^2}{bM} + 40 \gamma^2 k L_{\max}^2 \varepsilon_0^2 +8 (1+ \gamma L_{\max})\varepsilon_0 k\Omega_z \notag\\
 &\hspace{0.4cm} + 4\varepsilon_0  \gamma \sum\limits_{t=0}^{k-1}\E\left[\|\bar g^{t+1/2} \| \right] + 4\varepsilon_0 \gamma \sum\limits_{t=0}^{k-1}\E\left[\|\bar g^{t} \|\right].
 \end{align}
It remains to estimate $\E\left[\|\bar g^{t+1/2} \| + \|\bar g^{t} \|\right]$:
 \begin{align*}
\E\left[\|\bar g^{t}\|\right] =& \E\Bigg[\| F(z^*) - F(z^*) + F(z^{t})  - F(z^{t}) + \frac{1}{M}\sum\limits_{m=1}^M F_m(z^{t}_m) - \frac{1}{M}\sum\limits_{m=1}^M F_m(z^{t}_m) + \bar g^{t} \|\Bigg] \\
\leq&  \| F(z^*) \| + \E\left[\left\| F(z^{t}) - F(z^{*}) \right\|\right] +
\E\left[\left\| \frac{1}{M}\sum\limits_{m=1}^M F_m(z^{t}_m) - F(z^{t}) \right\|\right] \\
&+\E\left[\left\| \frac{1}{bM} \sum\limits_{m=1}^M \sum\limits_{i=1}^b (F_m(z^t_m, \xi^{t,i}_m) -  F_m(z^t_m)) \right\|\right].
\end{align*}
From \eqref{t432dd} we have that $\E\left[\left\| \frac{1}{bM} \sum\limits_{m=1}^M \sum\limits_{i=1}^b (F_m(z^t_m, \xi^{t,i}_m) -  F_m(z^t_m)) \right\|^2\right] \leq \frac{\sigma^2}{bM}$ and from \eqref{t505d} we have $\E\left[\left\| \frac{1}{M}\sum\limits_{m=1}^M F_m(z^{t}_m) - F(z^{t}) \right\|\right] \leq 2 L_{\max} \varepsilon_0$, then
 \begin{align*}
\E\left[\|g^{t}\|\right] \leq&  \| F(z^*) \| + \E\left[\left\| F(z^{t}) - F(z^{*}) \right\|\right] +
2 L_{\max} \varepsilon_0 + \frac{\sigma}{\sqrt{bM}} \\
\leq& Q + L \Omega_z +
2 L_{\max} \varepsilon_0 + \frac{\sigma}{\sqrt{bM}},
  \end{align*}
where $Q^2 = \frac{1}{M} \sum\limits_{m=1}^M \| F_m(z^*) \|^2$. 
Hence, we can rewrite \eqref{tempd1} as follows:
\begin{align*}
\E[\mathrm{gap}(\bar z^{k}_{avg})] 
 \leq& \frac{\Omega_z^2}{2\gamma k} + \frac{\gamma}{2k}\sum\limits_{t=0}^{k-1} \E\left[\|F(z^{t+1/2}) - \bar g^{t+1/2}\|^2\right] + \frac{5 \sigma^2  \gamma}{bM}\notag\\
 & + 20 \gamma L_{\max}^2 \varepsilon_0^2 +4 \left(\frac{1}{\gamma}+ L_{\max}\right)\varepsilon_0 \Omega_z \\
 &+ 4\varepsilon_0 \left( Q + L \Omega_z +
2 L_{\max} \varepsilon_0 + \frac{\sigma}{\sqrt{bM}}\right).
 \end{align*}
The same way as \eqref{t432dd}, one can estimate $E\left[\|F(z^{t+1/2}) - \bar g^{t+1/2}\|^2\right]$:
\begin{align*}
\E[\mathrm{gap}(\bar z^{k}_{avg})] 
 \leq& \frac{\Omega_z^2}{2\gamma k} + \frac{6 \sigma^2  \gamma}{bM}\notag\\
 & + 24 \gamma L_{\max}^2 \varepsilon_0^2 +4 \left(\frac{1}{\gamma}+ L_{\max}\right)\varepsilon_0 \Omega_z \\
 &+ 4\varepsilon_0 \left( Q + L \Omega_z +
2 L_{\max} \varepsilon_0 + \frac{\sigma}{\sqrt{bM}}\right).
 \end{align*}
Let $\gamma = \min\left\{\frac{1}{4L}; \Omega_z\sqrt{\frac{bM}{12k\sigma^2}}\right\}$ and $\varepsilon_0 = \mathcal{O}\left(\frac{\varepsilon}{\Omega_z L_{\max} + Q}\right)$, where $\varepsilon = \max\left(\frac{L \Omega_z^2}{k}; \frac{\sigma \Omega_z}{\sqrt{bMk}}\right)$. Then for the output of Algorithm \ref{alg3} it holds
\begin{equation*}
    \E[\mathrm{gap}(\bar z^{k}_{avg})] = \mathcal{O}\left(\frac{L \Omega_z^2}{k} + \frac{\sigma \Omega_z}{\sqrt{bMk}} \right).
\end{equation*}
Substituting the batch size $b$ and the number of iterations $k$ from the description of Algorithm \ref{alg2} and Corollary  \ref{cor45}:
\begin{equation*}
    \E[\mathrm{gap}(\bar z^{k}_{avg})] = \mathcal{\tilde O}\left(\frac{L \Omega_z^2 \sqrt{\chi}}{K} + \frac{\sigma \Omega_z}{\sqrt{MT}} \right).
\end{equation*}
\EndProof

\subsubsection{Strongly convex-strongly concave problems}

\begin{theorem}[Theorem \ref{th6}] 
Let $\{ z_m^t\}_{t \geq 0}$ denote the iterates of Algorithm~\ref{alg2} for solving the problem \eqref{distr}. Let Assumptions \ref{ass:as1g}, \ref{as:as1l}, \ref{ass:as2g} and \ref{as:as3} be satisfied. Then if $\gamma \leq \frac{1}{4L}$ and $P = \mathcal{O}\left(\sqrt{\chi} \log \frac{1}{\varepsilon}\right)$, we have the following estimate:
\begin{equation*}
    \E\left[\| \bar z^{k} - z^* \|^2\right]  = 
 \mathcal{\tilde O}\left(\| z^{0} - z^* \|^2 \exp\left( - \frac{\mu }{8L} \cdot\frac{K}{\sqrt{\chi}} \right) + \frac{\sigma^2}{\mu^2 M T}\right) .
\end{equation*}
\end{theorem}
\textbf{Proof:} We start with substituting $z = z^*$ in \eqref{1d} and taking full expectation. Then we use \eqref{t432dd} and get
\begin{align*}
2 \gamma \E\bigg[\langle F(z^{t+1/2}),&  z^{t+1/2} - z^* \rangle\bigg]  \leq \E\left[\| z^t - z^* \|^2\right] - \E\left[\| z^{t+1} - z^* \|^2\right] - \E\left[\| z^{t+1/2} -  z^t \|^2\right] \notag\\
 & + 5L^2\gamma^2 \E\left[\| z^{t+1/2} - z^{t} \|^2\right] + \frac{10 \sigma^2 \gamma^2 }{bM}\notag\\
 &+4\varepsilon_0 \E\left[\|z^{t+1} - z^*\|\right] + 4\varepsilon_0 \gamma\E\left[\|\bar g^{t+1/2} \| \right]
 \notag\\
 &+4 \varepsilon_0\E\left[\|z^{t+1/2} - z^{t+1}\|\right] + 4\varepsilon_0 \gamma\E\left[\|\bar g^{t} \|\right] \notag\\
 &+ 2 \gamma \E\left[\langle  F(z^{t+1/2}) - \bar g^{t+1/2},  z^{t+1/2} - z^* \rangle\right] + 40 \gamma^2 L_{\max}^2 \varepsilon_0^2 .
 \end{align*}
 The same way as \eqref{t505d}, one can get
 \begin{eqnarray*}
&&\hspace{-1cm}\E\bigg[ \langle F(z^{t+1/2})  - \bar g^{t+1/2},  z^{t+1/2} - z^* \rangle\bigg]\nonumber\\
&\leq&  \E\left[\E_{\{\xi^{t+1/2,i}_m\}_{i=1, m=1}^{b,M}}\left[\langle  F(z^{t+1/2}) - \bar g^{t+1/2},  z^{t+1/2} - z^* \rangle\right]\right] \nonumber\\
&=& \E\left[\langle \E_{\{\xi^{t+1/2,i}_m\}_{i=1, m=1}^{b,M}}\left[ F(z^{t+1/2}) - \bar g^{t+1/2}\right],  z^{t+1/2} - z^* \rangle\right] 
\nonumber\\
&=& \E\left[\langle \frac{1}{M} \sum\limits_{m=1}^M (F_m(z^{t+1/2}) - F_m(z_m^{t+1/2})),  z^{t+1/2} - z^* \rangle\right]  
\nonumber\\
&\leq& \E\left[\left\| \frac{1}{M} \sum\limits_{m=1}^M (F_m(z_m^{t+1/2}) - F_m(z^{t+1/2}))\right\| \cdot \|  z^{t+1/2} - z^* \|\right]
\nonumber\\
&\overset{\eqref{as1l}}{\leq}& \E\left[\left(\frac{L_{\max}}{M} \sum\limits_{m=1}^M \left\|  z_m^{t+1/2} - z^{t+1/2}\right\|\right) \cdot \|  z^{t+1/2} - z^* \|\right]
\nonumber\\
&\leq&  \E\Bigg[\Bigg(\frac{L_{\max}}{M} \sum\limits_{m=1}^M \bigg\|   \text{proj}_{\mathcal{Z}} (\tilde z^{t+1/2} + \delta^{t+1/2}_m) 
\nonumber\\
&&\hspace{4cm}- \frac{1}{M} \sum\limits_{j=1}^M \text{proj}_{\mathcal{Z}} (\tilde z^{t+1/2} + \delta^{t+1/2}_j)\bigg\|\Bigg) \cdot \|  z^{t+1/2} - z^*\|\Bigg] \nonumber\\
&\leq& \E\left[\left(\frac{L_{\max}}{M} \sum\limits_{m=1}^M \left\|   \text{proj}_{\mathcal{Z}} (\tilde z^{t+1/2} + \delta^{t+1/2}_m) - \text{proj}_{\mathcal{Z}} (\tilde z^{t+1/2})\right\|\right) \cdot \|  z^{t+1/2} - z^* \|\right]
\nonumber\\
&&+\E\left[\left(\frac{L_{\max}}{M} \sum\limits_{j=1}^M \left\| (\text{proj}_{\mathcal{Z}} (\tilde z^{t+1/2} + \delta^{t+1/2}_j) - \text{proj}_{\mathcal{Z}} (\tilde z^{t+1/2} ))\right\|\right) \cdot \|  z^{t+1/2} - z^* \|\right]
\nonumber\\
&\overset{\eqref{delt}}{\leq}& 2 L_{\max} \varepsilon_0 \E\left[\|  z^{t+1/2} - z^* \|\right] .
\end{eqnarray*}
and then 
\begin{align*}
2 \gamma \E\left[\langle F(z^{t+1/2}),  z^{t+1/2} - z^* \rangle\right] 
\leq& \E\left[\| z^t - z^* \|^2\right] - \E\left[\| z^{t+1} - z^* \|^2\right] - \E\left[\| z^{t+1/2} -  z^t \|^2\right] \notag\\
 &+ 5L^2\gamma^2 \E\left[\| z^{t+1/2} - z^{t} \|^2\right] + \frac{10 \sigma^2 \gamma^2 }{bM}\notag\\
 &+4\varepsilon_0 \E\left[\|z^{t+1} - z^*\|\right] + 4\varepsilon_0 \gamma\E\left[\|\bar g^{t+1/2} \| \right]\\
 &+4 \varepsilon_0\E\left[\|z^{t+1/2} - z^{t+1}\|\right] + 4\varepsilon_0 \gamma\E\left[\|\bar g^{t} \|\right] \notag\\
 &+ 4 \gamma L_{\max} \varepsilon_0 \E\left[\|  z^{t+1/2} - z^* \|\right] + 40 \gamma^2 L_{\max}^2 \varepsilon_0^2 .
 \end{align*}
 Next, we work with
 \begin{align*}
z^{t+1} &= \frac{1}{M} \sum\limits_{m=1}^M z^{t+1}_m = \frac{1}{M} \sum\limits_{m=1}^M \text{proj}_{\mathcal{Z}} (\tilde z^{t+1/2} + \delta^{t+1}_m) \nonumber\\
&= \text{proj}_{\mathcal{Z}} (\tilde z^{t+1}) + \frac{1}{M} \sum\limits_{m=1}^M \text{proj}_{\mathcal{Z}} (\tilde z^{t+1} + \delta^{t+1}_m) - \text{proj}_{\mathcal{Z}} (\tilde z^{t+1})
\nonumber\\
&= \text{proj}_{\mathcal{Z}} \left(\frac{1}{M}\sum\limits_{m=1}^M z^t_m - \gamma g^{t+1/2}_m\right) + \Delta^{t+1/2}
= \text{proj}_{\mathcal{Z}} \left(z^t - \gamma \bar g^{t+1/2}\right) + \Delta^{t+1/2},
\end{align*}
and \ans{using the triangle inequality of the forms $\|z^{t+1} - z^*\| \leq \|z^{t+1} - z^t\| + \|z^{t} - z^*\|$, $\|z^{t+1/2} - z^{t+1}\| \leq \|z^{t+1/2} - z^{t}\| + \|z^{t+1/2} - z^{t}\|$ and $\|  z^{t+1/2} - z^* \| \leq \|  z^{t} - z^* \| + \|  z^{t+1/2} - z^t \|$}, we get
\begin{eqnarray}
\label{8t8}
&&\hspace{-1cm}2 \gamma \E\left[\langle F(z^{t+1/2}),  z^{t+1/2} - z^* \rangle\right]  \nonumber\\
&\leq& \E\left[\| z^t - z^* \|^2\right] - \E\left[\| z^{t+1} - z^* \|^2\right] - \E\left[\| z^{t+1/2} -  z^t \|^2\right] \notag\\
 && + 5L^2\gamma^2 \E\left[\| z^{t+1/2} - z^{t} \|^2\right] + \frac{10 \sigma^2 \gamma^2 }{bM}\notag\\
 &&+8\varepsilon_0 \E\left[\|z^{t+1} - z^t\|\right] + 4\varepsilon_0 \gamma\E\left[\|\bar g^{t+1/2} \| \right] 
 + 4\varepsilon_0 \gamma\E\left[\|\bar g^{t} \|\right] \notag\\
 &&+ 4 \varepsilon_0 ( 1 + \gamma L_{\max}) \E\left[\|  z^{t+1/2} - z^t \|\right] \notag\\
 &&+ 4\varepsilon_0 (1 + \gamma L_{\max}) \E\left[\|  z^{t} - z^* \|\right] + 40 \gamma^2 L_{\max}^2 \varepsilon_0^2 \notag\\
 &\leq& 
\E\left[\| z^t - z^* \|^2\right] - \E\left[\| z^{t+1} - z^* \|^2\right] - \E\left[\| z^{t+1/2} -  z^t \|^2\right] \notag\\
 && + 5L^2\gamma^2 \E\left[\| z^{t+1/2} - z^{t} \|^2\right] + \frac{10 \sigma^2 \gamma^2 }{bM}\notag\\
 && +8\varepsilon_0 \E\left[\|\text{proj}_{\mathcal{Z}} \left(z^t - \gamma g^{t+1/2}\right) + \Delta^{t+1/2} - z^t\|\right] \notag\\
 && + 4\varepsilon_0 \gamma\E\left[\|\bar g^{t+1/2} \| \right] 
 + 4\varepsilon_0 \gamma\E\left[\|\bar g^{t} \|\right] \notag\\
 && + 4 \varepsilon_0 ( 1 + \gamma L_{\max}) \E\left[\|  z^{t+1/2} - z^t \|\right] \notag\\
 && + 4\varepsilon_0 (1 + \gamma L_{\max}) \E\left[\|  z^{t} - z^* \|\right] + 40 \gamma^2 L_{\max}^2 \varepsilon_0^2 \notag\\
 &\leq& \E\left[\| z^t - z^* \|^2\right] - \E\left[\| z^{t+1} - z^* \|^2\right] - \E\left[\| z^{t+1/2} -  z^t \|^2\right] \notag\\
 && + 5L^2\gamma^2 \E\left[\| z^{t+1/2} - z^{t} \|^2\right] + \frac{10 \sigma^2 \gamma^2 }{bM}\notag\\
 && +8\varepsilon_0 \E\left[\|\text{proj}_{\mathcal{Z}} \left(z^t - \gamma g^{t+1/2}\right) - \text{proj}_{\mathcal{Z}}(z^t)\|\right] +8\varepsilon_0^2 \notag\\
 && +4\varepsilon_0 (1 + \gamma L_{\max})\E\left[\|z^{t} - \add{z^*}\|\right] + 4\varepsilon_0 \gamma\E\left[\|\bar g^{t+1/2} \| \right]\notag\\
 && + 4\varepsilon_0 \gamma\E\left[\|\bar g^{t} \|\right] +  4\varepsilon_0(1 + \gamma L_{\max}) \E\left[\|  z^{t+1/2} - z^t \|\right] + 40 \gamma^2 L_{\max}^2 \varepsilon_0^2 \notag\\
 &\overset{\eqref{proj}}{\leq}& \E\left[\| z^t - z^* \|^2\right] - \E\left[\| z^{t+1} - z^* \|^2\right] - \E\left[\| z^{t+1/2} -  z^t \|^2\right] \notag\\
 && + 5L^2\gamma^2 \E\left[\| z^{t+1/2} - z^{t} \|^2\right] + \frac{10 \sigma^2 \gamma^2 }{bM}\notag\\
 && +8\varepsilon_0^2 +4\varepsilon_0 (1 + \gamma L_{\max})\E\left[\|z^{t} - z^*\|\right] + 12\varepsilon_0 \gamma\E\left[\|\bar g^{t+1/2} \| \right]\notag\\
 && + 4\varepsilon_0 \gamma\E\left[\|\bar g^{t} \|\right] +  4\varepsilon_0(1 + \gamma L_{\max}) \E\left[\|  z^{t+1/2} - z^t \|\right] + 40 \gamma^2 L_{\max}^2 \varepsilon_0^2.
 \end{eqnarray}
It remains to estimate $\E\left[\|\bar g^{t+1/2} \| + \|\bar g^{t} \|\right]$:
 \begin{align*}
\E\left[\|\bar g^{t}\|\right] =& \E\Bigg[\| F(z^*) - F(z^*) + F(z^{t}) - F(z^{t}) + \frac{1}{M}\sum\limits_{m=1}^M F_m(z^{t}_m) - \frac{1}{M}\sum\limits_{m=1}^M F_m(z^{t}_m) + \bar g^{t} \|\Bigg] \\
\leq&  \| F(z^*) \| + \E\left[\left\| F(z^{t}) - F(z^{*}) \right\|\right] +
\E\left[\left\| \frac{1}{M}\sum\limits_{m=1}^M F_m(z^{t}_m) - F(z^{t}) \right\|\right] \\
&+\E\left[\left\| \frac{1}{bM} \sum\limits_{m=1}^M \sum\limits_{i=1}^b (F_m(z^t_m, \xi^{t,i}_m) -  F_m(z^t_m)) \right\|\right].
  \end{align*}
 From \eqref{t432dd} we have that $\E\left[\left\| \frac{1}{bM} \sum\limits_{m=1}^M \sum\limits_{i=1}^b (F_m(z^t_m, \xi^{t,i}_m) -  F_m(z^t_m)) \right\|^2\right] \leq \frac{\sigma^2}{bM}$ and from \eqref{t505d} $\E\left[\left\| \frac{1}{M}\sum\limits_{m=1}^M F_m(z^{t}_m) - F(z^{t}) \right\|\right] \leq 2 L_{\max} \varepsilon_0$, then
 \begin{align*}
\E\left[\|\bar g^{t}\|\right] &\leq\| F(z^*) \| + \E\left[\left\| F(z^{t}) - F(z^{*}) \right\|\right] +
2 L_{\max} \varepsilon_0 + \frac{\sigma}{\sqrt{bM}} \\
&\leq Q + L \E\left[\|z^t - z^*\|\right] +
2 L_{\max} \varepsilon_0 + \frac{\sigma}{\sqrt{bM}}.
  \end{align*}
Substituting in \eqref{8t8}:
\begin{align}
\label{temp20808}
2 \gamma \E&\left[\langle F(z^{t+1/2}),  z^{t+1/2} - z^* \rangle\right]  \notag\\
\leq& \E\left[\| z^t - z^* \|^2\right] - \E\left[\| z^{t+1} - z^* \|^2\right] - \E\left[\| z^{t+1/2} -  z^t \|^2\right] \notag\\
 & + 5L^2\gamma^2 \E\left[\| z^{t+1/2} - z^{t} \|^2\right] + \frac{10 \sigma^2 \gamma^2 }{bM} +8\varepsilon_0^2 +4\varepsilon_0 (1 + \gamma L_{\max})\E\left[\|z^{t} - z^*\|\right] \nonumber\\
 &+ 12\varepsilon_0 \gamma\left( Q + L \E\left[\|z^{t+1/2} - z^*\|\right] +
2 L_{\max} \varepsilon_0 + \frac{\sigma}{\sqrt{bM}}\right)\notag\\
 &+ 4\varepsilon_0 \gamma\left( Q + L \E\left[\|z^t - z^*\|\right] +
2 L_{\max} \varepsilon_0 + \frac{\sigma}{\sqrt{bM}}\right) \notag\\
 &+  4\varepsilon_0(1 + \gamma L_{\max}) \E\left[\|  z^{t+1/2} - z^t \|\right] + 40 \gamma^2 L_{\max}^2 \varepsilon_0^2.
 \end{align}
 By simple fact $2ab \leq a^2 + b^2$, we get
 \begin{align}
 \label{a5a}
2 \gamma \E&\left[\langle F(z^{t+1/2}),  z^{t+1/2} - z^* \rangle\right]  \nonumber\\
\leq& \E\left[\| z^t - z^* \|^2\right] - \E\left[\| z^{t+1} - z^* \|^2\right] - \E\left[\| z^{t+1/2} -  z^t \|^2\right] \notag\\
 &+ 5L^2\gamma^2 \E\left[\| z^{t+1/2} - z^{t} \|^2\right] + \frac{10 \sigma^2 \gamma^2 }{bM}\notag\\
 &+20\varepsilon_0 (1 + \gamma L_{\max})\E\left[\|z^{t} - z^*\|\right] +  16\varepsilon_0(1 + \gamma L_{\max}) \E\left[\|  z^{t+1/2} - z^t \|\right]  \nonumber\\
 &+ 16\varepsilon_0 \gamma\left( Q + 2 L_{\max} \varepsilon_0 + \frac{\sigma}{\sqrt{bM}}\right) + 40 \gamma^2 L_{\max}^2 \varepsilon_0^2 +8\varepsilon_0^2 \nonumber\\
 \leq& (1 + 10\varepsilon_0)\E\left[\| z^t - z^* \|^2\right] - \E\left[\| z^{t+1} - z^* \|^2\right] \notag\\
 & + (5L^2\gamma^2 + 8 \varepsilon_0 - 1) \E\left[\| z^{t+1/2} - z^{t} \|^2\right] + \frac{10 \sigma^2 \gamma^2 }{bM}\notag\\
 &+20\varepsilon_0 (1 + \gamma L_{\max})^2 + 16\varepsilon_0 \gamma\left(Q + 2 L_{\max} \varepsilon_0 + \frac{\sigma}{\sqrt{bM}}\right) \notag\\
 &+ 40 \gamma^2 L_{\max}^2 \varepsilon_0^2 +8\varepsilon_0^2.
 \end{align}
By property of $z^*$, we get
\begin{align*}
\E\left[\langle  F(z^{t+1/2}),  z^{t+1/2} - z^* \rangle\right] &\geq \E\left[\langle F(z^{t+1/2}) -F(z^*),  z^{t+1/2} - z^* \rangle\right] \\
&\geq  \mu\E\left[\|z^{t+1/2} - z^*\|^2\right].
\end{align*}
Let use a simple fact $\|z^{t+1/2} - z^*\|^2   \geq \frac{1}{2}\|z^t - z^*\|^2 - \|z^{t+1/2} - z^t\|^2$, then
\begin{align*}
\E\left[\langle  F(z^{t+1/2}),  z^{t+1/2} - z^* \rangle\right] \geq  \frac{\mu}{2}\E\left[\|z^t - z^*\|^2 \right] - \mu \E\left[\|z^{t+1/2} - z^t\|^2 \right].
\end{align*}
Then \eqref{a5a} gives
\begin{align*}
 \E\left[\| z^{t+1} - z^* \|^2\right] &\leq (1 + 10\varepsilon_0 - \mu \gamma )\E\left[\| z^t - z^* \|^2\right] + \frac{10 \sigma^2 \gamma^2 }{bM} \notag\\
 &\hspace{0.4cm}+ (5L^2\gamma^2 + 2\gamma \mu + 8 \varepsilon_0 - 1) \E\left[\| z^{t+1/2} - z^{t} \|^2\right] \notag\\
 &\hspace{0.4cm}+20\varepsilon_0 (1 + \gamma L_{\max})^2 + 16\varepsilon_0 \gamma\left(Q + 2 L_{\max} \varepsilon_0 + \frac{\sigma}{\sqrt{bM}}\right) \\
 &\hspace{0.4cm}+ 40 \gamma^2 L_{\max}^2 \varepsilon_0^2 +8\varepsilon_0^2.
 \end{align*}
With $\varepsilon_0 \leq \min\left(\frac{1}{50}, \frac{\mu \gamma}{20}\right)$ and $\gamma \leq \frac{1}{4L}$, we have
\begin{align*}
 \E\left[\| z^{t+1} - z^* \|^2\right] &\leq \left(1 - \frac{\mu \gamma}{2} \right)\E\left[\| z^t - z^* \|^2\right] + \frac{10 \sigma^2 \gamma^2 }{bM} \notag\\
 &\hspace{0.4cm}+20\varepsilon_0 (1 + \gamma L_{\max})^2 + 16\varepsilon_0 \gamma\left(Q + 2 L_{\max} \varepsilon_0 + \frac{\sigma}{\sqrt{bM}}\right) \\
 &\hspace{0.4cm}+ 40 \gamma^2 L_{\max}^2 \varepsilon_0^2 +8\varepsilon_0^2.
 \end{align*}
Let us run the recursion from $0$ to $k-1$:
\begin{align*}
 \E\left[\| z^{k} - z^* \|^2\right] &\leq \left(1 - \frac{\mu \gamma}{2} \right)^k\E\left[\| z^0 - z^* \|^2\right] + \frac{20 \sigma^2 \gamma }{\mu bM} \notag\\
 &\hspace{0.4cm}+\frac{2\varepsilon_0}{\mu \gamma}\Bigg(20 (1 + \gamma L_{\max})^2 + 16 \gamma\left(Q + 2 L_{\max} \varepsilon_0 + \frac{\sigma}{\sqrt{bM}}\right) \\
 &\hspace{1.9cm}+ 40 \gamma^2 L_{\max}^2 \varepsilon_0 +8\varepsilon_0\Bigg).
 \end{align*}
Then we 
carefully choose $\gamma = \min\left\{\frac{1}{4L}; \frac{2\ln(\max\{2; b M \mu^2\|z^0 - z^*\|^2 k / 20\sigma^2\})}{\mu k}\right\}$ and $\varepsilon_0 = \mathcal{O}\left( \varepsilon \mu \gamma(1 + Q +\gamma L_{\max})^2 \right)$, where $\varepsilon = \max\left(\| z^{0} - z^* \|^2 \exp\left( - \frac{\mu k}{8L} \right); \frac{\sigma^2}{\mu^2 bM k}\right)$. Then the output of Algorithm \ref{alg3} it holds
 \begin{align*}
 \E\left[\| \bar z^{k} - z^* \|^2\right]  = 
 \mathcal{\tilde O}\left(\| z^{0} - z^* \|^2 \exp\left( - \frac{\mu k}{8L} \right) + \frac{\sigma^2}{\mu^2 bM k}\right) .
 \end{align*}
Substituting the batch size $b$ and the number of iterations $k$ from the description of Algorithm \ref{alg1}:
\begin{align*}
 \E\left[\| \bar z^{k} - z^* \|^2\right]  = 
 \mathcal{\tilde O}\left(\| z^{0} - z^* \|^2 \exp\left( - \frac{\mu }{8L} \cdot\frac{K}{P} \right) + \frac{\sigma^2}{\mu^2 M T}\right) .
 \end{align*}
Corollary  \ref{cor45} ends the proof.
\EndProof

\subsection{Non-convex-non-concave problems}

\begin{theorem}[Theorem \ref{th6}]
Let $\{ z_m^t\}_{t \geq 0}$ denote the iterates of Algorithm~\ref{alg2} for solving problem \eqref{distr}. Let Assumptions \ref{ass:as1g}, \ref{as:as1l}, \ref{ass:as2n}, \ref{as:as3} and \ref{as:as4} be satisfied. Then if $\gamma \leq \frac{1}{4L}$ and $P = \mathcal{O}\left(\sqrt{\chi} \log \frac{1}{\varepsilon}\right)$, we have the following estimate:
\begin{equation*}
    \E\left[ \frac{1}{k}\sum\limits_{t=0}^{k-1}\|F(z^t)\|^2\right] = \mathcal{\tilde O}\left(\frac{L^2 \Omega_z^2 \sqrt{\chi}}{K} + \frac{\sigma^2 K}{MT\sqrt{\chi}} \right).
\end{equation*}
\end{theorem}

\textbf{Proof:} We start from \eqref{temp20808} with using diameter $\Omega_z$:
\begin{align*}
2 \gamma \E&\left[\langle F(z^{t+1/2}),  z^{t+1/2} - z^* \rangle\right]  \notag\\
&\leq \E\left[\| z^t - z^* \|^2\right] - \E\left[\| z^{t+1} - z^* \|^2\right] - (1 - 5L^2\gamma^2)\E\left[\| z^{t+1/2} -  z^t \|^2\right] + \frac{10 \sigma^2 \gamma^2 }{bM}\notag\\
 &\hspace{0.4cm}+ 16\varepsilon_0 \gamma\left( L \Omega_z +
2 L_{\max} \varepsilon_0 + \frac{\sigma}{\sqrt{bM}}\right)+  8\varepsilon_0(1 + \gamma L_{\max}) \Omega_z + 8 (1 + 5\gamma^2 L_{\max}^2) \varepsilon_0^2.
 \end{align*}
With the Minty assumption it transforms to
\begin{align*}
0&\leq \E\left[\| z^t - z^* \|^2\right] - \E\left[\| z^{t+1} - z^* \|^2\right] - (1 - 5L^2\gamma^2)\E\left[\| z^{t+1/2} -  z^t \|^2\right] + \frac{10 \sigma^2 \gamma^2 }{bM}\notag\\
 &\hspace{0.4cm}+ 16\varepsilon_0 \gamma\left( Q + L \Omega_z +
2 L_{\max} \varepsilon_0 + \frac{\sigma}{\sqrt{bM}}\right)+  8\varepsilon_0(1 + \gamma L_{\max}) \Omega_z + 8 (1 + 5\gamma^2 L_{\max}^2) \varepsilon_0^2 \\
&= \E\left[\| z^t - z^* \|^2\right] - \E\left[\| z^{t+1} - z^* \|^2\right] - \gamma^2(1 - 5L^2\gamma^2)\E\left[\| g^t\|^2\right] + \frac{10 \sigma^2 \gamma^2 }{bM}\notag\\
 &\hspace{0.4cm}+ 16\varepsilon_0 \gamma\left( Q + L \Omega_z +
2 L_{\max} \varepsilon_0 + \frac{\sigma}{\sqrt{bM}}\right)+  8\varepsilon_0(1 + \gamma L_{\max}) \Omega_z + 8 (1 + 5\gamma^2 L_{\max}^2) \varepsilon_0^2.
 \end{align*}
After the choice of $\gamma \leq \frac{1}{4L}$ we get
\begin{align*}
0&\leq \E\left[\| z^t - z^* \|^2\right] - \E\left[\| z^{t+1} - z^* \|^2\right] - \frac{\gamma^2}{2}\E\left[\| g^t\|^2\right] + \frac{10 \sigma^2 \gamma^2 }{bM}\notag\\
 &\hspace{0.4cm}+ 16\varepsilon_0 \gamma\left( Q + L \Omega_z +
2 L_{\max} \varepsilon_0 + \frac{\sigma}{\sqrt{bM}}\right)+  8\varepsilon_0(1 + \gamma L_{\max}) \Omega_z + 8 (1 + 5\gamma^2 L_{\max}^2) \varepsilon_0^2.
 \end{align*}
 The fact: $-\| g^t\|^2 \leq -\frac{1}{2} \| F(z^t)\|^2 + \| g^t - F(z^t)\|^2 $, gives
 \begin{align*}
0\leq& \E\left[\| z^t - z^* \|^2\right] - \E\left[\| z^{t+1} - z^* \|^2\right] - \frac{\gamma^2}{4}\E\left[\| F(z^t)\|^2\right] + \frac{\gamma^2}{2}\E\left[\| g^t - F(z^t)\|^2\right] \notag\\
 &+ \frac{10 \sigma^2 \gamma^2 }{bM} + 16\varepsilon_0 \gamma\left( Q + L \Omega_z +
2 L_{\max} \varepsilon_0 + \frac{\sigma}{\sqrt{bM}}\right)+  8\varepsilon_0(1 + \gamma L_{\max}) \Omega_z 
\notag\\
&+ 8 (1 + 5\gamma^2 L_{\max}^2) \varepsilon_0^2.
 \end{align*}
The term $\| \bar g^t - F(z^t)\|^2$ was estimated, when we deduced \eqref{t432dd}. Then 
 \begin{align*}
\frac{\gamma^2}{4}\E\left[\| F(z^t)\|^2\right]&\leq \E\left[\| z^t - z^* \|^2\right] - \E\left[\| z^{t+1} - z^* \|^2\right]  + \frac{11 \sigma^2 \gamma^2 }{bM}\notag\\
 &\hspace{0.4cm}+ 16\varepsilon_0 \gamma\left( Q + L \Omega_z +
2 L_{\max} \varepsilon_0 + \frac{\sigma}{\sqrt{bM}}\right) \\
&\hspace{0.4cm}+  8\varepsilon_0(1 + \gamma L_{\max}) \Omega_z + 8 (1 + 6\gamma^2 L_{\max}^2) \varepsilon_0^2.
 \end{align*}
 Summing over all $t$ from $0$ to $k-1$:
\begin{align*}
\E\left[\frac{1}{k} \sum\limits_{t=0}^{k-1}\left\|  F(z^t) \right\|^2\right]  \leq&  \frac{4\E\left[\| z^0 - z^* \|^2\right]}{\gamma^2 k} +  \frac{44 \sigma^2}{bM} + \frac{64\varepsilon_0}{\gamma} \left( Q + L \Omega_z +
2 L_{\max} \varepsilon_0 + \frac{\sigma}{\sqrt{bM}}\right)\\
&+  \frac{32\varepsilon_0}{\gamma^2}(1 + \gamma L_{\max}) \Omega_z + \frac{32}{\gamma^2} (1 + 6\gamma^2 L_{\max}^2) \varepsilon_0^2.  
 \end{align*}
Let $\gamma = \frac{1}{4L}$ and $\varepsilon_0 = \mathcal{O}\left(\frac{\varepsilon}{\Omega_z L_{\max} L}\right)$, where $\varepsilon = \max\left(\frac{L^2 \Omega_z^2}{k}; \frac{\sigma^2}{bM}\right)$. Then for the output of Algorithm \ref{alg3} it holds
\begin{equation*}
\E\left[\frac{1}{k} \sum\limits_{t=0}^{k-1} \left\|  F(z^t) \right\|^2\right]  = \mathcal{O} \left(\frac{\E\left[L^2\| z^0 - z^* \|^2\right]}{k} +  \frac{\sigma^2}{bM}\right).
 \end{equation*}
Substituting the batch size $b$ and the number of iterations $k$ from the description of Algorithm \ref{alg2} and Corollary  \ref{cor45}:
\begin{equation*}
    \E\left[ \frac{1}{k}\sum\limits_{t=0}^{k-1}\|F(z^t)\|^2\right] = \mathcal{\tilde O}\left(\frac{L^2 \Omega_z^2 \sqrt{\chi}}{K} + \frac{\sigma^2 K}{MT\sqrt{\chi}} \right).
\end{equation*}
\EndProof

\section{Proof of Theorems from Section \ref{sec:na}} \label{lsgd}

Here we also introduce auxiliary sequences (Algorithm \ref{alg4} does not compute them):
\begin{eqnarray}
\label{seq}
\bar z^t = \frac{1}{M} \sum\limits_{m=1}^M z^{t}_m, &&
\bar g^t = \frac{1}{M} \sum\limits_{m=1}^M F_m(z^t_m, \xi^t_m), \notag\\
\bar z^{t+1/2} = \bar z^t - \gamma \bar g^t, && \bar z^{t+1} = \bar z^t - \gamma \bar g^{t+1/2}
\end{eqnarray}

\subsection{Strongly convex-strongly concave problems}

\begin{theorem}[Theorem \ref{th7}]
Let $\{ z^t_m\}_{t \geq 0}$ denote the iterates of Algorithm~\ref{alg4} for solving the problem \eqref{distr}. Let Assumptions \ref{as:as1l}, \ref{ass:as2g}, \ref{as:as3} and \ref{as:as5} be satisfied. Also let $H = \max_p |k_{p+1} - k_p|$ is a maximum distance between moments of communication ($k_p \in I$). Then if $\gamma \leq \frac{1}{21HL_{\max}}$, we have the following estimate:
\begin{align*}
 \E[\|\bar z^{T} - z^* \|^2 ] =\mathcal{\tilde O} \left( \exp\left(- \frac{\mu K}{42 H L_{\max}}\right) \|z^0 - z^* \|^2 + \frac{\sigma^2}{\mu^2 M T} + \frac{(D^2 H + \sigma^2) H L_{\max}^2}{\mu^4 T^2}\right).
 \end{align*}
\end{theorem}

We start our proof with the following lemma.

\begin{lemma} 
\label{l11}
Let $z,y \in \mathbb{R}^n$. We set $z^+ = z - y$, then for all $u \in \mathbb{R}^n$:
$$\|z^+ - u\|^2 \leq \|z - u \|^2 - 2 \langle y, z^+ - u\rangle - \|z^+ - z \|^2.$$
\end{lemma}
\textbf{Proof:} Simple manipulations give
 \begin{eqnarray*}
 \|z^+ - u \|^2 &=& \|z^+ - z + z - u \|^2 \\
&=& \|z-u \|^2 + 2 \langle z^+ - z, z - u \rangle + \|z^+ -z \|^2 \\
&=& \|z-u \|^2 + 2 \langle z^+ - z, z^+ - u \rangle - \|z^+ - z \|^2 \\
&=& \|z-u \|^2 + 2 \langle z^+ - (z - y), z^+ - u \rangle - 2 \langle y, z^+ - u \rangle - \|z^+ - z \|^2 \\
&=& \|z - u\|^2 - 2 \langle y, z^+ - u \rangle - \|z^+ - z \|^2.
 \end{eqnarray*}
\EndProof

\textbf{Proof of Theorem: } 
Applying this Lemma with $z = \bar z^{t+1}$,  $z=\bar z^{t}$, $u = z^*$ and $y = \gamma \bar g^{t+1/2}$, we get
\begin{equation*}
 \|\bar z^{t+1} - z^* \|^2 = \|\bar z^t - z^* \|^2 - 2 \gamma \langle \bar g^{t+1/2}, \bar z^{t+1} - z^* \rangle - \|\bar z^{t+1} - \bar z^t \|^2,
 \end{equation*}
and with $z = \bar z^{t+1/2}$,  $z=\bar z^{t}$, $u = z^{t+1}$, $y = \gamma \bar g^{t}$:
\begin{equation*}
 \|\bar z^{t+1/2} - \bar z^{t+1} \|^2 = \|\bar z^t - \bar z^{t+1} \|^2 - 2 \gamma \langle \bar g^{t}, \bar z^{t+1/2} - \bar z^{t+1} \rangle - \|\bar z^{t+1/2} - \bar z^t \|^2.
 \end{equation*}
Next, we sum up the two previous equalities
 \begin{align*}
 \|\bar z^{t+1} - z^* \|^2 + \|\bar z^{t+1/2} - \bar z^{t+1} \|^2 =& \|\bar z^t - z^* \|^2 - \|\bar z^{t+1/2} - \bar z^t \|^2 \\
 & - 2 \gamma \langle \bar g^{t+1/2}, \bar z^{t+1} - z^* \rangle - 2 \gamma \langle \bar g^{t}, \bar z^{t+1/2} - \bar z^{t+1} \rangle.
 \end{align*}
A small rearrangement gives
\begin{align*}
 \|\bar z^{t+1} - z^* \|^2 + \|\bar z^{t+1/2} - \bar z^{t+1} \|^2 =& \|\bar z^t - z^* \|^2  - \|\bar z^{t+1/2} - \bar z^t \|^2 \\ 
 & - 2 \gamma \langle \bar g^{t+1/2}, \bar z^{t+1/2} - z^* \rangle \\
 &+ 2 \gamma \langle \bar g^{t+1/2} - \bar g^{t}, \bar z^{t+1/2} - \bar z^{t+1} \rangle \\
 \leq& \|\bar z^t - z^* \|^2  - \|\bar z^{t+1/2} - \bar z^t \|^2 \\ 
 & - 2 \gamma \langle \bar g^{t+1/2}, \bar z^{t+1/2} - z^* \rangle + \gamma^2 \|\bar g^{t+1/2} - \bar g^{t}\|^2 \\
 &+ \| \bar z^{t+1/2} - \bar z^{t+1}\|^2,
 \end{align*}
Then we take the total expectation of both sides of the equation
\begin{eqnarray}
\label{1lsgd}
 \E\left[\|\bar z^{t+1} - z^* \|^2\right]  =& \E\left[\|\bar z^t - z^* \|^2\right] - \E\left[\|\bar z^{t+1/2} - \bar z^t \|^2\right] \notag\\
 &- 2 \gamma \E\left[\langle \bar g^{t+1/2}, \bar z^{t+1/2} - z^* \rangle\right] + \gamma^2 \E\left[\|\bar g^{t+1/2} - \bar g^{t}\|^2\right].  
 \end{eqnarray}
Further, we need to additionally estimate two terms $- 2 \gamma \langle \bar g^{t+1/2}, \bar z^{t+1/2} - z^* \rangle$ and $\|\bar g^{t+1/2} - \bar g^{t}\|^2$. For this we prove the following two lemmas, but before that we introduce the additional notation:
\begin{eqnarray}
\label{err}
 \text{Err}(t) = \frac{1}{M}\sum\limits_{m=1}^M \|\bar z^{t} - z_m^{t}\|^2.  
 \end{eqnarray}
\begin{lemma} \label{lem12}
The following estimate is valid:
 \end{lemma}
 \begin{eqnarray}
 \label{2}
 - 2 \gamma \E\left[\langle \bar g^{t+1/2}, \bar z^{t+1/2} - z^* \rangle\right] \leq - \gamma \mu \E\left[\| \bar z^{t+1/2} - z^* \|^2\right]  + \frac{\gamma L^2_{\max}}{\mu} \E\left[\text{Err}(t+1/2) \right].
  \end{eqnarray}

\textbf{Proof:} We take into account the independence of all random vectors $\xi^{i} = (\xi^{i}_1, \ldots , \xi^{i}_m)$ and select only the conditional expectation $\E_{\xi^{t+1/2}}$ on vector $\xi^{t+1/2}$:
 \begin{eqnarray*}
  &&\hspace{-2cm}- 2 \gamma \E\left[\langle \bar g^{t+1/2}, \bar z^{t+1/2} - z^* \rangle\right] \\
  &=& - 2 \gamma \E\left[\left\langle  \frac{1}{M} \sum\limits_{m=1}^M \E_{\xi^{t+1/2}}[F_m(z_m^{t+1/2}, \xi_m^{t+1/2})], \bar z^{t+1/2} - z^* \right\rangle \right] \\
  &\overset{\eqref{as3}}{=}& -2 \gamma \E\left[\left\langle  \frac{1}{M} \sum\limits_{m=1}^M F_m(z_m^{t+1/2}), \bar z^{t+1/2} - z^* \right\rangle\right] \\
  &=& - 2 \gamma \E\left[\left\langle  \frac{1}{M} \sum\limits_{m=1}^M F_m(\bar z^{t+1/2}), \bar z^{t+1/2} - z^* \right\rangle\right] \\
  &&+ 2 \gamma \E\left[\left\langle  \frac{1}{M} \sum\limits_{m=1}^M [F_m(\bar z^{t+1/2}) - F_m(z_m^{t+1/2})], \bar z^{t+1/2} - z^* \right\rangle\right] \\
  &=& - 2 \gamma \E\left[\left\langle  F(\bar z^{t+1/2}), \bar z^{t+1/2} - z^* \right\rangle\right] \\
  &&+ 2 \gamma \E\left[\left\langle  \frac{1}{M} \sum\limits_{m=1}^M [F_m(\bar z^{t+1/2}) - F_m(z_m^{t+1/2})], \bar z^{t+1/2} - z^* \right\rangle\right].
  \end{eqnarray*}
Using the property of $z^*$, we have:
 \begin{eqnarray*}
 &&\hspace{-2cm}
  - 2 \gamma \E\left[\langle \bar g^{t+1/2}, \bar z^{t+1/2} - z^* \rangle\right]\\
  &=& - 2 \gamma \E\left[\left\langle  F(\bar z^{t+1/2}) - F(z^*), \bar z^{t+1/2} - z^* \right\rangle\right] \\
  &&+ 2 \gamma \E\left[\left\langle  \frac{1}{M} \sum\limits_{m=1}^M [F_m(\bar z^{t+1/2}) - F_m(z_m^{t+1/2})], \bar z^{t+1/2} - z^* \right\rangle\right] \\
  &\overset{\eqref{as2g}}{\leq}& - 2 \gamma \mu \E\left[\| \bar z^{t+1/2} - z^* \|^2\right] \\
  &&+ 2 \gamma \E\left[\left\langle  \frac{1}{M} \sum\limits_{m=1}^M [F_m(\bar z^{t+1/2}) - F_m(z_m^{t+1/2})], \bar z^{t+1/2} - z^* \right\rangle\right].
  \end{eqnarray*}
For $c >0$ it is true that $2\langle a , b\rangle \leq \frac{1}{c} \|a\|^2 + c \|b\|^2$, then
\begin{eqnarray*}
  &&\hspace{-2cm}- 2 \gamma \E\left[\langle \bar g^{t+1/2}, \bar z^{t+1/2} - z^* \rangle\right] \\
  &\leq& - 2 \gamma \mu \E\left[\| \bar z^{t+1/2} - z^* \|^2\right] \\
  &&+ \gamma \mu \E\left[\left\| \bar z^{t+1/2} - z^* \right\|^2\right] + \frac{\gamma}{\mu} \E\left[\left\| \frac{1}{M} \sum\limits_{m=1}^M [F_m(\bar z^{t+1/2}) - F_m(z_m^{t+1/2})] \right\|^2\right] \\
  &=& -\gamma \mu \E\left[\| \bar z^{t+1/2} - z^* \|^2\right]   + \frac{\gamma}{\mu M^2} \E\left[\left\| \sum\limits_{m=1}^M [F_m(\bar z^{t+1/2}) - F_m(z_m^{t+1/2})] \right\|^2\right] \\
  &\overset{\eqref{eq:squared_sum}}{\leq}& -\gamma \mu \E\left[\| \bar z^{t+1/2} - z^* \|^2\right]   + \frac{\gamma}{\mu M} \E\left[\sum\limits_{m=1}^M\left\| F_m(\bar z^{t+1/2}) - F_m(z_m^{t+1/2})\right\|^2\right] \\
  &\overset{\eqref{as1l}}{\leq}& -\gamma \mu \E\left[\| \bar z^{t+1/2} - z^* \|^2\right]  + \frac{\gamma L_{\max}^2}{\mu M} \E\left[ \sum\limits_{m=1}^M\left\|\bar z^{t+1/2} - z_m^{t+1/2}\right\|^2 \right].
  \end{eqnarray*}
Definition \eqref{err} ends the proof.
\EndProof

\begin{lemma} \label{lem13}
The following estimate is valid:
  \end{lemma}
 \begin{align}
 \label{3}
 \E\left[\|\bar g^{t+1/2} - \bar g^{t}\|^2\right]  &\leq 5L_{\max}^2 \E\left[ \|\bar z^{t+1/2} - \bar z^{t}\|^2\right] + \frac{10 \sigma^2}{M} \nonumber\\
 &\hspace{0.4cm}+ 5L_{\max}^2\E\left[\text{Err}(t+1/2) \right] +5L_{\max}^2\E\left[\text{Err}(t) \right].
  \end{align}
\textbf{Proof:} We make the following chain:
 \begin{eqnarray*}
 &\hspace{-1cm}\E\big[\|\bar g^{t+1/2} - & \bar g^{t}\|^2\big] \\
 &=& \E\left[\left\|\frac{1}{M}\sum\limits_{m=1}^M F_m(z^{t+1/2}_m, \xi^{t+1/2}_m) - \frac{1}{M}\sum\limits_{m=1}^M F_m(z^t_m, \xi^{t}_m)\right\|^2\right] \\
 &\overset{\eqref{eq:squared_sum}}{\leq}& 5 \E\left[\left\|\frac{1}{M}\sum\limits_{m=1}^M [F_m(z^{t+1/2}_m, \xi^{t+1/2}_m) - F_m(z^{t+1/2}_m)  ]\right\|^2\right]  \\
 && + 5  \E\left[\left\|\frac{1}{M}\sum\limits_{m=1}^M [F_m(z^{t}_m, \xi^k_m) - F_m(z^{t}_m)  ]\right\|^2\right] \\
 &&+5\E\left[\left\|\frac{1}{M}\sum\limits_{m=1}^M [ F_m(z^{t+1/2}_m) - F_m(\bar z^{t+1/2})  ]\right\|^2\right] \\
 && + 5\E\left[\left\|\frac{1}{M}\sum\limits_{m=1}^M [ F_m(z^{t}_m) - F_m(\bar z^{t})  ]\right\|^2\right] \\
 &&+ 5\E\left[\left\|\frac{1}{M}\sum\limits_{m=1}^M [ F_m(\bar z^{t+1/2}) - F_m(\bar z^{t})  ]\right\|^2\right]\\
 &\overset{\eqref{eq:squared_sum}}{\leq}& 5 \E\left[\left\|\frac{1}{M}\sum\limits_{m=1}^M [F_m(z^{t+1/2}_m, \xi^{t+1/2}_m) - F_m(z^{t+1/2}_m)  ]\right\|^2\right] \\
 && + 5  \E\left[\left\|\frac{1}{M}\sum\limits_{m=1}^M [F_m(z^{t}_m, \xi^k_m) - F_m(z^{t}_m)  ]\right\|^2\right] \\
 &&
 +\frac{5}{M}\sum\limits_{m=1}^M \E\left[\left\| F_m(z^{t+1/2}_m) - F_m(\bar z^{t+1/2})\right\|^2\right] \\
 &&  + \frac{5}{M} \sum\limits_{m=1}^M\E\left[\left\| F_m(z^{t}_m) - F_m(\bar z^{t})  \right\|^2\right] + 5\E\left[\left\|F(\bar z^{t+1/2}) - F(\bar z^{t})\right\|^2\right] \\
 &\overset{\eqref{as1l},\eqref{err}}{\leq}& 5\E\left[\left\|\frac{1}{M}\sum\limits_{m=1}^M [F_m(z^{t+1/2}_m, \xi^{t+1/2}_m) - F_m(z^{t+1/2}_m)  ]\right\|^2\right]  \\
 && + 5  \E\left[\left\|\frac{1}{M}\sum\limits_{m=1}^M [F_m(z^{t}_m, \xi^k_m) - F_m(z^{t}_m)  ]\right\|^2\right] \\
 && +5L_{\max}^2\E\left[\text{Err}(t+1/2) \right] +5L_{\max}^2\E\left[\text{Err}(t) \right] + 5 L_{\max}^2\E\left[ \|\bar z^{t+1/2} - \bar z^{t}\|^2 \right] \\
 &=& 5\E\left[\E_{\xi_{t+1/2}}\left[\left\|\frac{1}{M}\sum\limits_{m=1}^M [F_m(z^{t+1/2}_m, \xi^{t+1/2}_m) - F_m(z^{t+1/2}_m)  ]\right\|^2\right]\right]  \\
 &&+ 5  \E\left[\E_{\xi_{t}}\left[\left\|\frac{1}{M}\sum\limits_{m=1}^M [F_m(z^{t}_m, \xi^t_m) - F_m(z^{t}_m)  ]\right\|^2\right]\right] \\
 && +5L_{\max}^2\E\left[\text{Err}(t+1/2) \right] +5L_{\max}^2\E\left[\text{Err}(t) \right] + 5 L_{\max}^2\E\left[ \|\bar z^{t+1/2} - \bar z^{t}\|^2 \right].
  \end{eqnarray*}
 Using the independence of each machine and \eqref{as3}, we get:
  \begin{align*}
 \E\left[\|\bar g^{t+1/2} - \bar g^{t}\|^2\right]  \leq& \frac{10 \sigma^2}{M} +5L_{\max}^2\E\left[\text{Err}(t+1/2) \right] \\
 &+5L_{\max}^2\E\left[\text{Err}(t) \right] + 5 L_{\max}^2\E\left[ \|\bar z^{t+1/2} - \bar z^{t}\|^2 \right].
  \end{align*}
\EndProof
We are now ready to combine \eqref{1lsgd}, \eqref{2}, \eqref{3} and get
\begin{align}
\label{an6}
\E\left[\|\bar z^{t+1} - z^* \|^2\right]  \leq& \E\left[\|\bar z^t - z^* \|^2\right] - \E\left[\|\bar z^{t+1/2} - \bar z^t \|^2\right] \notag\\
 &- \gamma \mu \E\left[\| \bar z^{t+1/2} - z^* \|^2\right]  + \frac{\gamma L_{\max}^2}{\mu} \E\left[\text{Err}(t+1/2) \right] \notag\\
 &+ 5\gamma^2 L_{\max}^2 \E\left[\|\bar z^{t+1/2} - \bar z^{t}\|^2 \right] + \frac{10 \gamma^2\sigma^2}{M} \notag\\ 
 &+ 5\gamma^2 L_{\max}^2 \E\left[\text{Err}(t+1/2) \right] +5\gamma^2 L_{\max}^2 \E\left[\text{Err}(t)\right].
\end{align}
Together with $-\|\bar z^{t+1/2} - z^* \|^2 \leq \|\bar z^{t+1/2} - \bar z^t \|^2 - \nicefrac{1}{2} \| \bar z^t - z^*\|^2$ it transforms to
\begin{align*}
\E\left[\|\bar z^{t+1} - z^* \|^2\right]  \leq& \left(1 - \frac{\mu \gamma}{2} \right) \E\left[\|\bar z^t - z^* \|^2\right] + \frac{10 \gamma^2\sigma^2}{M}  \\
&+ (\mu \gamma + 5\gamma^2 L_{\max}^2-1) \|\bar z^{t+1/2} - \bar z^t \|^2 \\
 &  + \frac{\gamma L_{\max}^2}{\mu} \E\left[\text{Err}(t+1/2) \right] + 5\gamma^2 L_{\max}^2\E\left[\text{Err}(t+1/2) \right] \\
 &+5\gamma^2 L_{\max}^2\E\left[\text{Err}(t) \right].
\end{align*}
Taking $\gamma \leq \frac{1}{6HL_{\max}}$ gives
\begin{align}
\label{r405}
\E\left[\|\bar z^{t+1} - z^* \|^2\right]  \leq& \left(1 - \frac{\mu \gamma}{2} \right) \E\left[\|\bar z^t - z^* \|^2\right] + \frac{10 \gamma^2\sigma^2}{M}  \nonumber\\
&+ \frac{7 \gamma L_{\max}^2}{\mu} \E\left[\text{Err}(t+1/2)\right] +5\gamma^2 L_{\max}^2\E\left[\text{Err}(t) \right].
\end{align}
It remains to estimate $\E\left[\text{Err}(t+1/2)\right]$ and $ \E\left[\text{Err}(t)\right]$.

\begin{lemma} \label{lem14}
For $t \in [t_p + 1; t_{p+1}]$ the following estimate is valid:
  \end{lemma}
 \begin{eqnarray}
 \label{5}
 \E\left[\text{Err}(t+1/2)\right] &\leq  216 (D^2 H + \sigma^2) H \gamma^2.
  \end{eqnarray}
\textbf{Proof:}
First, let us look at the nearest past consensus point $t_p < t$, then $z^{t_p + 1}_m = \bar z^{t_p + 1}$:
\begin{eqnarray*}
 \E\left[\text{Err}(t+1/2)\right] &=& \frac{1}{M}\sum\limits_{m=1}^M \E\|\bar z^{t+1/2} - z_m^{t+1/2}\|^2 \\
 &=& \frac{1}{M}\sum\limits_{m=1}^M \E\|\bar z^{t+1/2} - \bar z^{t_p}  + z^{t_p}_m  - z_m^{t+1/2}\|^2 \\
 &=& \frac{\gamma^2}{M}\sum\limits_{m=1}^M \E\left\|F_m(z_m^{t}, \xi_m^{t}) -\bar g^{t}  +  \sum\limits_{k=t_p + 1}^{t-1} [F_m(z_m^{k+1/2}, \xi_m^{k+1/2}) - \bar g^{k+1/2}]\right\|^2.
 \end{eqnarray*}
Only $\bar g^{t}$ and $F_m(z_m^{k}, \xi_m^{k})$ depend on $\xi^{k}$, as well as the unbiasedness of $\bar g^{t}$ and $F_m(z_m^{k}, \xi_m^{k})$, we have
\begin{align*}
 \E\left[\text{Err}(t+1/2)\right] =& \frac{\gamma^2}{M}\sum\limits_{m=1}^M \E\Bigg\|-\frac{1}{M}\sum\limits_{i=1}^M  F_i(z_i^{t})-\sum\limits_{k=t_p + 1}^{t-1}\bar g^{k+1/2} 
 \\
 &+ F_m(z_m^{t}) + \sum\limits_{k=t_p + 1}^{t-1} F_m(z_m^{k+1/2}, \xi_m^{k+1/2})\Bigg\|^2 \\
 &+ \frac{\gamma^2}{M}\sum\limits_{m=1}^M \E\left\|\frac{1}{M}\sum\limits_{i=1}^M  F_i(z_i^{t}) -\bar g^{t} - F_m(z_m^{t}) + F_m(z_m^{t}, \xi_m^{t}) \right\|^2.
 \end{align*}
We continue the same way, but note that $z^{t}_i$ depends on $\xi^{k-1 + 1/2}$, then let us make the estimate rougher than in the previous case
\begin{align*}
 \E\left[\text{Err}(t+1/2)\right] &\leq (1 + \beta_0)\frac{\gamma^2}{M}\sum\limits_{m=1}^M \E\left\|-\sum\limits_{k=t_p + 1}^{t-1}\bar g^{k+1/2}  + \sum\limits_{k=t_p + 1}^{t-1} F_m(z_m^{k+1/2}, \xi_m^{k+1/2})\right\|^2 \\
 &\hspace{0.4cm}+ (1 + \beta^{-1}_0)\frac{\gamma^2}{M}\sum\limits_{m=1}^M \E\left\|-\frac{1}{M}\sum\limits_{i=1}^M  F_i(z_i^{t}) + F_m(z_m^{t}) \right\|^2 \\
 &\hspace{0.4cm}+ \frac{\gamma^2}{M}\sum\limits_{m=1}^M \E\left\|\frac{1}{M}\sum\limits_{i=1}^M  F_i(z_i^{t}) -\bar g^{t} - F_m(z_m^{t}) + F_m(z_m^{t}, \xi_m^{t}) \right\|^2.
 \end{align*}
Here $\beta_0$ is some positive constant, which we define later. Then 
\begin{align*}
 \E&\left[\text{Err}(t+1/2)\right] \\
 &\leq (1 + \beta_0)\frac{\gamma^2}{M}\sum\limits_{m=1}^M \E\Bigg\|-\frac{1}{M}\sum\limits_{i=1}^M  F_i(z_i^{t-1+1/2}) -\sum\limits_{k=t_p + 1}^{t-2}\bar g^{k+1/2} \\
 &\hspace{3.5cm}+ F_m(z_m^{t-1+1/2}, \xi_m^{t-1+1/2}) + \sum\limits_{k=t_p + 1}^{t-2} F_m(z_m^{k+1/2}, \xi_m^{k+1/2})\Bigg\|^2 \\
 &\hspace{0.4cm}+ (1 + \beta^{-1}_0)\frac{\gamma^2}{M}\sum\limits_{m=1}^M \E\left\|-\frac{1}{M}\sum\limits_{i=1}^M  F_i(z_i^{t}) + F_m(z_m^{t}) \right\|^2 \\
 &\hspace{0.4cm}+ (1 + \beta_0) \frac{\gamma^2}{M}\sum\limits_{m=1}^M \E\Bigg\|\frac{1}{M}\sum\limits_{i=1}^M  F_i(z_i^{t-1+1/2}) -\bar g^{t-1+1/2} 
 \\
 &\hspace{4cm}- F_m(z_m^{t-1+1/2}) + F_m(z_m^{t-1+1/2}, \xi_m^{t-1+1/2}) \Bigg\|^2\\
 &\hspace{0.4cm}+ \frac{\gamma^2}{M}\sum\limits_{m=1}^M \E\left\|\frac{1}{M}\sum\limits_{i=1}^M  F_i(z_i^{t}) -\bar g^{t} - F_m(z_m^{t}) + F_m(z_m^{t}, \xi_m^{t}) \right\|^2.
 \end{align*}
and
\begin{align*}
 \E&\left[\text{Err}(t+1/2)\right] \\
 &\leq (1 + \beta_0)(1 + \beta_1)\frac{\gamma^2}{M}\sum\limits_{m=1}^M \E\Bigg\|-\sum\limits_{k=t_p + 1}^{t-2}\bar g^{k+1/2} + \sum\limits_{k=t_p + 1}^{t-2} F_m(z_m^{k+1/2}, \xi_m^{k+1/2})\Bigg\|^2 \\
 &\hspace{0.4cm}+ (1 + \beta_0)(1 + \beta^{-1}_1)\frac{\gamma^2}{M}\sum\limits_{m=1}^M \E\Bigg\|-\frac{1}{M}\sum\limits_{i=1}^M  F_i(z_i^{t-1+1/2}) + F_m(z_m^{t-1+1/2}) \Bigg\|^2 \\
 &\hspace{0.4cm}+ (1 + \beta^{-1}_0)\frac{\gamma^2}{M}\sum\limits_{m=1}^M \E\left\|-\frac{1}{M}\sum\limits_{i=1}^M  F_i(z_i^{t}) + F_m(z_m^{t}) \right\|^2 \\
 &\hspace{0.4cm}+ (1 + \beta_0) \frac{\gamma^2}{M}\sum\limits_{m=1}^M \E\Bigg\|\frac{1}{M}\sum\limits_{i=1}^M  F_i(z_i^{t-1+1/2}) -\bar g^{t-1+1/2} \\
 &\hspace{4cm}- F_m(z_m^{t-1+1/2}) + F_m(z_m^{t-1+1/2}, \xi_m^{t-1+1/2}) \Bigg\|^2\\
 &\hspace{0.4cm}+ \frac{\gamma^2}{M}\sum\limits_{m=1}^M \E\left\|\frac{1}{M}\sum\limits_{i=1}^M  F_i(z_i^{t}) -\bar g^{t} - F_m(z_m^{t}) + F_m(z_m^{t}, \xi_m^{t}) \right\|^2.
 \end{align*}
One can continue this way for all terms, setting $\beta_i = \frac{1}{\alpha - i-1}$, where $\alpha = 4 H$. Then for all $i = 0, \ldots, (t- t_p - 1)$
$$(1 + \beta_0)(1 + \beta_1)(1 + \beta_2)\ldots (1 - \beta_i) = \frac{\alpha}{\alpha - i-1}.$$ 
Note that $t- t_p \leq 2H$, hence for all $i = 0, \ldots, (t- t_p - 1)$
$$(1 + \beta_0)(1 + \beta_1)(1 + \beta_2)\ldots (1 + \beta_{i}) \leq (1 + \beta_1)(1 + \beta_2)\ldots (1 + \beta_{t- t_p - 1}) \leq \frac{\alpha}{\alpha - 2 H} \leq 2.$$ 
Additionally, $1 + \beta_i^{-1} \leq \alpha$, then ($\alpha = 4 H$)
\begin{align*}
 \E\left[\text{Err}(t+1/2)\right]
 &\leq \frac{2 \alpha \gamma^2}{M} \sum\limits_{k=t_p + 1}^{t-1} \sum\limits_{m=1}^M \E\Bigg\|-\frac{1}{M}\sum\limits_{i=1}^M  F_i(z_i^{k+1/2}) + F_m(z_m^{k+1/2}) \Bigg\|^2 \\
 &\hspace{0.4cm}+ \frac{2 \alpha\gamma^2}{M}\sum\limits_{m=1}^M \E\left\|-\frac{1}{M}\sum\limits_{i=1}^M  F_i(z_i^{t}) + F_m(z_m^{t}) \right\|^2 \\
 &\hspace{0.4cm}+ \frac{2\gamma^2}{M} \sum\limits_{k=t_p + 1}^{t-1} \sum\limits_{m=1}^M \E\Bigg\|\frac{1}{M}\sum\limits_{i=1}^M  F_i(z_i^{k+1/2}) -\bar g^{k+1/2} \\
 &\hspace{4cm}
 - F_m(z_m^{k+1/2}) + F_m(z_m^{k+1/2}, \xi_m^{k+1/2}) \Bigg\|^2\\
 &\hspace{0.4cm}+ \frac{2\gamma^2}{M}\sum\limits_{m=1}^M \E\Bigg\|\frac{1}{M}\sum\limits_{i=1}^M  F_i(z_i^{t}) -\bar g^{t} - F_m(z_m^{t}) + F_m(z_m^{t}, \xi_m^{t}) \Bigg\|^2 \\
 &= \frac{8 \gamma^2 H}{M} \sum\limits_{k=t_p + 1}^{t-1} \sum\limits_{m=1}^M \E\Bigg\|-\frac{1}{M}\sum\limits_{i=1}^M  F_i(z_i^{k+1/2}) + F_m(z_m^{k+1/2}) \Bigg\|^2 \\
 &\hspace{0.4cm}+ \frac{8\gamma^2 H}{M}\sum\limits_{m=1}^M \E\left\|-\frac{1}{M}\sum\limits_{i=1}^M  F_i(z_i^{t}) + F_m(z_m^{t}) \right\|^2 \\
 &\hspace{0.4cm}+ \frac{8\gamma^2}{M} \sum\limits_{k=t_p + 1}^{t-1} \sum\limits_{m=1}^M \E\Bigg\|\frac{1}{M}\sum\limits_{i=1}^M  F_i(z_i^{k+1/2}) -\bar g^{k+1/2} 
 \\
 &\hspace{4cm}- F_m(z_m^{k+1/2}) + F_m(z_m^{k+1/2}, \xi_m^{k+1/2}) \Bigg\|^2\\
 &\hspace{0.4cm}+ \frac{8\gamma^2 }{M}\sum\limits_{m=1}^M \E\left\|\frac{1}{M}\sum\limits_{i=1}^M  F_i(z_i^{t}) -\bar g^{t} - F_m(z_m^{t}) + F_m(z_m^{t}, \xi_m^{t}) \right\|^2.
 \end{align*}
It remains to estimate 
\begin{align*}
\frac{1}{M}\sum\limits_{m=1}^M \E\Bigg\|-\frac{1}{M}&\sum\limits_{i=1}^M  F_i(z_i^{k+1/2}) + F_m(z_m^{k+1/2}) \Bigg\|^2 \\
&\overset{\eqref{eq:squared_sum}}{\leq} \frac{3}{M}\sum\limits_{m=1}^M \E\Bigg\|-\frac{1}{M}\sum\limits_{i=1}^M  F_i(z_i^{k+1/2}) + \frac{1}{M}\sum\limits_{i=1}^M  F_i(\bar z^{k+1/2}) \Bigg\|^2 \\
&\hspace{0.4cm}+  \frac{3}{M}\sum\limits_{m=1}^M \E\Bigg\|- \frac{1}{M}\sum\limits_{i=1}^M  F_i(\bar z^{k+1/2}) +  F_m(\bar z^{k+1/2})\Bigg\|^2 \\
&\hspace{0.4cm}+ \frac{3}{M}\sum\limits_{m=1}^M \E\Bigg\|- F_m(\bar z^{k+1/2}) + F_m(z_m^{k+1/2}) \Bigg\|^2 \\
&\overset{\eqref{as4}}{\leq} \frac{6}{M}\sum\limits_{m=1}^M \E\Bigg\|- F_m(\bar z^{k+1/2}) + F_m(z_m^{k+1/2}) \Bigg\|^2 +  3 D^2
\\
&\overset{\eqref{as1l}}{\leq} \frac{6 L^2_{\max}}{M}\sum\limits_{m=1}^M \E\|\bar z^{k+1/2} - z_m^{k+1/2} \|^2 +  3 D^2 \\
&\overset{}{=} 6 L^2_{\max} \E\left[\text{Err}(k+1/2)\right] +  3 D^2
\end{align*}
and
\begin{align*}
 \frac{1}{M} \sum\limits_{m=1}^M \E\Bigg\|\frac{1}{M}\sum\limits_{i=1}^M  F_i(z_i^{k+1/2}) &-\bar g^{k+1/2} - F_m(z_m^{k+1/2}) + F_m(z_m^{k+1/2}, \xi_m^{k+1/2}) \Bigg\|^2 \\
 &\overset{\eqref{eq:squared_sum}}{\leq} 2 \Bigg[ \E\left\| \frac{1}{M}\sum\limits_{i=1}^M  F_i(z_i^{k+1/2}) -\bar g^{k+1/2} \right\|^2 \\
 &\hspace{0.4cm}+ \frac{2}{M} \sum\limits_{m=1}^M\E\left\|F_m(z_m^{k+1/2}) + F_m(z_m^{k+1/2}, \xi_m^{k+1/2}) \right\|^2 \Bigg]\\
 &\overset{\eqref{as3}}{\leq} 4 \sigma^2.
  \end{align*}
Finally, we get
\begin{align}
\label{r4041}
 \E\left[\text{Err}(t+1/2)\right] &\leq  48 \gamma^2 L^2_{\max} H \sum\limits_{k=t_p + 1}^{t-1}  \E\left[\text{Err}(k+1/2)\right] +  48\gamma^2 L^2_{\max}  H \E\left[\text{Err}(t)\right] \nonumber\\
 &\hspace{0.4cm} +32 (D^2 H + \sigma^2)  \sum\limits_{k=t_p + 1}^{t-1} \gamma^2 + 32\gamma^2 \left(\sigma^2 + D^2\right).
 \end{align}
The estimate for $\E\left[\text{Err}(t+1/3)\right]$ is done in a similar way:
\begin{align}
\label{r4042}
 \E\left[\text{Err}(t)\right] &\leq  48 \gamma^2 L^2_{\max} H \sum\limits_{k=t_p + 1}^{t-1}  \E\left[\text{Err}(k+1/2)\right] +32 (D^2 H + \sigma^2)  \sum\limits_{k=t_p + 1}^{t-1} \gamma^2 .
 \end{align}
Substituting $\E\left[\text{Err}(t)\right]$ to $\E\left[\text{Err}(t+1/2)\right]$, we get
\begin{align*}
 \E\left[\text{Err}(t+1/2)\right] &\leq  48 \gamma^2 L^2_{\max} H \sum\limits_{k=t_p + 1}^{t-1}  \E\left[\text{Err}(k+1/2)\right] \\
 &\hspace{0.4cm}+  48\gamma^2 L^2_{\max}  H \Bigg( 48 \gamma^2 L^2_{\max} H \sum\limits_{k=t_p + 1}^{t-1}  \E\left[\text{Err}(k+1/2)\right] \\
 &\hspace{0.4cm} +32 (D^2 H + \sigma^2)  \sum\limits_{k=t_p + 1}^{t-1} \gamma^2 \Bigg) \\
 &\hspace{0.4cm} +32 (D^2 H + \sigma^2)  \sum\limits_{k=t_p + 1}^{t-1} \gamma^2 + 32\gamma^2 \left(\sigma^2 + D^2\right).
 \end{align*}
With $\gamma \leq \frac{1}{21 H L_{\max}}$
\begin{align*}
 \E\left[\text{Err}(t+1/2)\right] &\leq  \frac{1}{8H} \sum\limits_{k=t_p + 1}^{t-1}  \E\left[\text{Err}(k+1/2)\right] +72 (D^2 H + \sigma^2) \gamma^2 (t - t_p - 1).
 \end{align*}
Let us run the recursion:
 \begin{align*}
 \E\left[\text{Err}(t+1/2)\right] &\leq  \frac{1}{8H} \left(1 + \frac{1}{8H}\right) \sum\limits_{k=t_p + 1}^{t-2}  \E\left[\text{Err}(k+1/2)\right]  \\
 &\hspace{0.4cm}+ \frac{1}{8H} \cdot 72 (D^2 H + \sigma^2)  \gamma^2 (t - t_p - 2)   \gamma^2 (t - t_p - 1)\\
 &\leq 72 (D^2 H + \sigma^2) \gamma^2   \sum\limits_{k=t_p + 1}^{t-1}  \left( 1 + \frac{1}{8H}\right)^{t - 1 - j} .
 \end{align*}
 Then one can note that $\left( 1 + \frac{1}{8H}\right)^{t - 1 - j} \leq \left( 1 + \frac{1}{2H}\right)^{2H} \leq \exp(1) \leq 3$ and then
\begin{align*}
\E\left[\text{Err}(t+1/2)\right] &\leq 216 (D^2 H + \sigma^2) \sum\limits_{k=t_p + 1}^{t-1} \gamma^2 \leq 216 (D^2 H + \sigma^2) H \gamma^2.
\end{align*}
\EndProof

Note that in the general case $\E\left[\text{Err}(t+1/3)\right]$ may be less than $\E\left[\text{Err}(t)\right]$, but since the recurrent \eqref{r4041} is stronger than \eqref{r4042}, we assume for the simplicity that $\E\left[\text{Err}(k+1/3)\right] \geq \E\left[\text{Err}(k)\right]$. Then \eqref{r405} can be rewritten as
\begin{align*}
\E\left[\|\bar z^{t+1} - z^* \|^2\right]  \leq& \left(1 - \frac{\mu \gamma}{2} \right) \E\left[\|\bar z^t - z^* \|^2\right] + \frac{10 \gamma^2\sigma^2}{M} \nonumber\\ & + \left(\frac{7 \gamma L_{\max}^2}{\mu} +5\gamma^2 L_{\max}^2\right)\E\left[\text{Err}(t+1/2)\right] \\
\leq& \left(1 - \frac{\mu \gamma}{2} \right) \E\left[\|\bar z^t - z^* \|^2\right] + \frac{10 \gamma^2\sigma^2}{M} \nonumber\\ & + \left(\frac{7 \gamma L_{\max}^2}{\mu} +5\gamma^2 L_{\max}^2\right)\left( 216 (D^2 H + \sigma^2) H \gamma^2 \right).
\end{align*}
Running the recursion, we obtain:
\begin{align*}
\E\left[\|\bar z^{T} - z^* \|^2\right]  &= \mathcal{O} \left( \left(1 - \frac{\mu \gamma}{2} \right)^T \|z^0 - z^* \|^2 + \frac{\gamma\sigma^2}{\mu M} + \frac{\gamma^2  (D^2 H + \sigma^2) H L_{\max}^2}{\mu^2} \right),
\end{align*}
or
\begin{align*}
\E\left[\|\bar z^{T} - z^* \|^2\right] &=\mathcal{O} \left( \exp\left(- \frac{\mu \gamma T}{2} \right) \|z^0 - z^* \|^2 + \frac{\gamma\sigma^2}{\mu M} + \frac{\gamma^2  (D^2 H + \sigma^2) H L_{\max}^2}{\mu^2} \right)
\end{align*}

Finally, we need tuning of $\gamma = \min\left\{\frac{1}{21 H L_{\max}}; \frac{2\ln\left( \max\{2, \mu  \|z^0 - z^* \|^2 T M/\sigma^2 \} \right)}{\mu T}\right\}$ to get
    \begin{align*}
   \mathcal{\tilde O} \left( \exp\left(- \frac{\mu T}{42 H L_{\max}}\right) \|z^0 - z^* \|^2 + \frac{\sigma^2}{\mu^2 M T} + \frac{(D^2 H + \sigma^2) H L_{\max}^2}{\mu^4 T^2}\right).  
    \end{align*}
\EndProof

\subsection{Non-convex-non-concave problems}

\begin{theorem}[Theorem \ref{th7}]
Let $\{ z^t_m\}_{t \geq 0}$ denote the iterates of Algorithm~\ref{alg4} for solving the problem \eqref{distr}. Let Assumptions \ref{as:as1l}, \ref{ass:as2n}, \ref{as:as3} and \ref{as:as5} be satisfied. Also let $H = \max_p |k_{p+1} - k_p|$ is a maximum distance between moments of communication ($k_p \in I$) and $\|\bar z^t\| \leq \Omega$ (for all $t$). Then if $\gamma \leq \frac{1}{4L_{\max}}$, we have the following estimate:
\begin{align*}
\E\left[ \frac{1}{T}\sum\limits_{t=0}^{T-1}\|F(\bar z^t)\|^2\right] &= \mathcal{O} \Bigg(\frac{L^2_{\max} \|\bar z^0 - z^* \|^2}{T}  + \frac{(L_{\max} \Omega (D^2 H + \sigma^2) H)^{2/3}}{T^{1/3}} \\
&\hspace{1cm}+ \frac{\sigma^2}{M}  + L_{\max} \Omega \sqrt{(D^2 H + \sigma^2) H}  \Bigg).
\end{align*}
\end{theorem}

\textbf{Proof:} Most of the necessary estimates have already been made in the previous subsection. In particular, Lemmas \ref{lem13} and \ref{lem14} are valid for us. But Lemma \ref{lem12} needs modification:

\begin{lemma} 
The following estimate is valid:
 \end{lemma}
 \begin{align}
 \label{temp1005}
 - 2 \gamma \E\left[\langle \bar g^{t+1/2}, \bar z^{t+1/2} - z^* \rangle\right] &\leq  2 \gamma L_{\max} \sqrt{\E\left[ \| \bar z^{t+1/2} - z^*\|^2\right]} \sqrt{\E\left[\text{Err}(t+1/2)\right]} \notag\\
 &\hspace{0.4cm}+ \gamma L_{\max} \E\left[ \| \bar z^{t+1/2} - \bar z^t\|^2\right] + \gamma L \E\left[ \text{Err}(t+1/2)\right].
  \end{align}
  
 \textbf{Proof:} First of all, we use the independence of all random vectors $\xi^{i} = (\xi^{i}_1, \ldots , \xi^{i}_m)$ and select only the conditional expectation $\E_{\xi^{t+1/2}}$ on vector $\xi^{t+1/2}$ and get the following chain of inequalities: 
 \begin{align*}
  - 2 \gamma \E &\left[\langle \bar g^{t+1/2}, \bar z^{t+1/2} - z^* \rangle\right]\\
  &= - 2 \gamma \E\left[\left\langle  \frac{1}{M} \sum\limits_{m=1}^M \E_{\xi^{k+1/2}}[F_m(z_m^{t+1/2}, \xi_m^{t+1/2})], \bar z^{t+1/2} - z^* \right\rangle \right] \\
  &\overset{\eqref{as3}}{=} 
  -2 \gamma \E\left[\left\langle  \frac{1}{M} \sum\limits_{m=1}^M F_m(z_m^{t+1/2}), \bar z^{t+1/2} - z^* \right\rangle\right] \\
  &= - 2 \gamma \E\left[\left\langle  \frac{1}{M} \sum\limits_{m=1}^M F_m(\bar z^{t+1/2}), \bar z^{t+1/2} - z^* \right\rangle\right] \\
  &\hspace{0.4cm}+ 2 \gamma \E\left[\left\langle  \frac{1}{M} \sum\limits_{m=1}^M [F_m(\bar z^{t+1/2}) - F_m(z_m^{t+1/2})], \bar z^{t+1/2} - z^* \right\rangle\right] \\
  &= - 2 \gamma \E\left[\left\langle  F(\bar z^{t+1/2}), \bar z^{t+1/2} - z^* \right\rangle\right] \\
  &\hspace{0.4cm}+ 2 \gamma \E\left[\left\langle  \frac{1}{M} \sum\limits_{m=1}^M [F_m(\bar z^{t+1/2}) - F_m(z_m^{t+1/2})], \bar z^{t+1/2} - z^* \right\rangle\right] \\
  &\overset{\eqref{as2m}}{\leq}2 \gamma \E\left[\left\langle  \frac{1}{M} \sum\limits_{m=1}^M [F_m(\bar z^{t+1/2}) - F_m(z_m^{k+1/2})], \bar z^{t+1/2} - z^* \right\rangle\right] \\
  &\leq 2 \gamma \E\left[ \| \bar z^{t+1/2} - z^*\| \cdot \left\|\frac{1}{M} \sum\limits_{m=1}^M    F_m(\bar z^{t+1/2}) - F_m(z_m^{t+1/2}) \right\| \right]\\
  &\leq 2 \gamma \E\left[ \| \bar z^{t+1/2} - z^*\| \cdot \frac{1}{M} \sum\limits_{m=1}^M  \left\|   F_m(\bar z^{t+1/2}) - F_m(z_m^{t+1/2}) \right\| \right] \\
  &\overset{\eqref{as1l}}{\leq} 2 \gamma L_{\max} \E\left[ \| \bar z^{t+1/2} - z^*\| \cdot \frac{1}{M} \sum\limits_{m=1}^M  \left\|  z_m^{t+1/2} - \bar z^{t+1/2} \right\| \right] \\
 &\overset{}{\leq} 2 \gamma L_{\max} \E\left[ \| \bar z^{t} - z^*\| \cdot \frac{1}{M} \sum\limits_{m=1}^M  \left\|  z_m^{t+1/2} - \bar z^{t+1/2} \right\| \right] \\
  &\hspace{0.4cm}+ 2 \gamma L_{\max} \E\left[ \| \bar z^{t+1/2} - \bar z^t\| \cdot \frac{1}{M} \sum\limits_{m=1}^M  \left\|  z_m^{t+1/2} - \bar z^{t+1/2} \right\| \right]\\
  &\overset{}{\leq} 2 \gamma L_{\max} \sqrt{\E\left[ \| \bar z^{t} - z^*\|^2\right]} \cdot \sqrt{\E\left[ \left(\frac{1}{M} \sum\limits_{m=1}^M  \left\|  z_m^{t+1/2} - \bar z^{t+1/2} \right\| \right)^2 \right]} \\
  &\hspace{0.4cm}+\gamma L_{\max} \E\left[ \| \bar z^{t+1/2} - \bar z^t\|^2\right] + \gamma L_{\max} \E\left[ \left(\frac{1}{M} \sum\limits_{m=1}^M  \| \bar z^{t+1/2} - z_m^{t+1/2}\|  \right)^2\right].
  \end{align*}
By \eqref{eq:squared_sum} it is easy to see that $$ \E\left[ \left(\frac{1}{M} \sum\limits_{m=1}^M  \| \bar z^{t+1/2} - z_m^{t+1/2}\|  \right)^2\right]\leq \E\left[ \frac{1}{M} \sum\limits_{m=1}^M   \| \bar z^{t+1/2} - z_m^{t+1/2}\|^2 \right].$$
This completes the proof.
\EndProof
Then we have the same as \eqref{an6}:
\begin{align*}
 \E&\left[\|\bar z^{t+1} - z^* \|^2\right]  \leq \E\left[\|\bar z^t - z^* \|^2\right] - \E\left[\|\bar z^{t+1/2} - \bar z^t \|^2\right] \notag\\
 &\hspace{0.4cm}+ 2 \gamma L_{\max} \sqrt{\E\left[ \| \bar z^{t} - z^*\|^2\right]} \sqrt{\E\left[\text{Err}(t+1/2)\right]} \\
 &\hspace{0.4cm}+ \gamma L_{\max} \E\left[ \| \bar z^{t+1/2} - \bar z^t\|^2\right] + \gamma L_{\max} \E\left[ \text{Err}(t+1/2)\right]\\
 &\hspace{0.4cm}+ \gamma^2 \left( 5L^2_{\max} \E\left[ \|\bar z^{t+1/2} - \bar z^{t}\|^2\right] + \frac{10 \sigma^2}{M}  + 5L^2_{\max}\E\left[\text{Err}(t+1/2) \right] +5L^2_{\max}\E\left[\text{Err}(t) \right] \right).
  \end{align*}
Choosing $\gamma \leq \frac{1}{4L_{\max}}$ gives
\begin{align*}
   \frac{1}{2}\E&\left[\|\bar z^{t+1/2} - \bar z^t \|^2\right] \leq \E\left[\|\bar z^t - z^* \|^2\right] - \E\left[\|\bar z^{t+1} - z^* \|^2\right] \notag\\
 &\hspace{0.4cm}+ 2 \gamma L_{\max} \sqrt{\E\left[ \| \bar z^{t} - z^*\|^2\right]} \sqrt{\E\left[\text{Err}(t+1/2)\right]}\\
 &\hspace{0.4cm}+ (5\gamma^2L^2_{\max} + \gamma L_{\max})\E\left[\text{Err}(t+1/2) \right] +5\gamma^2L^2_{\max}\E\left[\text{Err}(t) \right]  + \frac{10 \gamma^2 \sigma^2}{M}.
  \end{align*}
Next we work with 
\begin{align*}
 \E&\left[\|\bar z^{t+1/2} - \bar z^{t}\|^2 \right] \\
 &= \gamma^2\E\left[\left\|\frac{1}{M} \sum\limits_{m=1}^M F_m(z^t_m, \xi^{t}_m) - F_m(z^t_m) + F_m(z^t_m) - F_m(\bar z^t) + F_m(\bar z^t) \right\|^2 \right] \\
 &\overset{}{\geq} \frac{\gamma^2}{2} \E\left\| F(\bar z^t) \right\|^2 - \gamma^2\E\left[\left\|\frac{1}{M} \sum\limits_{m=1}^M F_m(z^t_m, \xi^{t}_m) - F_m(z^t_m) + F_m(z^t_m) - F_m(\bar z^t) \right\|^2 \right] \\
 &\overset{}{\geq} \frac{\gamma^2}{2} \E\left\| F(\bar z^t) \right\|^2 - 2\gamma^2\E\left[\left\|\frac{1}{M} \sum\limits_{m=1}^M F_m(z^t_m, \xi^{t}_m) - F_m(z^t_m) \right\|^2 \right] \\
 &\hspace{0.4cm}- 2\gamma^2\E\left[\left\|\frac{1}{M} \sum\limits_{m=1}^M  F_m(z^t_m) - F_m(\bar z^t) \right\|^2 \right] \\
 &\overset{\eqref{as1l}}{\geq} \frac{\gamma^2}{2} \E\left\| F(\bar z^t) \right\|^2 - \frac{2 \gamma^2 \sigma^2}{M} - \frac{2 \gamma^2 L^2_{\max}}{M} \sum\limits_{m=1}^M \E\left[\left\|  z^t_m-\bar z^t \right\|^2 \right] \\
 &= \frac{\gamma^2}{2} \E\left\| F(\bar z^t) \right\|^2 - \frac{2 \gamma^2 \sigma^2}{M} - 2 \gamma^2 L^2_{\max} \E\left[\text{Err}(t)\right].
\end{align*}
Connecting with previous gives
\begin{align*}
   \frac{\gamma^2}{4}\E\left[\|F(\bar z^t)\|^2\right] &\leq \E\left[\|\bar z^t - z^* \|^2\right] - \E\left[\|\bar z^{t+1} - z^* \|^2\right] \notag\\
 &\hspace{0.4cm}+ 2 \gamma L_{\max} \sqrt{\E\left[ \| \bar z^{t} - z^*\|^2\right]} \sqrt{\E\left[\text{Err}(t+1/2)\right]}\\
 &\hspace{0.4cm}+  (\gamma L_{\max} + 5\gamma^2L^2_{\max})\E\left[\text{Err}(t+1/2) \right] +6\gamma^2 L^2_{\max}\E\left[\text{Err}(t) \right] + \frac{11 \gamma^2 \sigma^2}{M}.
  \end{align*}
With result of Lemma \ref{lem14}, we get
\begin{align*}
   \frac{\gamma^2}{4}\E\left[\|F(\bar z^t)\|^2\right] &\leq \E\left[\|\bar z^t - z^* \|^2\right] - \E\left[\|\bar z^{t+1} - z^* \|^2\right] \notag\\
 &\hspace{0.4cm}+ 2 \gamma L_{\max} \sqrt{\E\left[ \| \bar z^{t} - z^*\|^2\right]} \sqrt{216 (D^2 H + \sigma^2) H \gamma^2}\\
 &\hspace{0.4cm}+  \frac{11 \gamma^2\sigma^2}{M}  + 216 (\gamma L_{\max}  + 11\gamma^2 L_{\max} ^2)  (D^2 H + \sigma^2) H \gamma^2 .
  \end{align*}
Summing over all $t$ from $0$ to $T-1$ and averaging gives
\begin{align}
\label{temp202}
\E\left[ \frac{1}{T}\sum\limits_{t=0}^{T-1}\|F(\bar z^t)\|^2\right] &\leq \frac{4\|z^0 - z^* \|^2}{\gamma^2 T} + \frac{44 \sigma^2}{M}\notag\\
 &\hspace{0.4cm}+ 1000 (\gamma L_{\max}  + 11\gamma^2 L_{\max} ^2)  (D^2 H + \sigma^2) H  \notag\\
 &\hspace{0.4cm}+ \frac{120  L_{\max} \sqrt{(D^2 H + \sigma^2) H} }{T} \sum\limits_{t=0}^{T-1} \sqrt{\E\left[ \| \bar z^{t} - z^*\|^2\right]} 
.
\end{align}
Under the additional assumption that $\|z^*\| \leq \Omega$ and $\|\bar z^t\| \leq \Omega$, from \eqref{temp202}, we obtain
\begin{align*}
\E\left[ \frac{1}{T}\sum\limits_{t=0}^{T-1}\|F(\bar z^t)\|^2\right] &= \mathcal{O} \Bigg(\frac{\|z^0 - z^* \|^2}{\gamma^2 T} + (\gamma L_{\max}  + \gamma^2 L_{\max} ^2)  (D^2 H + \sigma^2) H  \notag\\
 &\hspace{1cm}+ \frac{\sigma^2}{M} + L_{\max} \Omega \sqrt{(D^2 H + \sigma^2) H} \Bigg)
.
\end{align*}
With $\gamma = \min\left\{\frac{1}{4L_{\max}}; \left(\frac{\|\bar z^0 - z^* \|^2}{TL_{\max} (D^2 H + \sigma^2) H}\right)^{1/3}\right\}$, we have
\begin{align*}
\E\left[ \frac{1}{T}\sum\limits_{t=0}^{T-1}\|F(\bar z^t)\|^2\right] &= \mathcal{O} \Bigg(\frac{L^2_{\max} \|z^0 - z^* \|^2}{T}  + \frac{(L_{\max} \Omega (D^2 H + \sigma^2) H)^{2/3}}{T^{1/3}} \\
&\hspace{1cm}+ \frac{\sigma^2}{M}  + L_{\max} \Omega \sqrt{(D^2 H + \sigma^2) H}  \Bigg).
\end{align*}
\EndProof

\section{Experiments}\label{add_exp}

We implement all methods in Python 3.8 using
PyTorch \cite{pytorch} and Ray \cite{ray} and run on a machine with 24 AMD EPYC 7552 @ 2.20GHz processors,
2 GPUs  NVIDIA A100-PCIE with 40536 Mb of memory each (Cuda 11.3).

\subsection{Federated GAN on MNIST}\label{sec:mnist_exp}

We continue with further experiments on Generative Adversarial Networks. But first, a short introduction. A simple GAN setup consists of two parts -- the discriminator $D$, which aims to distinguish real samples $x$ from adversarial ones by giving a probability that the sample is real, and the generator $G$, which tries to fool the discriminator by generating realistic samples from random noise $z$. Following \cite{goodfellow2014generative}, the value function $V(G, D)$ used in such a min-max game can be expressed as 
\begin{eqnarray}\label{eq::gan_vf}
    \min _{G} \max _{D} V(D, G)=\mathbb{E}_{\boldsymbol{x} \sim p_{\mathrm{data}}(\boldsymbol{x})}[\log D(\boldsymbol{x})] +\mathbb{E}_{\boldsymbol{z} \sim p_{\boldsymbol{z}}(\boldsymbol{z})}[\log (1-D(G(\boldsymbol{z})))].
\end{eqnarray}

As mentioned in main part, we use Deep Convolutional GAN \cite{dcgan}. 
As optimizers we use Algorithm \ref{alg4} and a combination of Adam with Algorithm \ref{alg4}.



We make up to 3 or 4 replicas and train them for 200 epochs. In these experiments we try to vary the synchronisation frequencies for the generator and the discriminator separately.
We split the data as follows: one half of the data set is divided equally between the replicas, and from the other half we take only those digits that correspond to the order number of the replica.

Usually, to get better performance, researchers vary the number of training steps done for the generator and the discriminator, pretrain one of the parts or use specific optimizer. We are more interested in numerical convergence, that is why we do not do such fine-tuning.

\begin{figure}[h]
\centering
\begin{minipage}{0.25\textwidth}
\includegraphics[width=\textwidth]{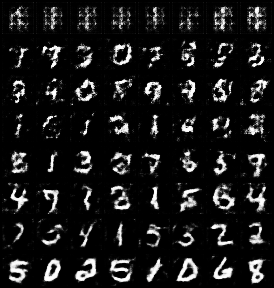}
\end{minipage}%
\begin{minipage}{0.05\linewidth}
~~~
\end{minipage}
\begin{minipage}{0.25\textwidth}
\includegraphics[width=\textwidth]{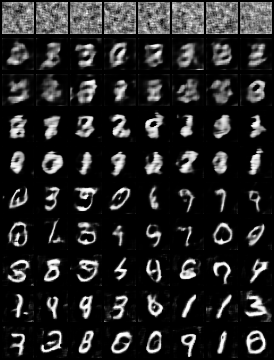}
\end{minipage}
\caption{Digits generated by global generator during training. 4 replicas, Local SGD (left) and 4 replicas, Local Adam (right) $H_g=H_d =20$.}
\label{mnistprogress}
\end{figure}

\begin{figure}[h!]
\centering
\begin{minipage}{0.5\textwidth}
\includegraphics[width =  \textwidth ]{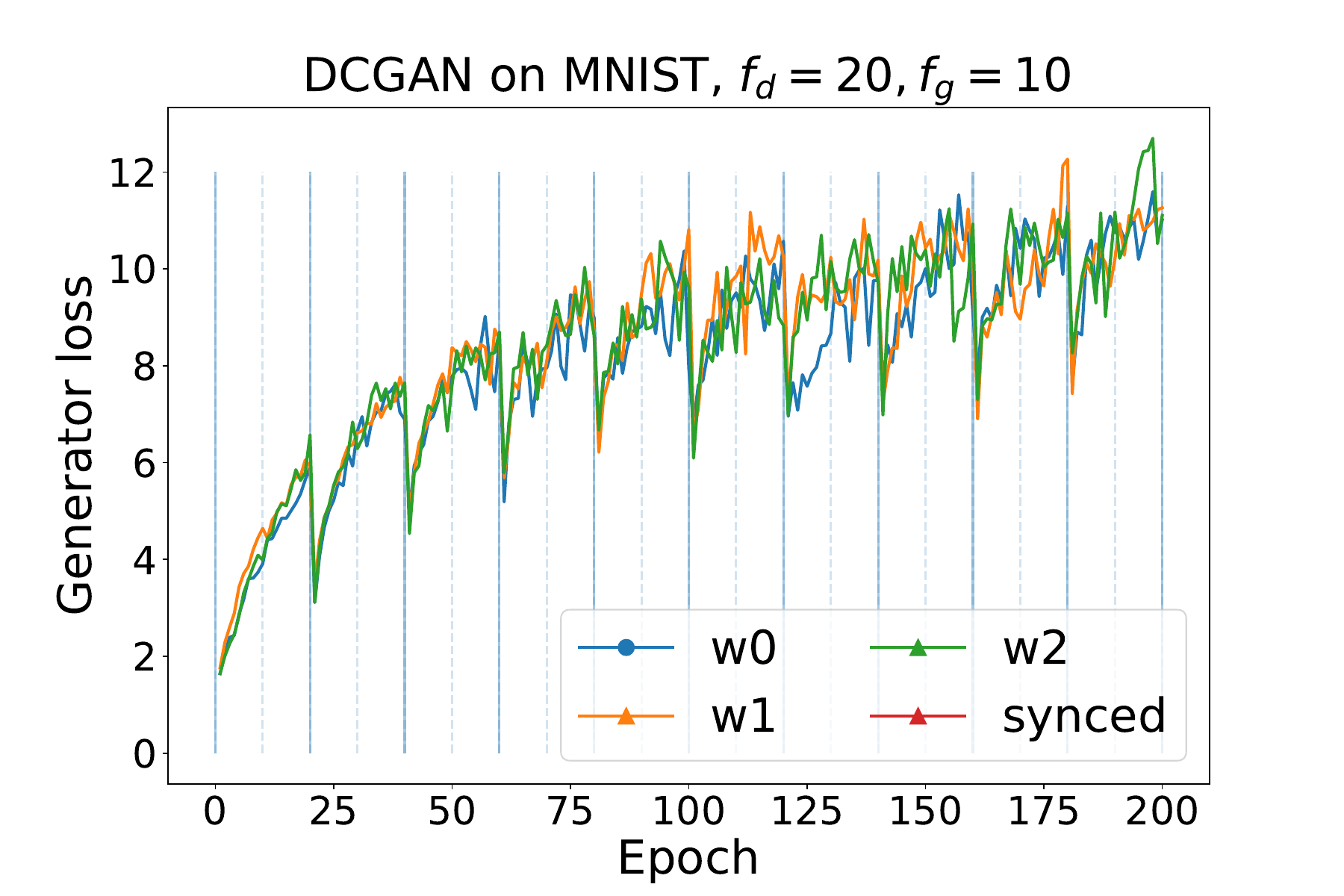}
\end{minipage}%
\begin{minipage}{0.5\textwidth}
  \centering
\includegraphics[width =  \textwidth ]{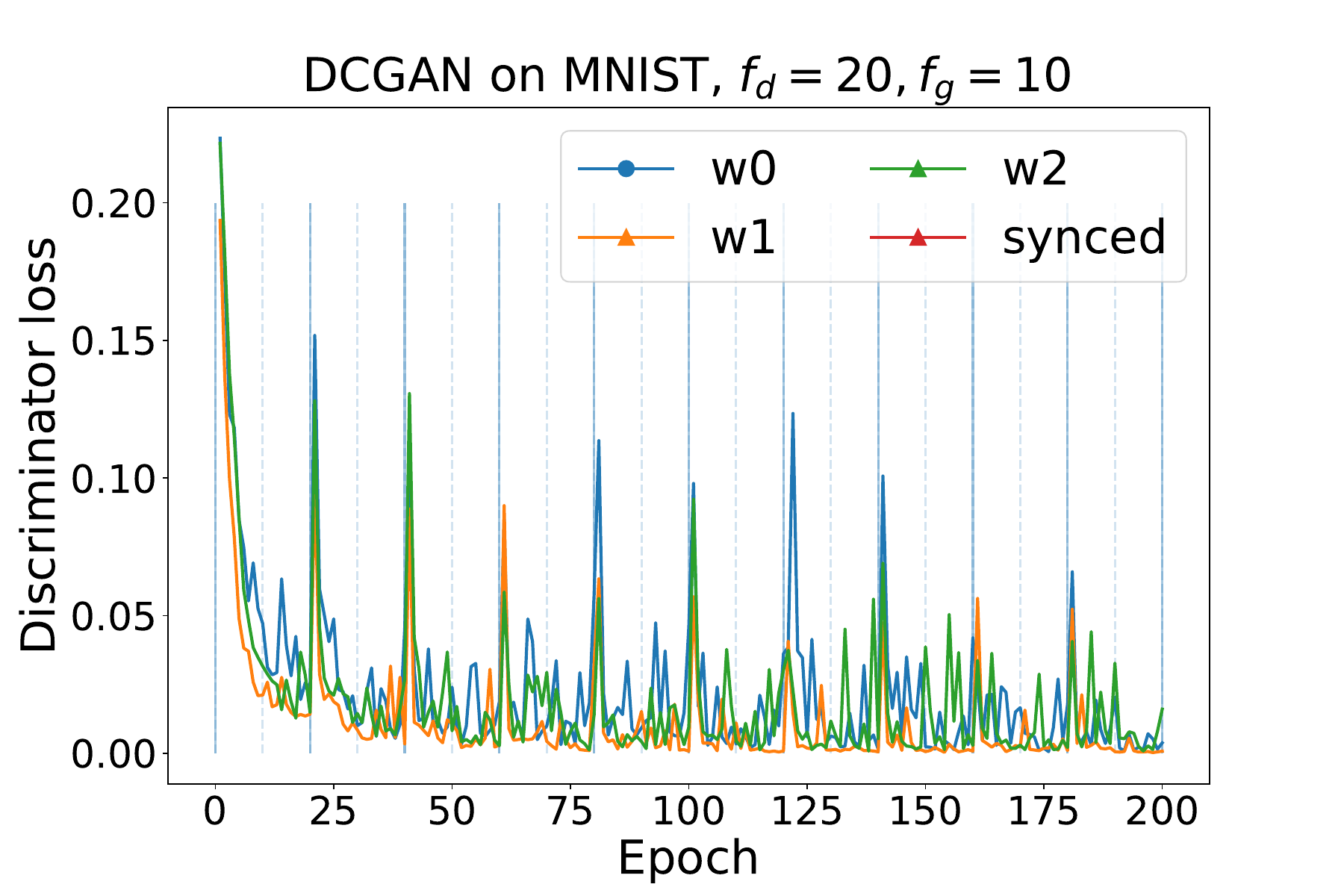}
\end{minipage}%
\caption{Generator and Discriminator Empirical Loss on MNIST during training, Local SGD, 3 replicas, $H_g = 10,\ H_d = 20$.}
\label{fig:mnist_gan_10_20_3}
\end{figure}

\begin{figure}[h!]
\centering
\begin{minipage}{0.5\textwidth}
\includegraphics[width =  \textwidth ]{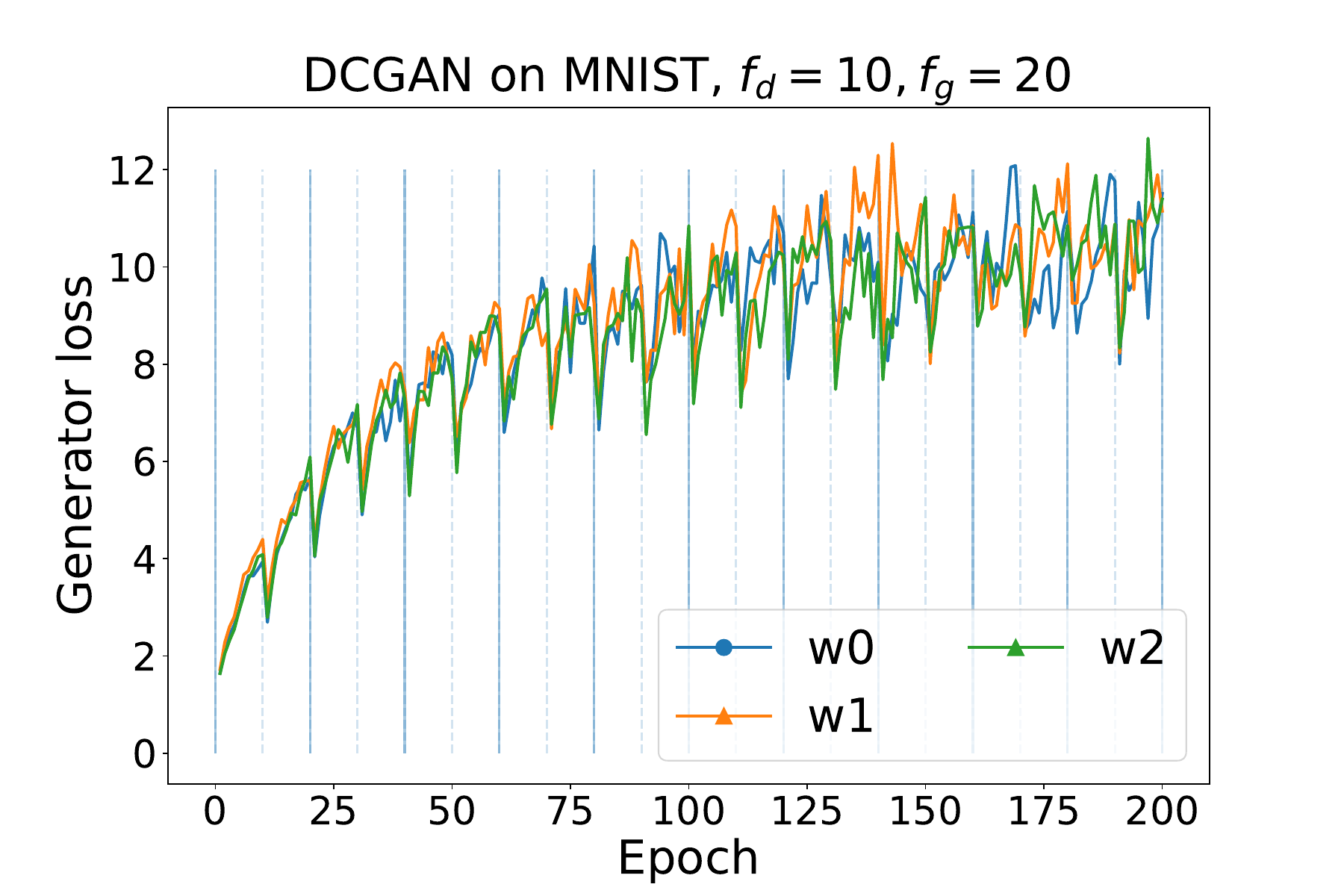}
\end{minipage}%
\begin{minipage}{0.5\textwidth}
  \centering
\includegraphics[width =  \textwidth ]{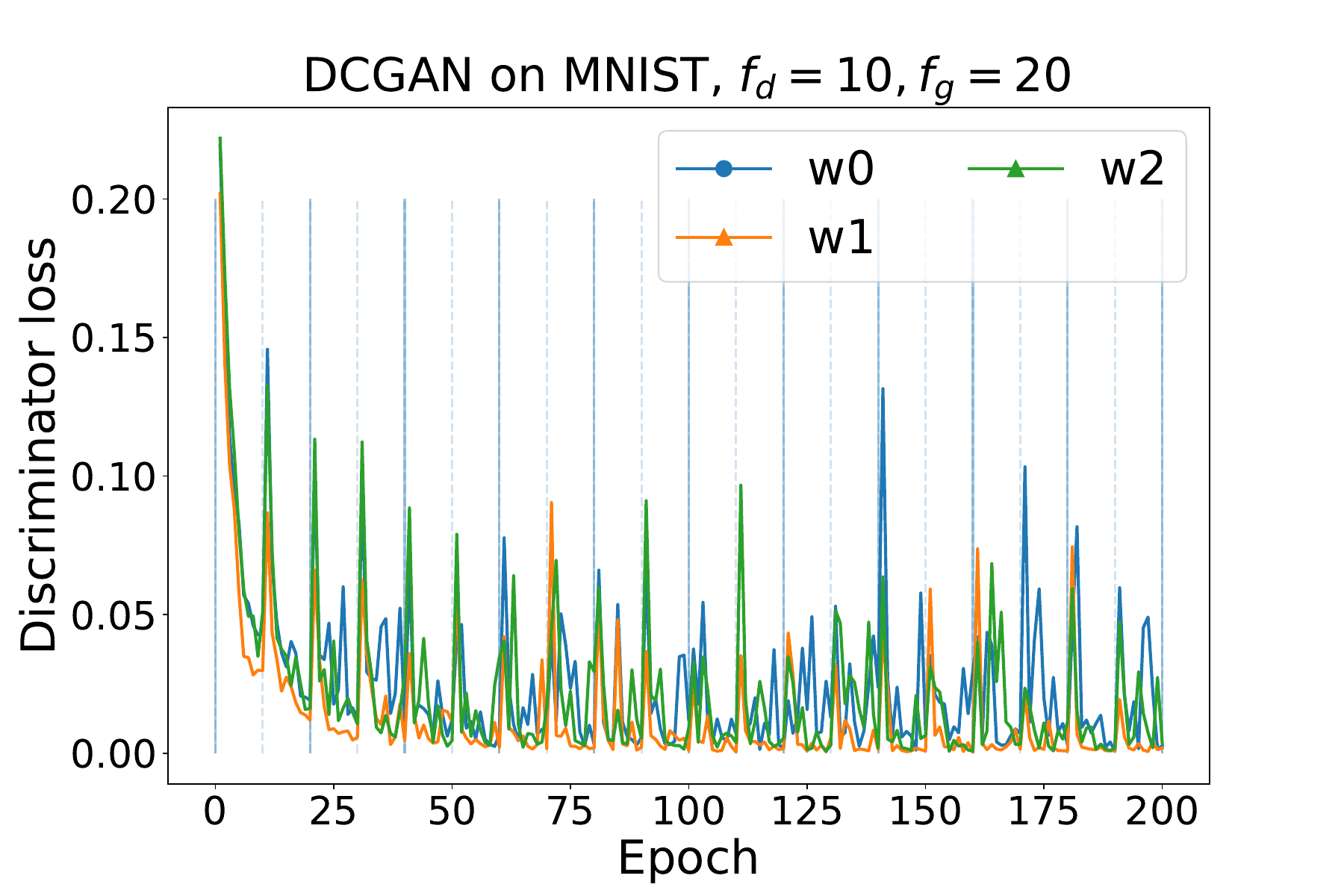}
\end{minipage}%
\caption{Generator and Discriminator Empirical Loss on MNIST during training, Local SGD, 3 replicas, $H_g = 20,\ H_d = 10$.}
\label{fig:mnist_gan_20_10_3}
\end{figure}

\begin{figure}[h!]
\centering
\begin{minipage}{0.5\textwidth}
\includegraphics[width =  \textwidth ]{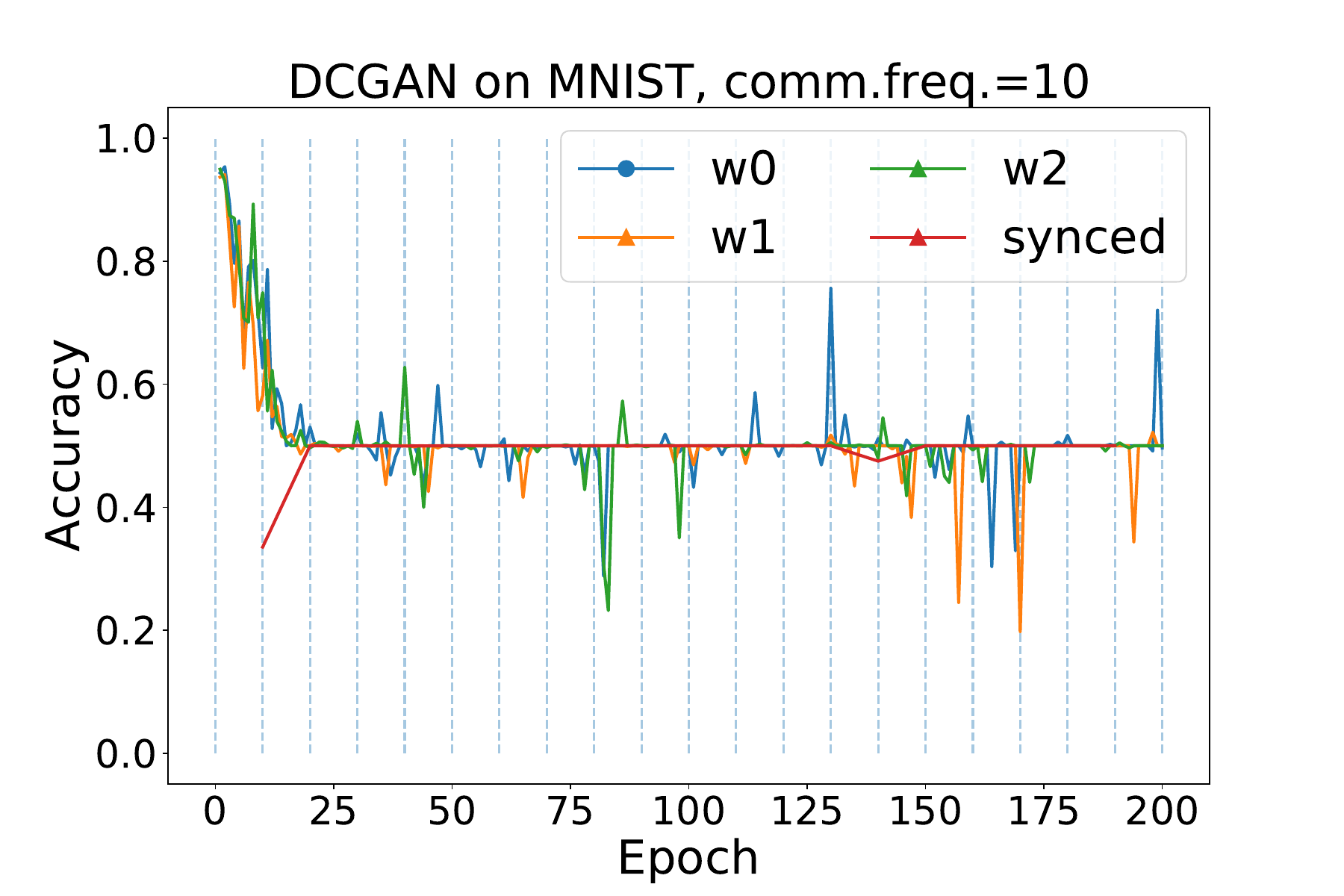}
\end{minipage}%
\begin{minipage}{0.5\textwidth}
\includegraphics[width =  \textwidth ]{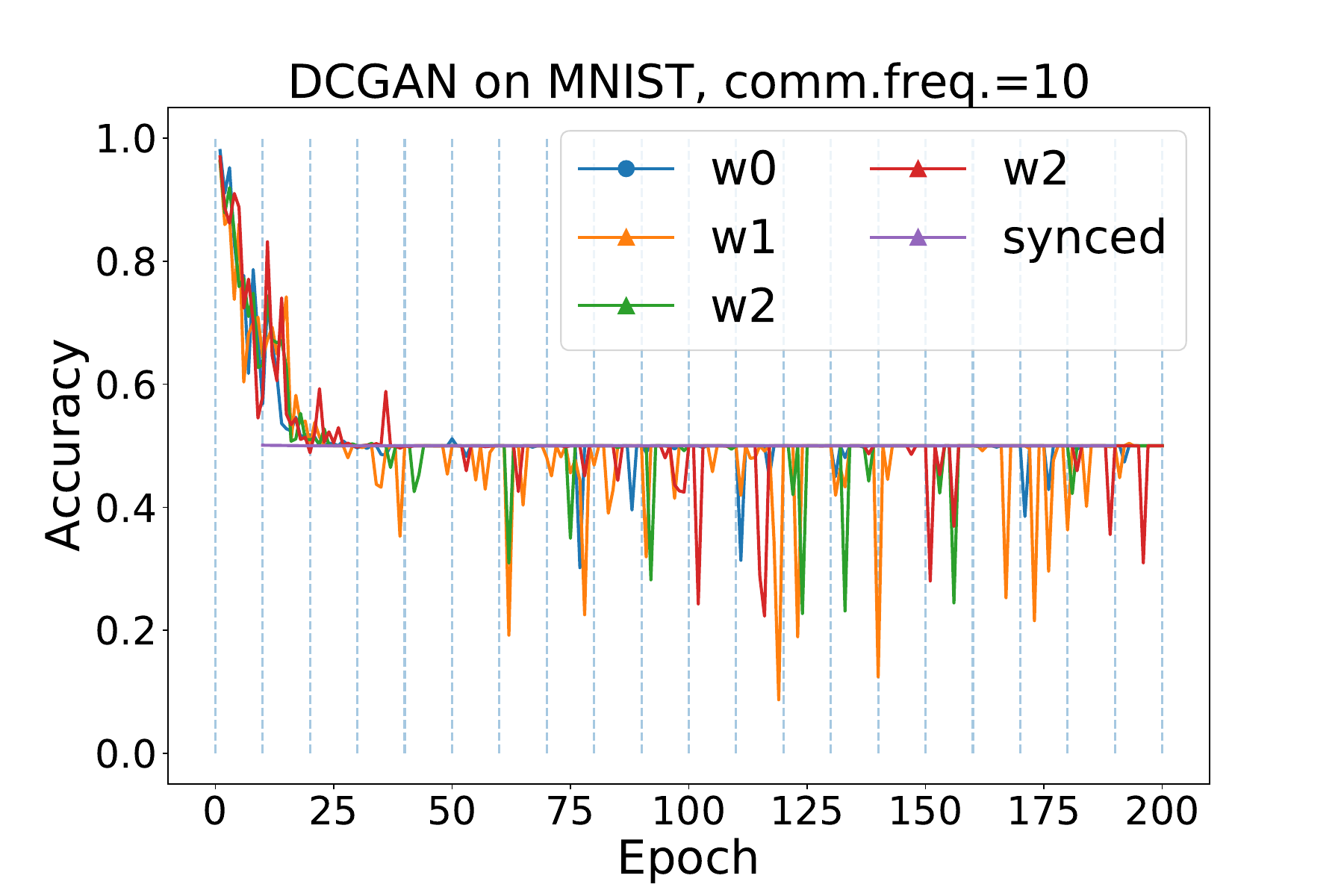}
\end{minipage}%
\caption{Accuracy on MNIST, Local SGD, 3 and 4 replicas, $H_g = H_d = 20$.}
\label{fig:mnist_acc_3_4}
\end{figure}

\begin{figure}[h!]
\centering
\begin{minipage}{0.5\textwidth}
\includegraphics[width =  \textwidth ]{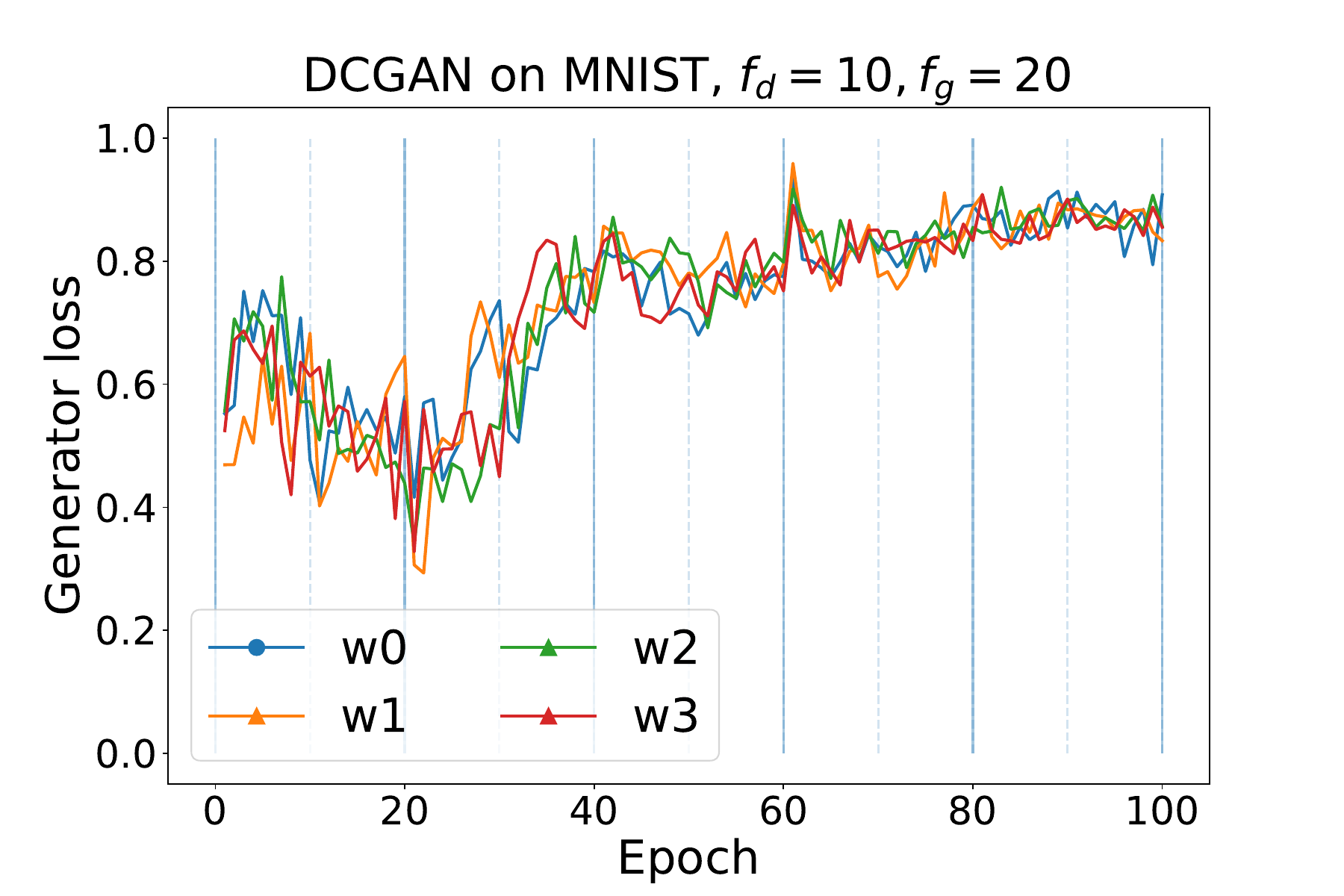}
\end{minipage}%
\begin{minipage}{0.5\textwidth}
  \centering
\includegraphics[width =  \textwidth ]{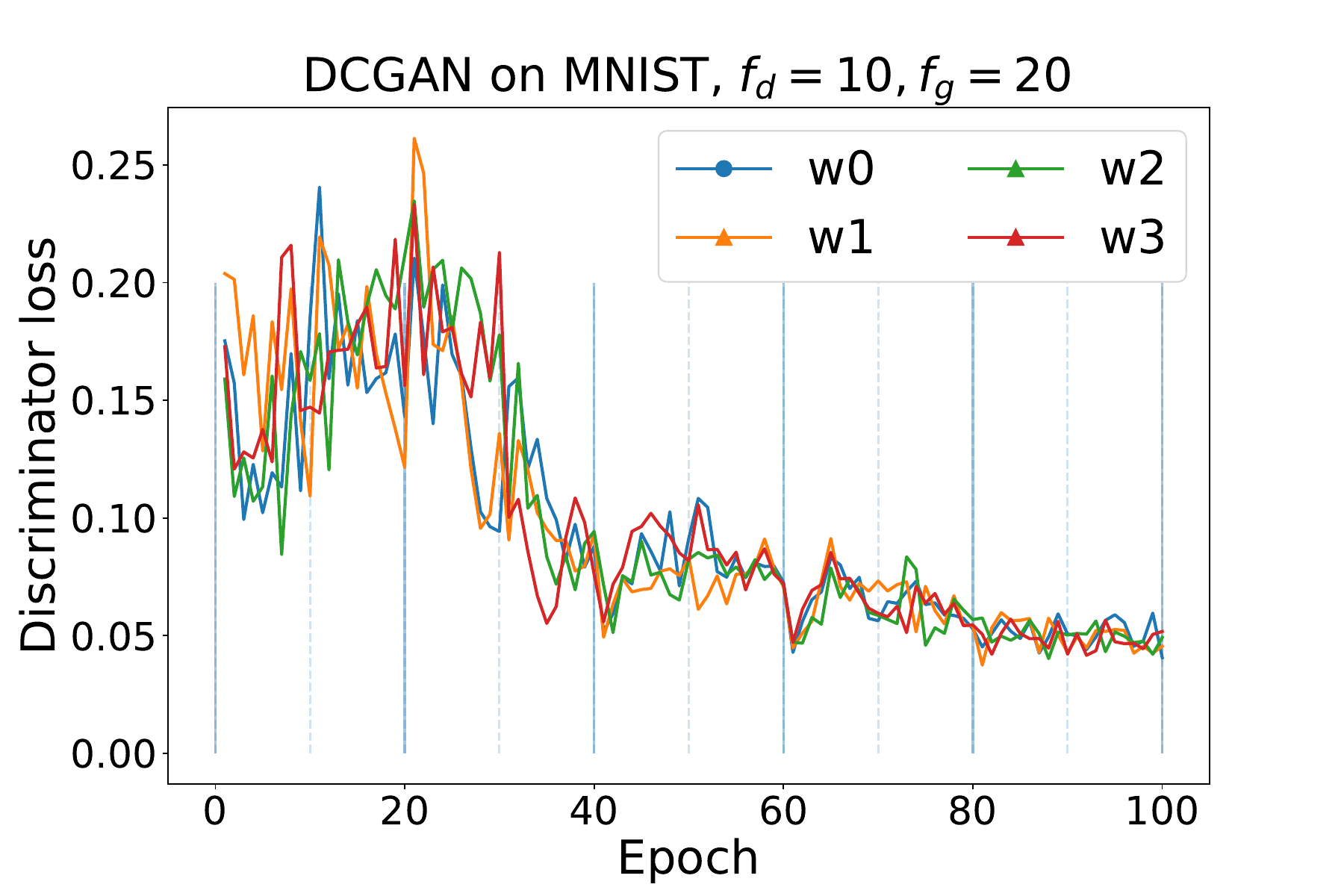}
\end{minipage}%
\caption{Generator and Discriminator Empirical Loss on MNIST during training, Local Adam, 4 replicas, $H_g = 20,\ H_d = 10$.}
\label{fig:mnist_gan_adam_20_10_4}
\end{figure}

\begin{figure}[h!]
    \centering
    \includegraphics[width=0.5\textwidth]{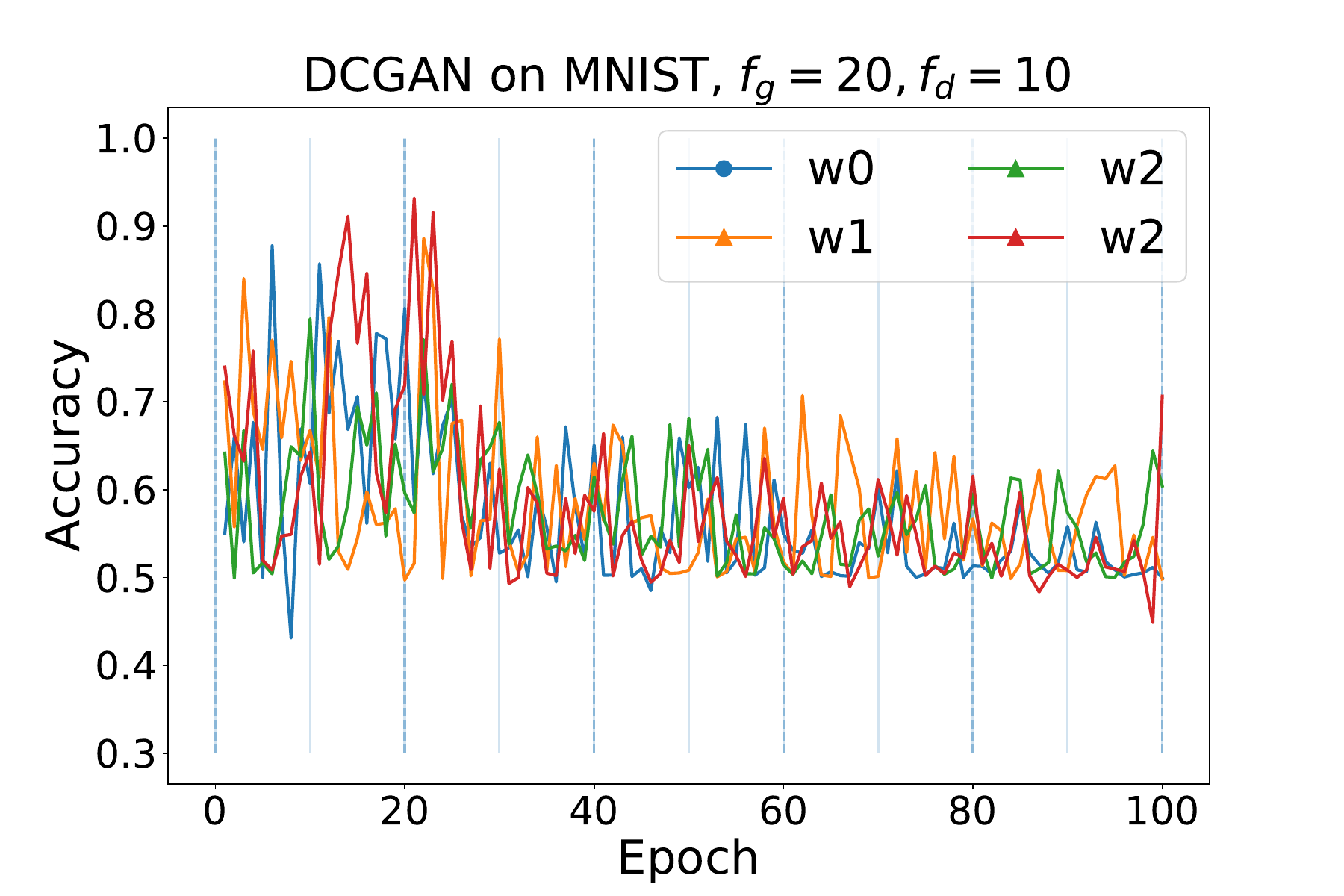}
    \caption{Accuracy on MNIST, Local Adam, 4 replicas, $H_g = 20,\ H_d = 10$.}
    \label{fig:my_label}
\end{figure}

The results of the experiment for Algorithm \ref{alg4} and Local Adam are reflected in Figures \ref{mnistprogress}, \ref{fig:mnist_acc_3_4}, \ref{fig:mnist_gan_10_20_3}, \ref{fig:mnist_gan_20_10_3}, \ref{fig:mnist_gan_adam_20_10_4} and \ref{fig:my_label}. Here $H_g$, $H_d$ -- communication frequencies for generator and discriminator.

The experiment shows that we have good global images despite the fact that the data is heterogeneous. 
The global (synced) discriminator converges to random guessing, which is indicated by a binary classification accuracy equal to $0.5$, see Figures \ref{fig:mnist_acc_3_4}, \ref{fig:my_label}. Based on \cite{goodfellow2014generative}, this behaviour is expected.

\end{document}